\pdfoutput=1 
\documentclass[12pt]{article}
\usepackage{arxiv}
\usepackage[T1]{fontenc}
\usepackage[british]{babel}
\usepackage{csquotes}
\usepackage{microtype}
\usepackage{times}
\usepackage{xcolor}
\usepackage{mathtools,amssymb,amsthm}
\mathtoolsset{showonlyrefs,showmanualtags}
\usepackage{bm}
\usepackage{dsfont}
\usepackage{my_notation}
\usepackage[authoryear,square]{natbib}
\usepackage{graphicx}
\usepackage{placeins}
\usepackage{float}
\usepackage{algorithm}
\usepackage{algpseudocode}
\usepackage{booktabs}
\usepackage{longtable}
\usepackage{tikz}
\usepackage{enumitem}
\usepackage{array}
\usepackage{multirow}
\usepackage{alphalph}
\usepackage{etoolbox}
\usepackage{xurl}
\usepackage{hyperref}
\usepackage{bookmark}

\theoremstyle{plain}
\newtheorem{theorem}{Theorem}
\newtheorem{lemma}[theorem]{Lemma}
\newtheorem{proposition}[theorem]{Proposition}
\newtheorem{corollary}[theorem]{Corollary}
\theoremstyle{definition}
\newtheorem{definition}[theorem]{Definition}

\newenvironment{proofof}[1]{\begin{proof}[Proof of #1]}{\end{proof}}

\usepackage{paper_layout}
\renewenvironment{keywords}
 {\par\small\noindent\textbf{Keywords:}\ }
 {\par}

\algrenewcommand{\algorithmiccomment}[1]{\hfill\textcolor{black!55}{\itshape #1}}
\newcounter{algeq}
\renewcommand{\thealgeq}{A\arabic{algeq}}
\newcommand{\algtag}[1]{\hfill\refstepcounter{algeq}\textup{(\thealgeq)}\label{#1}}

\newlist{lemenum}{enumerate}{1}
\setlist[lemenum]{label=(\alph*),ref=\thetheorem(\alph*),beginpenalty=10000,midpenalty=10000,%
  leftmargin=*,topsep=2pt,itemsep=1pt,parsep=0pt}

\definecolor{dkblue}{cmyk}{1,.54,.04,.19}
\hypersetup{
 colorlinks=true,
 linkcolor=black,
 citecolor=dkblue,
 urlcolor=dkblue,
 filecolor=dkblue,
 pdftitle={Curvature-Matched Exploration: High-Probability Õ(√(dT)) Regret for Adversarial Linear Bandits on the Euclidean Ball},
 pdfauthor={Jack J. Mayo},
 pdfsubject={Curvature-matched exploration and second-order regret for adversarial linear bandits},
 pdfkeywords={adversarial linear bandits, barrier curvature, data-dependent regret, implicit exploration}
}

\renewcommand{\R}{\mathbb R}
\renewcommand{\E}{\mathbb E}
\newcommand{\bbP}{\mathbb P}
\newcommand{\transpose}{^\mathsf{\scriptscriptstyle T}}
\DeclarePairedDelimiterX{\twonorm}[1]{\lVert}{\rVert}{{#1}}
\DeclarePairedDelimiterX{\iprod}[2]{\langle}{\rangle}{{#1,#2}}
\let\II\relax
\DeclarePairedDelimiterXPP{\II}[1]{\mathds 1}{\{}{\}}{}{#1}
\newcommand{\opnorm}[1]{\norm{#1}_{\mathrm{op}}}
\newcommand{\normt}[2]{\norm{#2}_{#1}}
\newcommand{\Reg}{\mathrm{Reg}}
\newcommand{\tr}{\operatorname{tr}}
\newcommand{\sphere}{\mathbb S}
\newcommand{\ball}{B_2^d}
\newcommand{\Qcurv}{\mathsf V}
\newcommand{\EestObs}{\widehat V}
\renewcommand{\argmin}{\operatorname*{arg\,min}}

\providecommand{\qixproofstage}[2]{\paragraph{Step #1: #2.}}

\title{Curvature-Matched Exploration: High-Probability \texorpdfstring{\(\widetilde O(\sqrt{dT})\)}{Õ(√(dT))}\\[3pt]
  Regret for Adversarial Linear Bandits on the Euclidean Ball}
\renewcommand{\shorttitle}{Curvature-Matched Exploration: High-Probability \(\widetilde O(\sqrt{dT})\) Regret on the Euclidean Ball}

\author{Jack J. Mayo\\
Kurtos.ai\\
Korteweg-de Vries Institute for Mathematics\\
University of Amsterdam\\
\texttt{jack@kurtos.ai}}
\date{}

\begin{document}

\raggedbottom
\setlength{\emergencystretch}{3em}
\medmuskip=4mu plus 2mu minus 2mu
\maketitle

\begin{abstract}
We study adversarial linear bandits on the \(d\)-dimensional Euclidean unit ball
with exact scalar feedback, giving an algorithm attaining \(\widetilde O(\sqrt{dT})\) anytime regret for every nonanticipating adaptive adversary, in high-probability, and using \(O(d)\) time and memory per round. By matching the second
moment of the exploratory distribution to the inverse Hessian of the spherical log-barrier, the variance of
each comparator's estimated loss difference can be explicitly controlled through its squared distance
from the leader. Then, by occasionally contracting the accumulated estimate, the algorithm can earn credit proportional to that same squared distance in the regret bound. Uniform control over the ball then follows through a martingale dilation argument, avoiding the \(\sqrt d\) naive covering introduces when controlling estimator deviation. A second-order bound in the
cumulative local squared losses and a term associated with radial drift also falls naturally out of the same analysis. On aligned losses
bounded away from zero, the algorithm incurs \(\widetilde\Theta(T^{1/3})\)
regret at fixed dimension and confidence; conversely, we show that any nonanticipating
learner guaranteeing regret at most an additive term plus a multiple of the
square root of the cumulative local squared losses at confidence
\(1-1/(16T)\) requires the additive term or squared multiplier to grow as
\(\Omega(T^{1/4})\).
\end{abstract}


\section{Introduction}
\label{sec:introduction}
On each round of a linear bandit game on the Euclidean unit ball
\(\ball\subset\R^d\), the adversary chooses a loss vector \(\theta_t\in\ball\)
before the learner draws an action \(a_t\in\ball\). The learner observes only
\(\iprod{a_t}{\theta_t}\). We measure performance in terms of regret against the best fixed action
on the realised loss sequence. Since an adaptive adversary can choose later
losses in response to earlier actions, this comparator can depend on the
learner's random draws.

\citet[Theorem~6]{bubeck2012minimax} obtain \(\widetilde O(\sqrt{dT})\)
expected regret against a fixed comparator and leave the high-probability
extension open. The rate is optimal up to logarithms for \(T\ge d^3\)
\citep[Theorem~5.2]{jacobsen2026perturbation}. For realised regret against the
best fixed action in hindsight, the best previous high-probability bound is
\(\widetilde O(d\sqrt T)\), obtained for general convex action sets by
\citet[Theorem~5]{zimmert2022bias}. On the ball we recover the expected rate
for this stronger requirement. Write
\(\Reg_T=\sum_{t\le T}\iprod{a_t}{\theta_t}-\min_{u\in\ball}\sum_{t\le T}\iprod{u}{\theta_t}\). Our main result is:
\begin{theorem}[Anytime regret on the Euclidean ball]
\label{thm:worst-case}
Let \(d\ge2\) and \(\delta\in(0,1)\). Against every nonanticipating
adaptive adversary, with probability at least \(1-\delta\),
Algorithm~\ref{alg:radial-qix} satisfies, simultaneously for all \(T\ge1\),
\[
 \Reg_T\le C_0\sqrt{dT\log(eT)\,\bigl\{\log(e/\delta)+\log\log(e^2T)\bigr\}},
\]
where \(C_0>0\) is an absolute constant.
\end{theorem}
The bound matches the \(\sqrt{dT\log T}\) expected rate of
\citet{bubeck2012minimax} for any $T\geq 1$ with sub-Gaussian \(\sqrt{\log(e/\delta)}\)
confidence scaling, up to the anytime \(\log\log(e^2T)\) term.
To choose the comparator in hindsight, we need the regret bound to hold
for every action on the same event. Covering the unit ball of
any norm on \(\R^d\) at scale \(\varepsilon\in(0,1]\) requires between
\(\varepsilon^{-d}\) and \((2/\varepsilon+1)^d\) points
\citep[Section~4.2]{vershynin2018high}, so a union bound over a fixed-scale net
is insensitive to volume: the cube and the ball incur the same \(\sqrt d\)
covering cost. Such a union bound treats net points separately, ignoring the fact that comparators in any open neighbourhood on the boundary have closely
coupled incurred losses, which are measured through strongly correlated noise from our loss estimates. This correlation is a direct consequence of the boundary curvature of \(\ball\), which can be exploited both for better dimension
dependence in linear bandits \citep{bubeck2012minimax,kerdreux2021linear}
and for fast rates under full information
\citep{huang2017following,kerdreux2021projectionfree,tsuchiya2024fast,mayo2026smallloss}. A key insight of our approach is that the same curvature also lets us avoid a separate fixed-scale union bound: instead, we use it to
cover each comparator's excess estimated loss online with a single quadratic
form, so one vector bound controls the estimation errors \emph{jointly}.

\Needspace{7\baselineskip}
\paragraph{Approach.}
Let \(e_t\) be a unit action that minimises loss under accumulated
estimate \(\widehat\Theta_{t-1}\). For any unit comparator \(v\), the excess estimated
loss over the leader \(e_t\) is
\begin{equation*}
\iprod{v-e_t}{\widehat\Theta_{t-1}}=\tfrac12\twonorm{\widehat\Theta_{t-1}}\twonorm{v-e_t}^2,	
\end{equation*}
since \(1-\iprod v{e_t}=\tfrac12\twonorm{v-e_t}^2\) on the sphere.
This identity lets us use the same gap to bound the noise in a comparison
and to offset that noise in the regret bound.

\begin{figure}[htbp]
\centering
\begin{tikzpicture}[>=stealth,font=\small]
\begin{scope}[scale=1.8]
 \draw[dashed,fill=gray!20,rotate=35] (0,0) ellipse (0.5 and 1);
 \draw[thick] (0,0) circle (1);
 \draw[gray] (0,0) -- (35:1);
 \draw[->,thick,gray!70!black] (0,0) -- (215:0.32);
 \node at (215:0.18) [below right=1pt] {\(\widehat\Theta_{t-1}\)};
 \fill (0,0) circle (0.022);
 \draw[<->] ([shift={(125:0.1)}]0,0) -- ([shift={(125:0.1)}]35:0.5)
  node[midway,above left=-2pt] {\(\sqrt{p_t}\)};
 \draw[<->] ([shift={(125:0.1)}]35:0.75) -- ([shift={(125:0.1)}]35:1)
  node[midway,above left=-1pt] {\(p_t\)};
 \fill (35:1) circle (0.04);
 \node[align=left,anchor=west,inner sep=1pt] at ([shift={(0.09,0.02)}]35:1) {\(e_t\)\\[-2pt]\footnotesize w.p.\ \(1-p_t\)};
 \draw[fill=white] (35:0.75) circle (0.035);
 \node at (35:0.75) [below right=-1pt] {\(z_t\)};
 \draw[gray] (-0.700,0.566) -- (-0.93,0.95);
 \node[align=right,anchor=south east,inner sep=1pt] at (-0.93,0.95)
   {\footnotesize explore with\\[-2pt]\footnotesize probability \(p_t\)};
 \node at (0,-1.24) {(a) one round of sampling};
\end{scope}
\begin{scope}[xshift=6.0cm,yshift=-1.9cm,scale=2.15]
 \draw[thick] ([shift={(0,-1)}]68:2) arc (68:112:2);
 \node[gray] at (-0.75,0.81) [left] {\footnotesize boundary};
 \draw[gray] (-0.6,1) -- (1.0,1);
 \draw[gray,densely dotted] (0,1) -- (0,0.4);
 \node[gray] at (0,0.45) [left] {\footnotesize radial};
 \fill (0,1) circle (0.035) node[above=1pt] {\(e_t\)};
 \draw[->,very thick] (0,1) -- (0.75,0.5);
 \node at (0.33,0.72) [below left=-2pt] {\(\theta_t\)};
 \draw[->,dashed,thick] (0,1) -- (0.75,1);
 \node[align=center] at (0.42,1) [above=1pt] {\footnotesize tangential\\[-2pt]\footnotesize weight \(1\)};
 \draw[->,dashed,thick] (0.75,1) -- (0.75,0.52);
 \node[align=left] at (0.75,0.76) [right=2pt] {\footnotesize radial\\[-2pt]\footnotesize weight \(p_t\)};
 \node at (0.25,-0.15) {(b) weights of the loss quadratic form};
\end{scope}
\end{tikzpicture}
\caption{One QIX round on the ball with \(d=2\) and \(p_t=1/4\).
(a) On an exploration round the learner plays \(G_t^{1/2}U_t\) on the
dashed ellipse. Its mean action over both branches is \(z_t\), at distance
\(p_t\) from the boundary. (b) The weights of the loss
quadratic form \eqref{eq:q-clock}.}
\label{fig:qix-geometry}
\end{figure}
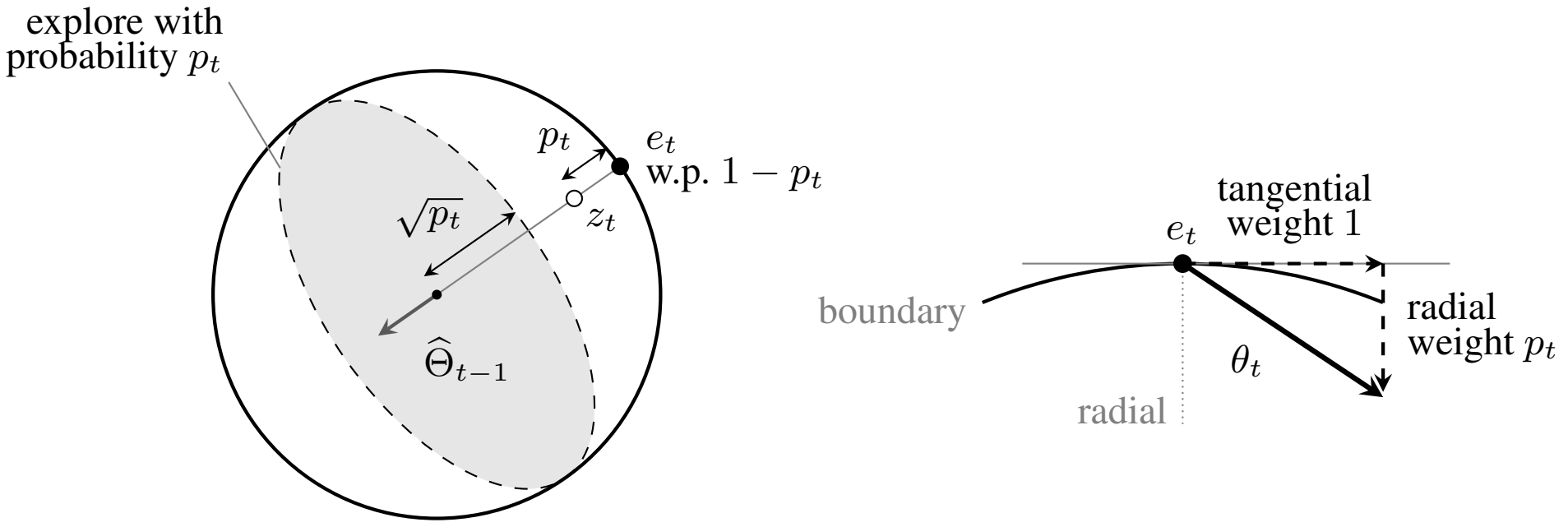

We base our algorithm on Follow the Regularised Leader (FTRL), with the radial barrier of
\citet{bubeck2012minimax}. The learner plays \(e_t\) with probability proportional to its mean radius, and explores proportionally to its distance from the boundary \citep[Section~5.2]{abernethy2009beating}. In place of the signed-coordinate exploration these works utilise, we introduce a scheme whose second moment is designed to match the inverse barrier Hessian,
as in SCRiBLe \citep{AHR08}. This relatively standard idea allocates more exploration perpendicular to the leader than along it near the boundary (Figure~\ref{fig:qix-geometry}).
The same metric measures stability, the observed squares used to tune the
inverse learning rate, and the loss quadratic form of Section~\ref{sec:main-result}.
\emph{Quadratic implicit exploration} (QIX) adapts implicit exploration
\citep{kocak2014efficient,neu2015explore} to this metric to control stability under rare, high-variance observations.

A matrix test adapted from \citet[Algorithm~2]{lee2020bias} tells us when
to shorten \(\widehat\Theta_{t-1}\) (Section~\ref{sec:algorithm},
Figure~\ref{fig:trigger-test}). Each contraction contributes a
linear bias in the FTRL analysis \citep[Section~2.1]{zimmert2022bias},
which allows us to subtract a fraction of each comparator's gap from its regret bound.
This gap also bounds the variance of the comparison, as in
\citet[Section~5.2]{abernethy2009beating}. The test keeps each unit
comparator's scaled gap below the squared norm of \((-v,1)\) in a single
positive definite matrix, so bounding the noise vector in the dual norm
controls all comparators at once. Because this matrix grows online,
normalising by it turns the noise sum into a Hilbert-space recursion under predictable contractions. The crucial step to control all comparators without a incurring a log-determinant
penalty is then to dilate each contraction to an isometry
(Section~\ref{sec:contractive-concentration}). The resulting
noise bound then grows with the square root of the accumulated gap between each comparator and the leader, while the
credit from contractions grows linearly in the same quantity (Figure~\ref{fig:cover-main}b).
The credit also absorbs the QIX bias, as in earlier bonus methods
\citep{AuerCFS02,bartlett2008high,abernethy2009beating}.
Section~\ref{sec:proof-strategy} gives the argument.

\paragraph{Data-dependent bounds and their limits.}
Keeping the cumulative loss quadratic form
\(\sum_{t\le T}(\twonorm{\theta_t}^2-\twonorm{z_t}\iprod{e_t}{\theta_t}^2)\)
(where \(z_t\) is the mean action at round \(t\)) yields a second-order
regret bound with a residual for excess radial drift that vanishes under
sufficient tangential loss. Even when all losses point along a single ray
(\(\theta_t=-\gamma_tu\) with \(\twonorm u=1\) and \(\gamma_t\in[\gamma_0,1]\)),
however, exploration has a cost: QIX incurs \(\widetilde\Theta(T^{1/3})\)
regret for fixed \(d,\delta,\gamma_0>0\). More generally, any
nonanticipating learner guaranteeing
\(\Reg_T\le\mathsf F+\mathsf B\sqrt{\sum_{t\le T}(\twonorm{\theta_t}^2-\twonorm{z_t}\iprod{e_t}{\theta_t}^2)}\)
at confidence \(1-1/(16T)\) must satisfy
\(\mathsf F+\mathsf B^2+1=\Omega(T^{1/4})\). From the second-order bound,
Appendix~\ref{sec:consequences} also derives guarantees in terms of the
cancellation slack
\(\Delta_T=\sum_t\twonorm{\theta_t}-\twonorm{\sum_t\theta_t}\) and, for
adaptive random losses with fixed mean and conditional covariance
\(\Sigma_t\), the trace-free bound
\(\widetilde O(d+\sqrt{d\sum_t\tr\Sigma_t}+\sqrt T)\) (improving to
\(\widetilde O(d+\sqrt T)\) when \(\tr\Sigma_t=\widetilde O(1/d)\)).

\paragraph{Outline.}
Sections~\ref{sec:setting} and~\ref{sec:algorithm} present the model and
algorithm. Section~\ref{sec:regret-analysis} proves the worst-case rate
and the contractive martingale bound, and Section~\ref{sec:main-result}
establishes the upper and lower bounds in the loss quadratic form together
with the aligned-loss separation; Appendix~\ref{sec:consequences} records
the cancellation and random-loss consequences.

\subsection{Related work}
\label{sec:related-work}
Table~\ref{tab:closest-adversarial} compares rates and per-round
computation on the ball; Appendix~\ref{sec:extended-consequences} discusses the sources of the various dimensional factors in more detail, and derives
the specialisation on \(\ball\) for \citet[Theorem~3.1]{lee2020bias} and
\citet[Theorem~4]{zimmert2022bias}.

\suppressfloats[t]
\begin{table}[htbp]
\caption{Bounds on the Euclidean ball with exact scalar feedback, up to
logarithmic factors. The first row is optimal for \(T\ge d^3\)
\citep[Theorem~5.2]{jacobsen2026perturbation}. ``Not supplied'': no
efficient implementation is given.}
\label{tab:closest-adversarial}
\small
\centering
\setlength{\tabcolsep}{4pt}
\renewcommand{\arraystretch}{1.22}
\begin{tabular}{@{}>{\raggedright\arraybackslash\hangindent=1em\hangafter=1}p{0.4\linewidth}lll@{}}
\toprule
Reference & \(\widetilde O(\cdot)\) rate & Horizon & Work per round\\
\midrule
\multicolumn{4}{@{}l}{\emph{Optimal rate on the ball: oblivious adversary, expected regret}}\\
\citet[Theorem~6]{bubeck2012minimax} & \(\sqrt{dT}\) & tuned & \(O(d)\)\\
\midrule
\multicolumn{4}{@{}l}{\emph{Adaptive adversary, high-probability realised regret}}\\
\citet{lee2020bias} & \(d^{5/2}\sqrt T+d^3\) & tuned & efficient\\[1pt]
\citet{zimmert2022bias}, entropic barrier, general convex sets & \(d^2\sqrt T\) & tuned & polynomial\\[1pt]
\citet{abernethy2009beating} & \(d^{3/2}\sqrt T\) & tuned & \(O(d)\)\\[1pt]
\citet[Theorem~4]{zimmert2022bias}, \(O(1)\) ball barrier & \(d\sqrt T\) & tuned & efficient\\[1pt]
\citet{zimmert2022bias}, log-determinant, general convex sets & \(d\sqrt T\) & tuned & not supplied\\[1pt]
\textbf{Theorem~\ref{thm:worst-case}} & \(\sqrt{dT}\) & anytime & \(O(d)\)\\
\bottomrule
\end{tabular}
\end{table}

\citet{AuerCFS02} first established high-probability guarantees against adaptive adversaries for multi-armed bandits; the adaptation to linear bandits was developed by \citet{bartlett2008high} and \citet{abernethy2009beating}.
\citet[Theorem~11]{kerdreux2021linear} extend the radial barrier to smooth,
centrally symmetric, strongly convex sets, with horizon-tuned
pseudo-regret \(\widetilde O(\sqrt{dT})\).
\citet[Theorem~4.3]{jacobsen2026perturbation}
give high-probability guarantees in unconstrained linear bandits for a
comparator chosen before play. Concurrently, \citet{di2026corruption} prove high-probability
\(\widetilde O(\sqrt{dT\log K}+dC)\) regret against adaptive adversaries
with \(K\) available action vectors and corruption level \(C\); a discretised ball gives
\(\widetilde O(d\sqrt T)\) with exponentially many arms.

Appendix~\ref{sec:relation} compares our trigger test and contraction credit with the arguments of
\citet{lee2020bias}, \citet{zimmert2022bias}, and \citet{abernethy2009beating}.
\citet[Theorem~5]{zimmert2022bias} use a fixed-point bias with lifted actions
\((a,1)\), and \citet{fokkema2024online} use negative quadratic bonuses similarly in the convex bandit setting.
Implicit exploration controls rare observations by enlarging estimator
denominators \citep{kocak2014efficient,neu2015explore}. QIX uses a
quadratic denominator and tunes its scale through observed squared local
norms.

The variation and first-order bounds of
\citet{hazan2011better,ito2020tight,ito2023variance} use different
statistics and hold in expectation. The adaptive high-probability
small-loss bound of \citet[Theorem~3.1]{lee2020bias}, applied to
\(\ell_t(x)+1\) on \(\ball\times\{1\}\), charges
\(\sum_t(1-\twonorm{\theta_t})+\Delta_T\), which equals \(\Delta_T\)
only for unit-norm losses. The cancellation slack is the linear instance of the constraint-aware small
loss of \citet[Section~4.3]{gao2026small}.

Much of our guiding logic is derived from observations taken from full information, where the curvature of the action set affords coupling information between comparators, and can be used to achieve fast rates \citep{huang2017following,tsuchiya2024fast,kerdreux2021projectionfree, mayo2026smallloss}.

\section{Problem setting}
\label{sec:setting}

Let \(d\ge2\) and \(\delta\in(0,1)\). On round \(t=1,2,\ldots\), the adversary chooses
\(\theta_t\in\ball\). The learner draws fresh randomness, plays
\(a_t\in\ball\), and observes only \(Y_t=\iprod{a_t}{\theta_t}\in[-1,1]\).
The adversary is adaptive and nonanticipating, so the losses are \emph{predictable}, in the sense that \(\theta_t\) is \(\mathcal F_{t-1}\)-measurable,
where \(\mathcal F_t\) is generated by the adversary's randomisation and the
learner's draws up to round \(t\)\footnote{Appendix~\ref{sec:calibration} specifies the probability space and measurability assumptions more precisely.}. For conditional expectation, we use the shorthand \(\E_t[\cdot]=\E[\cdot\mid\mathcal F_{t-1}]\).

For \(T\ge1\), write \(S_T=\sum_{t=1}^T\theta_t\). The best hindsight
comparator is \(u^\star=-S_T/\twonorm{S_T}\) when \(S_T\ne0\), and an arbitrary unit vector otherwise. The realised regret is
\begin{equation}
 \Reg_T=\sum_{t=1}^T\iprod{a_t-u^\star}{\theta_t}
 =\sum_{t=1}^TY_t+\twonorm{S_T}.
 \label{eq:ball-realised-regret}
\end{equation}

\paragraph{Notation.} Write \(I_d\) for the identity matrix, \(\normt{M}{x}^2=x\transpose Mx\) for
\(M\succeq0\), \([x]_+=\max\{x,0\}\), and \(a\vee b=\max\{a,b\}\). Unmarked norms are Euclidean, logarithms are natural, and \(\preceq\) is the Loewner order. The symbols \(O,\Omega,\asymp\) and \(\widetilde O,\widetilde\Omega,\widetilde\Theta\) hide absolute constants; the tilde forms also hide factors polylogarithmic in \(T\) and \(1/\delta\). A guarantee is \emph{anytime} if it holds simultaneously for all \(T\ge1\) and the algorithm does not use \(T\).

\section{Adaptive quadratic implicit exploration}
\label{sec:algorithm}

At the start of round \(t\), the state \(\widehat\Theta_{t-1}\) holds the accumulated
loss estimate. Given this state, the scale \(\beta_t\) sets the exploration
probability. A separate test decides whether to shorten the state before
adding the next estimate. Algorithm~\ref{alg:radial-qix} takes only \(d\ge2\) and \(\delta\in(0,1)\),
using the epochwise schedules \((\ell_j,a_j,\kappa_j)\) introduced in
Section~\ref{sec:estimating-and-scaling} and summarised in
Table~\ref{tab:tuning} of Appendix~\ref{sec:public-tuning}.

\subsection{Sampling from the barrier geometry}

With the radial barrier \(F(z)=-\twonorm{z}-\log(1-\twonorm{z})\) and scale
\(\beta_t>0\), FTRL sets the mean action to
\begin{equation}
 z_t=\argmin_{\twonorm{z}<1}
 \left\{\iprod{\widehat\Theta_{t-1}}{z}+\beta_tF(z)\right\}.
 \label{eq:main-ftrl-leader}
\end{equation}
Let the leader direction be \(e_t=-\widehat\Theta_{t-1}/\twonorm{\widehat\Theta_{t-1}}\) when \(\widehat\Theta_{t-1}\ne0\), and a fixed unit vector otherwise. Differentiating \eqref{eq:main-ftrl-leader} gives \(z_t=(1-p_t)e_t\), where
\(p_t=\beta_t/(\twonorm{\widehat\Theta_{t-1}}+\beta_t)\) is the mean action's distance from the boundary and the exploration probability (Figure~\ref{fig:qix-geometry}a). Increasing \(\beta_t\) at a fixed
nonzero state moves the mean action inward and increases exploration.
At \(z_t=(1-p_t)e_t\), the barrier Hessian \(\nabla^2F(z_t)\) has
eigenvalue \(p_t^{-2}\) along \(e_t\) and \(p_t^{-1}\) perpendicular to it
(Proposition~\ref{prop:qix-boundary-release}(a)), so its inverse factors as
\(\{\nabla^2F(z_t)\}^{-1}=p_tG_t\), where the metric
\(G_t=I_d-\twonorm{z_t}e_te_t\transpose=I_d-(1-p_t)e_te_t\transpose\) has
weight \(p_t\) along \(e_t\) and one perpendicular to it.
The learner draws \(Z_t\sim\mathrm{Bernoulli}(p_t)\) and
\(U_t\sim\mathrm{Unif}(\sphere^{d-1})\) independently, and plays
\(a_t=(1-Z_t)e_t+Z_tG_t^{1/2}U_t\). These actions lie in \(\ball\)
and satisfy
\begin{equation}
 \E_ta_t=(1-p_t)e_t,\qquad
 d\,\E_t[Z_ta_ta_t\transpose]=p_tG_t=\{\nabla^2F(z_t)\}^{-1}.
 \label{eq:experiment-moment}
\end{equation}
Thus \(d\) times the exploratory second moment equals the inverse barrier
Hessian at \(z_t\): the exploration is matched to the curvature of the
barrier.

The loss quadratic form in this metric is
\begin{equation}
 \normt{G_t}{\theta_t}^2
 =\twonorm{\theta_t}^2-\twonorm{z_t}\iprod{e_t}{\theta_t}^2
 =\twonorm{\theta_t}^2-(1-p_t)\iprod{e_t}{\theta_t}^2\in[0,1].
 \label{eq:q-clock}
\end{equation}
It subtracts the mean action's radius \(\twonorm{z_t}=1-p_t\) times the
squared radial projection from the full squared norm; equivalently,
it weights tangential loss fully and radial loss by \(p_t\)
(Figure~\ref{fig:qix-geometry}b). Its cumulative value
\(\sum_{t\le T}{(\twonorm{\theta_t}^2-\twonorm{z_t}\iprod{e_t}{\theta_t}^2)}\)
depends on the learner's trajectory; each summand is predictable but
unobserved.
With coordinate exploration, the stability term of a round has conditional mean
\(d\twonorm{\theta_t}^2\) per unit learning rate \citep[proof of Theorem~6]{bubeck2012minimax}; with curvature-matched exploration it is of order at most
\(d\normt{G_t}{\theta_t}^2\) (Lemma~\ref{lem:qix:sandwich}).

\subsection{Estimating losses and setting the scale}
\label{sec:estimating-and-scaling}
The learner cannot compute \(\normt{G_t}{\theta_t}^2\) from its single
observation \(Y_t=\iprod{a_t}{\theta_t}\). To tune the scale, it uses
the squared size of a computable loss estimate. With tuning index
\(j_t=\floor{\log_2(\beta_{t-1}/\beta_0)}\) and
\(\nu_t=d^2\ell_{j_t}^2/(4\beta_t^2)\), the QIX estimate is
\begin{equation}
 \widehat{\theta}_t=(1-Z_t)e_tY_t
 +\frac{dZ_tY_t}{p_t+\nu_tY_t^2}G_t^{1/2}U_t.
 \label{eq:qix-estimator}
\end{equation}
On an exploration round, weighting by \(1/p_t\) alone would amplify the
observed loss when exploration is rare. The extra term \(\nu_tY_t^2\)
limits this amplification to quadratic order. Interestingly, the resulting bias has the direction
\(G_t\theta_t\), as Lemma~\ref{lem:qix:bias} shows.

The \emph{observed squares}
\(\EestObs_\tau=\sum_{t\le\tau}p_t\normt{G_t}{\widehat\theta_t}^2
=\sum_{t\le\tau}\normt{\{\nabla^2F(z_t)\}^{-1}}{\widehat\theta_t}^2\) tune the scale:
\[
 \beta_t=\max\bigl\{\beta_{t-1},\;a_{j_t}\sqrt{\EestObs_{t-1}+1},\;
 \rho_{\kappa_{j_t}}(\twonorm{\widehat\Theta_{t-1}})\bigr\}.
\]
In the second argument, the decreasing coefficient \(a_{j_t}\) both splits the confidence level over dyadic scale intervals indexed by \(j_t\) and balances barrier regularisation against stability by shrinking
as \((j_t+1)^{-1/2}\).

The third term keeps exploration frequent enough for the observed squares
to track the losses and controls QIX bias. For one-step stability, we require
\(\beta_t^2p_t\ge\kappa_{j_t}\); substituting
\(p_t=\beta_t/(\twonorm{\widehat\Theta_{t-1}}+\beta_t)\) gives the cubic
inequality \(\beta_t^3\ge\kappa_{j_t}(\beta_t+\twonorm{\widehat\Theta_{t-1}})\).
For \(\kappa>0\) and \(h\ge0\), the closed-form cubic floor
\begin{equation}
 \rho_\kappa(h):=\bigl(\kappa^{3/2}+2\kappa h\bigr)^{1/3}
 \label{eq:overview-rho-definition}
\end{equation}
meets this requirement (since \((\sqrt\kappa+h)^3\ge\kappa^{3/2}+2\kappa h\)
gives \(\rho_\kappa(h)^3\ge\kappa\{\rho_\kappa(h)+h\}\); Lemma~\ref{lem:cubic-support})
and anchors \(\rho_\kappa(0)=\sqrt\kappa\). With this floor, even \emph{after} shrinkage, the
conditional mean of each observed square lies between
\(\tfrac d8\normt{G_t}{\theta_t}^2\) and \(d\normt{G_t}{\theta_t}^2\) (Lemma~\ref{lem:qix:sandwich}),
allowing a scale tuned from scalar feedback to track the unobserved loss
quadratic form.
Concretely, with \(\delta_0=\delta/8\), epoch \(j\ge0\) uses
\(\ell_j=\log\{\pi^2(j+1)^2/(6\delta_0)\}\),
\(a_j=\sqrt{\ell_j/\{48(j+1)\}}\), and
\(\kappa_j=\max\{1,(j+1)/d\}(\ell_j/\ell_0)^2\kappa_0\asymp\max\{d,j+1\}\ell_j^2\),
anchored at \(\beta_0=\sqrt{\kappa_0}\) with
\(\kappa_0=\ell_0^2\cdot12d^2(d+3)/\{(d+4)(7d+6)\}\) (Table~\ref{tab:tuning}).

Write \(\xi_t=\widehat\theta_t-\E_t\widehat\theta_t\) for the estimation noise; the moments of QIX enjoy the following useful properties.

\begin{lemma}[QIX moments]
\label{lem:qix}
There is a predictable \(\lambda_t\in[0,1]\) such that, almost surely,
\begin{lemenum}
\item \(\theta_t-\E_t\widehat{\theta}_t=\lambda_tG_t\theta_t\);
 \label{lem:qix:bias}
\item \(\operatorname{Cov}_t(\widehat{\theta}_t)\preceq5\normt{G_t}{\theta_t}^2G_t/p_t\);
 \label{lem:qix:cov}
\item \(\normt{p_tG_t^{-1}}{\widehat{\theta}_t}\le\beta_t/\ell_{j_t}\) and
 \(\normt{p_tG_t^{-1}}{\xi_t}\le\beta_t/\ell_{j_t}+1\);
 \label{lem:qix:range}
\item \(\normt{\{\nabla^2F(z_t)\}^{-1}}{\widehat{\theta}_t}=\sqrt{p_t}\normt{G_t}{\widehat\theta_t}\le\beta_t/\ell_{j_t}\);
 \label{lem:qix:local}
\item \(\tfrac d8\normt{G_t}{\theta_t}^2\le\E_t[p_t\normt{G_t}{\widehat\theta_t}^2]\le d\normt{G_t}{\theta_t}^2\).
 \label{lem:qix:sandwich}
\end{lemenum}
\end{lemma}
Appendix~\ref{sec:qix-moments} gives \(\lambda_t\) explicitly and proves
the lemma. Part~(b) controls the variance of a loss comparison in any
direction, while parts~(c) and~(d) bound individual updates even on rare
exploration rounds.

\subsection{Choosing when to contract}
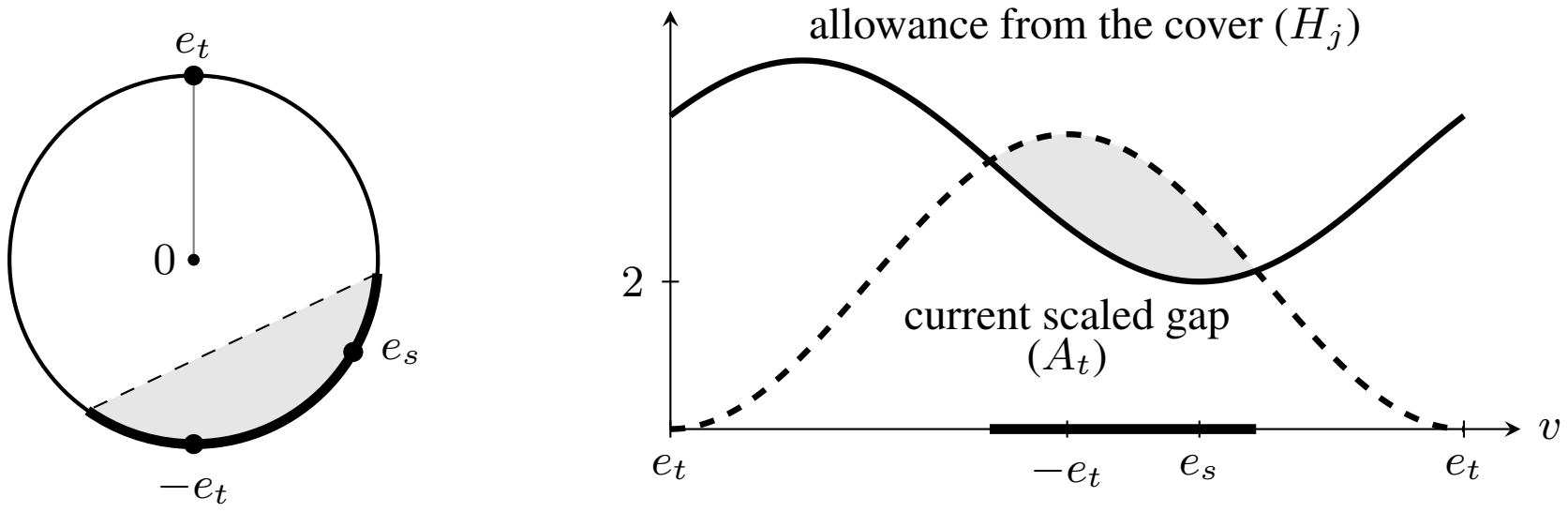
\begin{figure}[ht]
\centering
\begin{tikzpicture}[>=stealth,font=\small,baseline=(testa.base)]
 \fill[gray!20] (-4.27:1.3) arc (-4.27:-125.17:1.3) -- cycle;
 \draw[thick] (0,0) circle (1.3);
 \draw[line width=2pt] (-4.27:1.3) arc (-4.27:-125.17:1.3);
 \draw[dashed] (-4.27:1.3) -- (-125.17:1.3);
 \draw[gray] (0,0) -- (90:1.3);
 \fill (0,0) circle (1.2pt) node[left] {\footnotesize \(0\)};
 \fill (90:1.3) circle (2pt) node[above] {\(e_t\)};
 \fill (-30:1.3) circle (2pt) node[right=2pt] {\(e_s\)};
 \fill (-90:1.3) circle (2pt) node[below=2pt] {\(-e_t\)};
 \node (testa) at (0.1,-2.15) {(a) the cap where the test fails};
\end{tikzpicture}\hspace{0.7cm}
\begin{tikzpicture}[>=stealth,font=\small,baseline=(testb.base)]
 \fill[gray!20]
  plot[domain=-125.17:-4.27,samples=40,variable=\x]
   ({(\x+270)/360*5.6},{0.52*2*(1-sin(\x))})
  -- plot[domain=-4.27:-125.17,samples=40,variable=\x]
   ({(\x+270)/360*5.6},{0.52*(2+1.5*(1-cos(\x+30)))}) -- cycle;
 \draw[->] (0,0) -- (6.0,0) node[right] {\(v\)};
 \draw[->] (0,0) -- (0,2.95);
 \draw[line width=2pt] ({(-125.17+270)/360*5.6},0) -- ({(-4.27+270)/360*5.6},0);
 \draw[very thick,domain=-270:90,samples=90,smooth,variable=\x]
  plot ({(\x+270)/360*5.6},{0.52*(2+1.5*(1-cos(\x+30)))});
 \draw[dashed,very thick,domain=-270:90,samples=90,smooth,variable=\x]
  plot ({(\x+270)/360*5.6},{0.52*2*(1-sin(\x))});
 \foreach \a/\l in {-270/e_t,-90/-e_t,-30/e_s,90/e_t}
  \draw ({(\a+270)/360*5.6},0.06) -- ({(\a+270)/360*5.6},-0.06)
   node[below] {\footnotesize \(\l\)};
 \draw (0.06,1.04) -- (-0.06,1.04) node[left] {\footnotesize \(2\)};
 \node[anchor=south west,inner sep=2pt] at (0.9,2.60)
   {\footnotesize allowance from the cover (\(H_j\))};
 \node[align=center,inner sep=1pt] at (2.8,0.64)
   {\footnotesize current scaled gap\\[-2pt]\footnotesize (\(A_t\))};
 \node (testb) at (2.9,-0.95) {(b) the two quadratic forms at \((-v,1)\)};
\end{tikzpicture}
\caption{The matrix test for \(d=2\), with one earlier trigger at \(e_s\).
The thick arc in (a) marks the directions \(v\) for which the current
scaled gap exceeds the allowance. These directions form the shaded interval in (b),
where the dashed curve rises above the solid curve. If the interval is
nonempty, the learner contracts the state and adds the dashed curve to
the cover.}
\label{fig:trigger-test}
\end{figure}

The scale controls the size of estimates, but regret against the hindsight
comparator requires a bound valid for every comparator at once.
Before adding \(\widehat{\theta}_t\), the learner occasionally contracts the previous state:
\(\widehat\Theta_t=(1-\epsilon_t)\widehat\Theta_{t-1}+\widehat{\theta}_t\), with \(\epsilon_t\in\{0,1/4\}\).
This shortens the state without changing its leader
direction; at a fixed scale, it increases exploration.
To decide when to contract, we compare each unit comparator's present gap \(\twonorm{\widehat\Theta_{t-1}}\{1-\iprod v{e_t}\}\) with its gaps
on earlier contraction rounds in the same dyadic scale epoch, measuring
each in units of that round's scale. The \emph{cover} gives each
comparator an initial allowance of two, plus the sum of those earlier
scaled gaps. We contract as soon as any current scaled gap exceeds its
allowance (Figure~\ref{fig:trigger-test}). Concretely, before the first contraction of an epoch, the
largest scaled gap is attained at \(v=-e_t\) and equals
\(2\twonorm{\widehat\Theta_{t-1}}/\beta_t\). The test then triggers precisely when
\(\twonorm{\widehat\Theta_{t-1}}>\beta_t\). Lines~7--11 of Algorithm~\ref{alg:radial-qix} perform this
test using one scalar and one vector. We derive the test below.

\FloatBarrier
\begin{algorithm}[ht]
\begin{algbox}{Adaptive radial QIX}{alg:radial-qix}
\Require \(d\ge2\), \(\delta\in(0,1)\) (with schedules
\(\ell_j,a_j,\kappa_j\) from Section~\ref{sec:algorithm} and
Table~\ref{tab:tuning})
\State Fix \(e_0\in\sphere^{d-1}\); set \(\widehat\Theta_0\gets0\), \(\EestObs_0\gets0\), and
       \(\beta_0\gets\sqrt{\kappa_0}\)
\For{\(t=1,2,\ldots\)}
 \State \(h_t\gets\twonorm{\widehat\Theta_{t-1}}\) and
        \(j_t\gets\floor{\log_2(\beta_{t-1}/\beta_0)}\)
        \Comment{scale}
 \State \(\beta_t\gets\max\{\beta_{t-1},a_{j_t}\sqrt{\EestObs_{t-1}+1},
        \rho_{\kappa_{j_t}}(h_t)\}\)
 \State \(e_t\gets-\widehat\Theta_{t-1}/h_t\) if \(h_t>0\), and \(e_t\gets e_0\) otherwise
 \State \(p_t\gets\beta_t/(h_t+\beta_t)\), \(z_t\gets(1-p_t)e_t\), and
        \(G_t\gets I_d-(1-p_t)e_te_t\transpose\)
        \algtag{eq:natural-depth}
 \State \(j\gets\floor{\log_2(\beta_t/\beta_0)}\); on entering a new epoch set
        \((\alpha_j,b_j)\gets(1,0)\) and discard the previous epoch
        \Comment{cover}
 \State \(r_t\gets h_t/\beta_t\), \(\epsilon_t\gets0\)
 \If{\(r_t\{\alpha_j-\iprod{b_j}{e_t}\}>\alpha_j^2-\twonorm{b_j}^2\)}
 \State \(\epsilon_t\gets1/4\),
        \(\alpha_j\gets\alpha_j+r_t/2\),
        \(b_j\gets b_j+r_te_t/2\)
 \EndIf
 \State Draw \(Z_t\sim\mathrm{Bernoulli}(p_t)\) and
        \(U_t\sim\mathrm{Unif}(\sphere^{d-1})\) independently
        \Comment{sampling}
 \State Play \(a_t\gets(1-Z_t)e_t+Z_tG_t^{1/2}U_t\), observe
        \(Y_t=\iprod{a_t}{\theta_t}\), and form \(\widehat{\theta}_t\) by
        \eqref{eq:qix-estimator} with \(\nu_t=d^2\ell_{j_t}^2/(4\beta_t^2)\)
 \State \(\widehat\Theta_t\gets(1-\epsilon_t)\widehat\Theta_{t-1}+\widehat{\theta}_t\) and
        \(\EestObs_t\gets\EestObs_{t-1}+p_t\normt{G_t}{\widehat\theta_t}^2\)
        \algtag{eq:state-update}
\EndFor
\end{algbox}
\end{algorithm}

The epochs are \(I_j=\{t:B_j\le\beta_t<2B_j\}\) with
\(B_j=2^j\beta_0\) for \(j\ge0\) (Table~\ref{tab:tuning}), so
\(j=\floor{\log_2(\beta_t/\beta_0)}\) throughout \(I_j\). The cover is built from the earlier
contraction rounds of the current epoch. With \(r_t=\twonorm{\widehat\Theta_{t-1}}/\beta_t\), the round's
matrix and the cover are
\[
 A_t=\frac{r_t}2
 \begin{pmatrix}I_d&e_t\\e_t\transpose&1\end{pmatrix},
 \qquad
 H_j=I_{d+1}+\sum_{s\in I_j,\,s<t,\,\epsilon_s>0}A_s
 =\begin{pmatrix}\alpha_jI_d&b_j\\b_j\transpose&\alpha_j\end{pmatrix}.
\]

Before each draw, the test checks whether
\(A_t\preceq H_j\), which holds exactly
when every unit \(v\)'s scaled gap \(r_t\{1-\iprod v{e_t}\}\) is at most
its cover allowance \(\bar v\transpose H_j\bar v\) at
\(\bar v=(-v,1)\). A failed test, a \emph{trigger}, adds
\(A_t\) to the cover and contracts the state; the cover
restarts at \(I_{d+1}\) whenever the scale enters a new epoch.
This adapts \citet[Algorithm~2]{lee2020bias}, who raise the learning rate
on a trigger. Lemma~\ref{lem:trigger} shows that Algorithm~\ref{alg:radial-qix}'s
scalar test is equivalent to \(A_t\preceq H_j\)
and bounds the number of triggers.

\subsection{Computation}
Storing and updating the cover needs only the pair \((\alpha_j,b_j)\). Since
\(G_t^{1/2}u=u+(\sqrt{p_t}-1)\iprod{e_t}{u}e_t\), sampling and
estimation require only vector operations. Once completed epochs are
discarded, a round takes one scalar cube root for
\(\rho_{\kappa_{j_t}}(\twonorm{\widehat\Theta_{t-1}})\) and \(O(d)\) operations and memory.

\Needspace{12\baselineskip}
\section{Regret analysis}
\label{sec:regret-analysis}
To bound the realised regret against the best action in hindsight, we
cannot fix a comparator before play: an adaptive adversary can choose
later losses in response to the learner's draws, making \(u^\star\)
random and dependent on the entire game. We therefore establish a regret
bound that holds simultaneously for every comparator \(v\in\ball\) on a
single high-probability event. Rather than bounding everything at once,
we first reduce the regret to two quantities produced by the algorithm:
the final regularisation scale \(\beta_T\) and the cumulative loss
quadratic form \(\sum_{t=1}^T\normt{G_t}{\theta_t}^2\), where
\(\normt{G_t}{\theta_t}^2=\twonorm{\theta_t}^2-\twonorm{z_t}\iprod{e_t}{\theta_t}^2\)
by \eqref{eq:q-clock}.

\begin{proposition}[Comparator-uniform regret reduction]
\label{prop:main-reduction} With probability
at least \(1-\delta\), simultaneously for all \(T\ge1\) and \(v\in\ball\),
Algorithm~\ref{alg:radial-qix} satisfies
\[
 \sum_{t=1}^T\iprod{a_t-v}{\theta_t}
 \le\widetilde O\!\left(\beta_T
   +\sqrt{d\sum_{t=1}^T \left(\twonorm{\theta_t}^2
                             -\twonorm{z_t}\iprod{e_t}{\theta_t}^2\right)}\right).
\]
\end{proposition}

Because the regret \(\sum_{t=1}^T\iprod{a_t-v}{\theta_t}\) is affine in
\(v\) and \(\ball\) is the convex hull of the unit sphere \(\sphere^{d-1}\),
it suffices to prove the bound simultaneously for every unit vector
\(v\in\sphere^{d-1}\). Below we fix a horizon \(T\ge1\) and walk through
the proof in seven steps, keeping every step uniform over
\(v\in\sphere^{d-1}\); Appendices~\ref{sec:local}--\ref{sec:assembly}
give explicit finite-sample constants and extend the guarantee to
all horizons simultaneously.

\subsection{Proof outline}
\label{sec:proof-strategy}

\qixproofstage{1}{Decompose the realised regret}
Algorithm~\ref{alg:radial-qix} maintains the mean action \(z_t=(1-p_t)e_t\),
plays a random action \(a_t\) with \(\E_t a_t=z_t\), and updates its state
using the loss estimate \(\widehat\theta_t\). Adding and subtracting both
\(z_t\) and \(\widehat\theta_t\) yields the pathwise identity
\begin{align}
 \sum_{t=1}^T\iprod{a_t-v}{\theta_t}
 =\underbrace{\sum_{t=1}^T\iprod{z_t-v}{\widehat{\theta}_t}}
             _{\text{estimated-loss regret}}
 +\underbrace{\sum_{t=1}^T\iprod{z_t-v}{\theta_t-\widehat{\theta}_t}}
             _{\text{estimation error}}
 +\underbrace{\sum_{t=1}^T\iprod{a_t-z_t}{\theta_t}}
             _{\text{sampling noise}}.
 \label{eq:regret-decomposition}
\end{align}

Let us inspect the three terms in \eqref{eq:regret-decomposition}.
The \emph{sampling noise} is the easiest because it does not involve the
comparator \(v\) at all. Each increment \(\iprod{a_t-z_t}{\theta_t}\) is a
centred martingale difference bounded in absolute value by \(\sqrt2\) with
conditional variance
\[
 \operatorname{Var}_t\bigl(\iprod{a_t-z_t}{\theta_t}\bigr)
 =p_t(1-p_t)\iprod{e_t}{\theta_t}^2+\frac{p_t}{d}\normt{G_t}{\theta_t}^2
 \le\normt{G_t}{\theta_t}^2,
\]
where the inequality uses \(p_t\iprod{e_t}{\theta_t}^2\le\normt{G_t}{\theta_t}^2\)
and \(d\ge2\). Freedman's inequality (Lemma~\ref{lem:played} in
Appendix~\ref{sec:sampling-error}) therefore bounds the sampling noise on
one high-probability event by
\(\widetilde O\bigl(1+\sqrt{\sum_{t=1}^T\normt{G_t}{\theta_t}^2}\bigr)\).

The \emph{estimated-loss regret} is the regret of the FTRL mean actions
\(z_t\) on the loss estimates \(\widehat\theta_t\), complicated only by the
occasional state contractions. The \emph{estimation error} is the delicate
term: it depends on the hindsight comparator \(v\), so a separate
martingale bound for each \(v\) would require a covering-net union bound
and lose a factor \(\sqrt d\). As we show next, the state contractions inside
the estimated-loss regret generate a comparator-dependent negative
term (which we refer to henceforth as the \emph{contraction credit}) that we will pair directly with the
estimation error.

\qixproofstage{2}{Apply FTRL to the effective losses and extract the contraction credit}
Because Algorithm~\ref{alg:radial-qix} updates its state by
\(\widehat\Theta_t=(1-\epsilon_t)\widehat\Theta_{t-1}+\widehat\theta_t\),
the \emph{effective loss} added to the FTRL state in round \(t\) is
\[
 g_t=\widehat\Theta_t-\widehat\Theta_{t-1}
    =\widehat\theta_t-\epsilon_t\widehat\Theta_{t-1}.
\]
By Fenchel duality, the FTRL mean action \eqref{eq:main-ftrl-leader} is
\(z_t=\nabla\Psi_{\beta_t}(-\widehat\Theta_{t-1})\), where
\(\Psi_\beta(y)=(\beta F)^*(y)=\twonorm y-\beta\log(1+\twonorm y/\beta)\)
is the dual potential. The one-round stability cost is the Bregman
divergence between consecutive states,
\[
 \mathrm{stab}_t
 =D_{\Psi_{\beta_t}}(-\widehat\Theta_t,-\widehat\Theta_{t-1})
 =\Psi_{\beta_t}(-\widehat\Theta_t)-\Psi_{\beta_t}(-\widehat\Theta_{t-1})
  +\iprod{z_t}{g_t}.
\]
For a unit comparator \(v\), define its \emph{gap} after round \(t\) by
\[
 \Gamma_t(v)=\twonorm{\widehat\Theta_t}+\iprod{v}{\widehat\Theta_t}\ge0.
\]
Since \(\widehat\Theta_{t-1}=-\twonorm{\widehat\Theta_{t-1}}e_t\), we have
\(\Gamma_{t-1}(v)=\twonorm{\widehat\Theta_{t-1}}\{1-\iprod{v}{e_t}\}
 =\tfrac12\twonorm{\widehat\Theta_{t-1}}\twonorm{v-e_t}^2\):
the gap is \(v\)'s excess estimated loss above the current leader \(e_t\),
and on the sphere it is proportional to the squared chord length
\(\twonorm{v-e_t}^2\) (Figure~\ref{fig:cover-main}a).

\begin{lemma}[Radial FTRL under state contractions]
\label{lem:contracted-ftrl}
For every unit \(v\),
\begin{equation}
 \sum_{t=1}^T\iprod{z_t-v}{\widehat{\theta}_t}
 =\sum_{t=1}^T\beta_t\log
 \frac{\beta_t+\twonorm{\widehat\Theta_t}}{\beta_t+\twonorm{\widehat\Theta_{t-1}}}
 +\sum_{t=1}^T\mathrm{stab}_t
 -\Gamma_T(v)-\sum_{t=1}^T\epsilon_t\Gamma_{t-1}(v)
 +\sum_{t=1}^T\epsilon_t\beta_t(1-p_t).
 \label{eq:contracted-ftrl}
\end{equation}
If \(\beta_t\) is nondecreasing, the first sum is at most
\(\beta_T\log(1+\twonorm{\widehat\Theta_T}/\beta_T)\).
\end{lemma}

The identity \eqref{eq:contracted-ftrl} comes from two elementary
calculations (detailed in Appendix~\ref{sec:decomposition}). First,
applying the standard FTRL potential argument \citep[Lemma~2]{ALST14} to
the effective losses \(g_t\) gives
\[
 \iprod{z_t}{g_t}
 =\Psi_{\beta_t}(\widehat\Theta_{t-1})-\Psi_{\beta_t}(\widehat\Theta_t)+\mathrm{stab}_t
 =\twonorm{\widehat\Theta_{t-1}}-\twonorm{\widehat\Theta_t}
  +\beta_t\log\frac{\beta_t+\twonorm{\widehat\Theta_t}}{\beta_t+\twonorm{\widehat\Theta_{t-1}}}
  +\mathrm{stab}_t,
\]
using the evenness of \(\Psi_\beta\). Summing from \(\widehat\Theta_0=0\)
telescopes the Euclidean norms to \(-\twonorm{\widehat\Theta_T}\), while
\(-\sum_{t=1}^T\iprod{v}{g_t}=-\iprod{v}{\widehat\Theta_T}\); together these
yield \(-\Gamma_T(v)\) (Lemma~\ref{lem:effective-ftrl}). Second, passing
from \(g_t\) back to \(\widehat\theta_t=g_t+\epsilon_t\widehat\Theta_{t-1}\)
adds \(\epsilon_t\iprod{z_t-v}{\widehat\Theta_{t-1}}\) on each round (the
linear bias term of \citealp[Theorem~1]{zimmert2022bias}, with bias
\(\epsilon_t\widehat\Theta_{t-1}=-\epsilon_t\beta_t\nabla F(z_t)\)).
Using \(\widehat\Theta_{t-1}=-\twonorm{\widehat\Theta_{t-1}}e_t\),
\(z_t=(1-p_t)e_t\), and \(p_t\twonorm{\widehat\Theta_{t-1}}=\beta_t(1-p_t)\),
this contribution splits cleanly into a negative gap term and a
comparator-independent cost:
\begin{align*}
 \epsilon_t\iprod{z_t-v}{\widehat\Theta_{t-1}}
 &\overset{\textup{(a)}}{=}\epsilon_t\twonorm{\widehat\Theta_{t-1}}
   \bigl\{\iprod{v}{e_t}-(1-p_t)\bigr\}\\
 &\overset{\textup{(b)}}{=}-\epsilon_t\twonorm{\widehat\Theta_{t-1}}
   \bigl\{1-\iprod{v}{e_t}\bigr\}+\epsilon_t p_t\twonorm{\widehat\Theta_{t-1}}\\
 &\overset{\textup{(c)}}{=}-\epsilon_t\Gamma_{t-1}(v)+\epsilon_t\beta_t(1-p_t).
\end{align*}
Here (a) substitutes \(z_t\) and \(\widehat\Theta_{t-1}\); (b) regroups
the radial term; and (c) applies the definitions of \(\Gamma_{t-1}(v)\) and
\(p_t\). Summing over \(t\) gives \eqref{eq:contracted-ftrl}.

Let us pause to inspect each of the five terms on the right-hand side of
\eqref{eq:contracted-ftrl}:
\begin{itemize}[leftmargin=1.5em,itemsep=2pt,topsep=2pt]
\item \emph{Terminal gap.} We have \(-\Gamma_T(v)\le0\) and simply drop it.
\item \emph{Regularisation cost.} Since \(\beta_t\) is nondecreasing, the
logarithmic sum is at most \(\beta_T\log(1+\twonorm{\widehat\Theta_T}/\beta_T)\).
Because the cubic floor keeps \(\twonorm{\widehat\Theta_T}\) polynomial in
\(\beta_T\) (Lemma~\ref{lem:terminal-cost}), this term is \(\widetilde O(\beta_T)\).
\item \emph{Stability.} Self-concordance of the radial barrier together with
the QIX local-norm bound \(\sqrt{p_t}\normt{G_t}{\widehat\theta_t}\le\beta_t/\ell_{j_t}\)
(Lemma~\ref{lem:qix:local}) bounds each round's stability cost by
(Lemma~\ref{lem:ftrl-stability})
\[
 \mathrm{stab}_t
 \le\frac{p_t\normt{G_t}{\widehat\theta_t}^2}{\beta_t}
 +\frac34\II{\epsilon_t>0}\beta_t(1-p_t).
\]
The first term is the usual local-norm stability, which we sum in Step~6.
\item \emph{Positive contraction cost.} Combining
\(\epsilon_t\beta_t(1-p_t)=(1/4)\II{\epsilon_t>0}\beta_t(1-p_t)\) from
\eqref{eq:contracted-ftrl} with the stability overhead
\((3/4)\II{\epsilon_t>0}\beta_t(1-p_t)\) charges at most
\(\beta_t(1-p_t)<2B_j\) per trigger in dyadic epoch
\(I_j=\{t:B_j\le\beta_t<2B_j\}\). As shown immediately below, each epoch
has fewer than \(10(j+1)\) triggers, \emph{independently of \(d\)}
(Lemma~\ref{lem:trigger}); summing over nonempty epochs
(\(\sum_{j\le j(\beta_T)}B_j<2\beta_T\)) therefore totals \(\widetilde O(\beta_T)\).
\item \emph{Contraction credit.} We retain the negative term
\(-\sum_{t=1}^T\epsilon_t\Gamma_{t-1}(v)\), which grows with \(v\)'s own
gap on trigger rounds (Figure~\ref{fig:cover-main}a), and pair it with the
estimation error in Steps~3--5.
\end{itemize}

\paragraph{The trigger count is dimension-free.} Write \(D=\alpha_j^2-\twonorm{b_j}^2\ge1\) for the cover parameters of
Algorithm~\ref{alg:radial-qix}. Whenever the scalar test
\(r_t\{\alpha_j-\iprod{b_j}{e_t}\}>D\) fires and updates
\((\alpha_j,b_j)\gets(\alpha_j+r_t/2,b_j+r_te_t/2)\), expanding the squares
cancels \(r_t^2/4\) and gives
\[
 D^+
 =\Bigl(\alpha_j+\frac{r_t}{2}\Bigr)^2-\twonorm[\Big]{b_j+\frac{r_t}{2}e_t}^2
 =D+r_t\bigl\{\alpha_j-\iprod{b_j}{e_t}\bigr\}
 >2D.
\]
Thus \(D\) more than doubles at every trigger, so after \(m\) triggers in
epoch \(j\) we have \(2^m<D_m\le\alpha_m^2\). Meanwhile, the cubic floor
\(\beta_t^2p_t\ge\kappa_0\) and \(\beta_t<2B_j=2^{j+1}\sqrt{\kappa_0}\) cap
\(r_t=\twonorm{\widehat\Theta_{t-1}}/\beta_t=1/p_t-1\le\beta_t^2/\kappa_0-1<4^{j+1}\)
throughout \(I_j\), so \(\alpha_m<1+\tfrac12m\cdot4^{j+1}\). Comparing
exponential growth \(2^{m/2}\) with linear growth in \(m\) forces
\(m<10(j+1)\).

Substituting Lemma~\ref{lem:contracted-ftrl} and the bounds above into
\eqref{eq:regret-decomposition} gives the intermediate bound
\begin{equation}
 \sum_{t=1}^T\iprod{a_t-v}{\theta_t}
 \le\widetilde O(\beta_T)
 +\sum_{t=1}^T\frac{p_t\normt{G_t}{\widehat\theta_t}^2}{\beta_t}
 +\sum_{t=1}^T\bigl\{\iprod{z_t-v}{\theta_t-\widehat\theta_t}
                    -\epsilon_t\Gamma_{t-1}(v)\bigr\}
 +\sum_{t=1}^T\iprod{a_t-z_t}{\theta_t}.
 \label{eq:step2-intermediate}
\end{equation}
The sampling noise is already controlled by Step~1. The heart of the proof
is the second sum on the right-hand side of \eqref{eq:step2-intermediate}: showing that, on a
single high-probability event simultaneously for every unit \(v\), the
contraction credit pays for the estimation error up to \(\widetilde O(\beta_T)\).

\qixproofstage{3}{Link directional noise and bias to the comparator's gap}
Recall \(\xi_t=\widehat\theta_t-\E_t\widehat\theta_t\). By
Lemma~\ref{lem:qix:bias}, \(\theta_t-\E_t\widehat\theta_t=\lambda_tG_t\theta_t\)
for a predictable \(\lambda_t\in[0,1]\), so the round-\(t\) estimation
error splits into centred noise and QIX bias:
\[
 \iprod{z_t-v}{\theta_t-\widehat\theta_t}
 =\underbrace{-\iprod{z_t-v}{\xi_t}}_{\text{estimation noise}}
 +\underbrace{\lambda_t\iprod{z_t-v}{G_t\theta_t}}_{\text{QIX bias}}.
\]
By Lemma~\ref{lem:qix:cov},
\(\operatorname{Cov}_t(\widehat\theta_t)\preceq5\normt{G_t}{\theta_t}^2G_t/p_t\),
which gives
\(\E_t\iprod{z_t-v}{\xi_t}^2\le5\normt{G_t}{\theta_t}^2\normt{G_t}{z_t-v}^2/p_t\).
Likewise, Cauchy--Schwarz in the \(G_t\) metric bounds the bias by
\(\lambda_t\abs{\iprod{z_t-v}{G_t\theta_t}}
 \le(\lambda_t\sqrt{p_t}\,\normt{G_t}{\theta_t})(\normt{G_t}{z_t-v}/\sqrt{p_t})\).
Thus both noise variance and QIX bias depend on the comparator through the
same ratio \(\normt{G_t}{z_t-v}^2/p_t\). The next lemma shows that on the
Euclidean sphere, curvature-matched exploration ties this ratio directly
to the comparator's scaled gap \(\Gamma_{t-1}(v)/\beta_t\).

\begin{lemma}[Directional geometry and bias]
\label{lem:directional-geometry}
For every unit \(v\), pathwise,
\begin{equation}
 \frac{\normt{G_t}{z_t-v}^2}{p_t}\le1+2\Gamma_{t-1}(v)/\beta_t,
 \label{eq:stock-geometry}
\end{equation}
and the bias satisfies
\begin{equation}
 \lambda_t\abs{\iprod{z_t-v}{G_t\theta_t}}\le
 \frac{d\ell_{j_t}\normt{G_t}{\theta_t}^2}{\sqrt{j_t+1}\,\beta_t}\sqrt{1+2\Gamma_{t-1}(v)/\beta_t}.
 \label{eq:bias-bound}
\end{equation}
\end{lemma}

The geometric inequality \eqref{eq:stock-geometry} is a three-line sphere
identity. Write \(c=\iprod{v}{e_t}\) and \(\eta=1-c\). Resolving
\(z_t-v=(1-p_t)e_t-v\) into its tangential part (squared norm
\(1-c^2=\eta(2-\eta)\)) and its radial part (\(\eta-p_t\)) gives
\(\normt{G_t}{z_t-v}^2=\eta(2-\eta)+p_t(\eta-p_t)^2\). Since
\(r_t=\twonorm{\widehat\Theta_{t-1}}/\beta_t\) satisfies \(p_tr_t=1-p_t\),
direct expansion gives
\[
 p_t(1+2r_t\eta)-\normt{G_t}{z_t-v}^2
 =(1-p_t)\bigl\{(\eta-p_t)^2+p_t\bigr\}\ge0.
\]
Dividing by \(p_t\) yields \eqref{eq:stock-geometry} because
\(r_t\eta=\twonorm{\widehat\Theta_{t-1}}\{1-\iprod{v}{e_t}\}/\beta_t=\Gamma_{t-1}(v)/\beta_t\).
Pairing \eqref{eq:stock-geometry} with Cauchy--Schwarz and the QIX bias
bound gives \eqref{eq:bias-bound} (Appendix~\ref{sec:bandit-error}).

Let us see what \eqref{eq:stock-geometry} already buys for a \emph{fixed}
comparator \(v\). Combining it with Lemma~\ref{lem:qix:cov} yields
\[
 \E_t\iprod{z_t-v}{\xi_t}^2
 \le5\normt{G_t}{\theta_t}^2\bigl\{1+2\Gamma_{t-1}(v)/\beta_t\bigr\}.
\]
Noise variance, QIX bias, and contraction credit are now all measured in
the same currency: the scaled gap \(\Gamma_{t-1}(v)/\beta_t\). What
remains is to turn this fixed-comparator variance control into a bound
that holds \emph{simultaneously} for every unit comparator \(v\), without
paying a \(\sqrt d\) covering-net penalty.

\begin{figure}[t]
\centering
\begin{tikzpicture}[>=stealth,font=\small,scale=2.25,baseline=(capa.base)]
 \fill[gray!20] (46.05:1) arc (46.05:133.95:1) -- cycle;
 \draw[thick] (22:1) arc (22:158:1);
 \draw[gray] (0,0) -- (0,1);
 \fill (0,0) circle (0.6pt) node[below] {\footnotesize \(0\)};
 \draw[gray] (-0.78,1) -- (0.95,1);
 \draw[dashed] (-0.78,0.72) -- (0.95,0.72);
 \draw[densely dotted,thick] (0,1) -- (30:1);
 \draw[<->] (30:1) -- (0.866,1);
 \draw[<->] (-0.66,0.72) -- (-0.66,1);
 \node[left] at (-0.66,0.86) {\footnotesize \(p_t\)};
 \node[right] at (0.866,0.86) {\footnotesize depth};
 \fill (0,1) circle (0.8pt) node[below left=-1pt] {\footnotesize \(e_t\)};
 \fill (0,0.72) circle (0.8pt) node[below right=-1pt] {\footnotesize \(z_t\)};
 \fill (30:1) circle (0.8pt) node[right] {\footnotesize \(v\)};
 \node (capa) at (0.08,-0.34) {(a) the gap on the ball};
\end{tikzpicture}\hspace{0.9cm}
\begin{tikzpicture}[>=stealth,font=\small,baseline=(capb.base)]
\begin{scope}[x=1cm,y=1cm]
 \draw[->] (0,0) -- (5.65,0);
 \draw[->] (0,0) -- (0,3.55);
 \node[below left,inner sep=1pt] at (0,0) {\footnotesize \(0\)};
 \node[below,inner sep=2pt] at (3.05,0)
   {\footnotesize accumulated scaled gap \(\bar\Gamma_j(v)\)};
 \draw[very thick,domain=0:42,samples=80,variable=\g]
   plot ({0.13*\g},{0.40*sqrt(2+\g)});
 \draw[dashed,very thick] (0,0) -- (5.46,3.36);
 \node[anchor=south east,inner sep=2pt] at (5.46,3.36)
   {\footnotesize contraction credit};
 \node[anchor=north east,inner sep=2pt] at (5.46,2.07)
   {\footnotesize noise bound};
 \draw[<->,thick] (0.5525,0.34) -- (0.5525,1.00);
 \draw[gray] (0.5525,0.67) -- (1.42,2.46);
 \node[anchor=south,align=center,inner sep=2pt] at (1.58,2.54)
   {\footnotesize largest excess of\\[-2pt]\footnotesize noise bound over credit};
 \node (capb) at (2.45,-0.75) {(b) noise bound and contraction credit};
\end{scope}
\end{tikzpicture}
\caption{Noise and contraction credit for a unit comparator \(v\).
(a) The depth of \(v\) below the tangent plane at \(e_t\) is half the
squared length of the dotted chord. Multiplying this depth by
\(\twonorm{\widehat\Theta_{t-1}}\) gives the gap. Below the dashed plane through
\(z_t\), the contraction credit exceeds its cost.
(b) As the accumulated gap grows, the linear credit \(B_j\bar\Gamma_j(v)/4\)
overtakes the noise bound \(\widetilde O(B_j)\sqrt{2+\bar\Gamma_j(v)}\).
The arrow marks the largest excess of the noise bound over the credit.}
\label{fig:cover-main}
\end{figure}

\qixproofstage{4}{Control the estimation noise uniformly via a lifted cover}
Work within a single dyadic epoch \(I_j=\{t:B_j\le\beta_t<2B_j\}\), where
the regularisation scale stays within a factor of two. Define \(v\)'s
\emph{accumulated scaled gap} on the epoch's contraction rounds by
\[
 \bar\Gamma_j(v)=\sum_{t\in I_j:\,\epsilon_t>0}\frac{\Gamma_{t-1}(v)}{\beta_t}.
\]
The noise \(-\iprod{z_t-v}{\xi_t}=\iprod{v}{\xi_t}-\iprod{z_t}{\xi_t}\) is
affine in \(v\). Following \citet[Section~3]{lee2020bias} and
\citet[Section~3.1]{zimmert2022bias}, appending a constant coordinate makes
it linear: setting \(\bar v=(-v,1)\in\R^{d+1}\) and
\(\bar\xi_t=-\bigl(\xi_t,\,(1-p_t)\iprod{e_t}{\xi_t}\bigr)\in\R^{d+1}\)
gives \(-\iprod{z_t-v}{\xi_t}=\iprod{\bar v}{\bar\xi_t}\), where
\(\bar\xi_t\) does not depend on \(v\).

The decisive property of the Euclidean sphere is that the \emph{same}
lifted vector \(\bar v=(-v,1)\) turns the scaled gap into an exact
quadratic form in the trigger matrix \(A_t\) and final epoch cover
\(H_j=I_{d+1}+\sum_{s\in I_j:\,\epsilon_s>0}A_s\) of
Section~\ref{sec:algorithm}: because \(\twonorm{v}=1\) (so
\(\bar v\transpose I_{d+1}\bar v=2\)),
\[
 \bar v\transpose A_t\bar v
 =\frac{r_t}{2}\bigl(\twonorm{v}^2-2\iprod{v}{e_t}+1\bigr)
 =\frac{\Gamma_{t-1}(v)}{\beta_t},
 \qquad
 \bar v\transpose H_j\bar v
 =2+\bar\Gamma_j(v)
\]
(Lemma~\ref{lem:trigger-cover}). Because the trigger test ensures
\(A_t\preceq H_j\) on every round \(t\in I_j\), every round's scaled gap
is bounded by the epoch's cover allowance:
\(\Gamma_{t-1}(v)/\beta_t\le2+\bar\Gamma_j(v)\).

Applying Cauchy--Schwarz in the norm induced by the final cover \(H_j\)
now separates the comparator completely from the noise:
\begin{equation}
 -\sum_{t\in I_j}\iprod{z_t-v}{\xi_t}
 =\iprod[\bigg]{\bar v}{\sum_{t\in I_j}\bar\xi_t}
 \le\norm[\bigg]{\sum_{t\in I_j}\bar\xi_t}_{H_j^{-1}}
    \sqrt{2+\bar\Gamma_j(v)}.
 \label{eq:lifted-cs}
\end{equation}
The right-hand side depends on \(v\) only through \(\bar\Gamma_j(v)\).
Thus a single high-probability bound on the random vector norm
\(\normt{H_j^{-1}}{\sum_{t\in I_j}\bar\xi_t}\) controls \emph{every} unit
comparator \(v\) at once, without a union bound over a net.

Bounding this dual norm requires two ingredients because the cover \(H_j\)
grows during the epoch and the variance depends on the unobserved losses:
\begin{enumerate}[label=(\roman*),leftmargin=1.8em,itemsep=2pt,topsep=2pt]
\item \emph{Moving-cover concentration.} Let \(H_{j,t}\) be the cover right
after the round-\(t\) trigger test; it is predictable (known before the
fresh draw) and nondecreasing in the Loewner order, with final value
\(H_j\). On round \(t\in I_j\), the post-test condition \(A_t\preceq H_{j,t}\)
implies that \(H_{j,t}\) dominates the lifted noise metric
(Lemma~\ref{lem:cover-domination}), yielding predictable range
\(\normt{H_{j,t}^{-1}}{\bar\xi_t}=\widetilde O(B_j)\) and conditional
variance trace
\[
 \E_t\normt{H_{j,t}^{-1}}{\bar\xi_t}^2\le10d\normt{G_t}{\theta_t}^2.
\]
Normalising the running noise sum by \(H_{j,t}^{-1/2}\) turns it into a
Hilbert-space recursion under predictable contractions
\(K_t=H_{j,t}^{-1/2}H_{j,t-1}^{1/2}\) (contractions because
\(H_{j,t-1}\preceq H_{j,t}\)). Section~\ref{sec:contractive-concentration}
develops a dimension-free Bennett bound for such recursions
(Proposition~\ref{prop:contractive-classical}) without a log-determinant
penalty.
\item \emph{Epoch variance cap.} To invoke
Proposition~\ref{prop:contractive-classical:fixed}, we need a deterministic
upper bound on the epoch's conditional variance
\(\sum_{t\in I_j}10d\normt{G_t}{\theta_t}^2\). Throughout epoch \(I_j\),
the scale rule keeps the observed squares \(\EestObs_{t-1}=\widetilde O(B_j^2)\)
(since \(a_{j_t}\sqrt{\EestObs_{t-1}+1}\le\beta_t<2B_j\)),
while Lemma~\ref{lem:qix:sandwich} guarantees
\(\E_t[p_t\normt{G_t}{\widehat\theta_t}^2]\ge(d/8)\normt{G_t}{\theta_t}^2\).
Comparing observed squares with their conditional means via an exponential
supermartingale (Lemma~\ref{lem:shell-reverse-clock}) caps the unobserved
epoch quadratic form on one high-probability event, simultaneously for all
epochs \(j\ge0\):
\begin{equation}
 d\sum_{t\in I_j}\normt{G_t}{\theta_t}^2=\widetilde O(B_j^2).
 \label{eq:epoch-variance-cap}
\end{equation}
\end{enumerate}

Applying Proposition~\ref{prop:contractive-classical:fixed} with the
variance cap \eqref{eq:epoch-variance-cap} yields one high-probability
event (Lemma~\ref{lem:lifted-uniform}) on which, simultaneously for every
epoch \(j\ge0\),
\[
 \norm[\bigg]{\sum_{t\in I_j}\bar\xi_t}_{H_j^{-1}}
 \le\varrho_j,
 \qquad
 \varrho_j=\widetilde O(B_j).
\]

\qixproofstage{5}{Pay for estimation noise and bias with contraction credit}
We now assemble the noise, bias, and credit inside each nonempty epoch
\(I_j\), on the intersection of the two events from Step~4:
\begin{itemize}[leftmargin=1.5em,itemsep=2pt,topsep=2pt]
\item \emph{Noise.} Substituting the lifted-noise bound into
\eqref{eq:lifted-cs} gives
\[
 -\sum_{t\in I_j}\iprod{z_t-v}{\xi_t}
 \le\varrho_j\sqrt{2+\bar\Gamma_j(v)}
 =\widetilde O(B_j)\sqrt{2+\bar\Gamma_j(v)}.
\]
\item \emph{Bias.} For every \(t\in I_j\), we have \(\beta_t\ge B_j\) and
\(\Gamma_{t-1}(v)/\beta_t\le2+\bar\Gamma_j(v)\). Moreover, the scale
increases by at most a factor of three per round (Lemma~\ref{lem:rate-jumps}),
so the lagged index \(j_t=j(\beta_{t-1})\) stays within two of \(j\),
giving \(\ell_{j_t}/\sqrt{j_t+1}\le\sqrt3\,\ell_j/\sqrt{j+1}\). Summing
\eqref{eq:bias-bound} over \(t\in I_j\) and applying the epoch cap
\eqref{eq:epoch-variance-cap} yields
\begin{align*}
 \sum_{t\in I_j}\lambda_t\iprod{z_t-v}{G_t\theta_t}
 &\le\frac{\sqrt3\,\ell_j}{\sqrt{j+1}\,B_j}
    \Biggl(d\sum_{t\in I_j}\normt{G_t}{\theta_t}^2\Biggr)
    \sqrt{1+2\bigl(2+\bar\Gamma_j(v)\bigr)}\\
 &=\widetilde O(B_j)\sqrt{2+\bar\Gamma_j(v)}.
\end{align*}
\item \emph{Credit.} On every trigger round in \(I_j\), \(\epsilon_t=1/4\)
and \(\beta_t\ge B_j\) (Lemma~\ref{lem:trigger-credit}; compare
\citealp[Lemma~B.13]{lee2020bias} and
\citealp[Appendix~F, Step~4]{zimmert2022bias}), so
\[
 -\sum_{t\in I_j}\epsilon_t\Gamma_{t-1}(v)
 =-\frac14\sum_{t\in I_j:\,\epsilon_t>0}\beta_t\frac{\Gamma_{t-1}(v)}{\beta_t}
 \le-\frac{B_j}{4}\bar\Gamma_j(v).
\]
\end{itemize}

Writing \(c_j=\widetilde O(B_j)\) for the combined noise and bias
coefficient in epoch \(j\), summing the three bounds across all nonempty
epochs yields, simultaneously for every unit comparator \(v\)
(Proposition~\ref{prop:epoch-error} and Lemma~\ref{lem:bandit-correction}),
\begin{align*}
 \sum_{t=1}^T\bigl\{\iprod{z_t-v}{\theta_t-\widehat\theta_t}
 -\epsilon_t\Gamma_{t-1}(v)\bigr\}
 &=\sum_j\sum_{t\in I_j}\bigl\{
   -\iprod{z_t-v}{\xi_t}
   +\lambda_t\iprod{z_t-v}{G_t\theta_t}
   -\epsilon_t\Gamma_{t-1}(v)\bigr\}\\
 &\overset{\textup{(a)}}{\le}\sum_j\Bigl\{c_j\sqrt{2+\bar\Gamma_j(v)}
   -\frac{B_j}{4}\bar\Gamma_j(v)\Bigr\}\\
 &\overset{\textup{(b)}}{\le}\sum_j\Bigl(\frac{B_j}{2}+\frac{c_j^2}{B_j}\Bigr)
  \overset{\textup{(c)}}{\le}\widetilde O(\beta_T).
\end{align*}
Here (a) inserts the epochwise noise, bias, and credit bounds; (b) uses
that noise and bias grow only as the square root of \(\bar\Gamma_j(v)\),
whereas credit grows linearly (Figure~\ref{fig:cover-main}b), so completing
the square eliminates the comparator regardless of its accumulated gap:
for every \(c>0\),
\[
 \sup_{x\ge0}\Bigl\{c\sqrt{2+x}-\frac{B_j}{4}x\Bigr\}
 \le\frac{B_j}{2}+\frac{c^2}{B_j};
\]
and (c) uses \(c_j^2/B_j=\widetilde O(B_j)\) and sums the geometric
thresholds \(B_j=2^j\beta_0\) over all nonempty epochs \(j\le j(\beta_T)\),
giving \(\sum_j B_j<2\beta_T\).

\qixproofstage{6}{Sum the stability costs and prove Proposition~\ref{prop:main-reduction}}
Returning to the intermediate bound \eqref{eq:step2-intermediate},
the only remaining term is the local-norm stability sum
\(\sum_{t=1}^T p_t\normt{G_t}{\widehat\theta_t}^2/\beta_t\). Because the
adaptive scale rule tracks the cumulative observed squares
\(\EestObs_t=\sum_{s=1}^t p_s\normt{G_s}{\widehat\theta_s}^2\), this sum
telescopes cleanly. Set \(w_t=\sqrt{\EestObs_t+1}\) (with \(w_0=1\)), so
that \(p_t\normt{G_t}{\widehat\theta_t}^2=w_t^2-w_{t-1}^2=(w_t+w_{t-1})(w_t-w_{t-1})\).
For every \(t\le T\), monotonicity of \(a_j\) and the scale rule give
\(w_{t-1}\le\beta_t/a_{j_t}\le\beta_t/a_{j_T}\), while the single-step
range bound \(p_t\normt{G_t}{\widehat\theta_t}^2\le\beta_t^2/\ell_{j_t}^2\)
(Lemma~\ref{lem:qix:local}) keeps
\(w_t=\sqrt{w_{t-1}^2+p_t\normt{G_t}{\widehat\theta_t}^2}<\sqrt2\,\beta_t/a_{j_T}\)
(as \(a_{j_T}<\ell_{j_t}\)). Since \(1+\sqrt2<3\), dividing by \(\beta_t\)
and telescoping yields (Lemma~\ref{lem:lagged-clock})
\[
 \sum_{t=1}^T\frac{p_t\normt{G_t}{\widehat\theta_t}^2}{\beta_t}
 =\sum_{t=1}^T\frac{w_t+w_{t-1}}{\beta_t}(w_t-w_{t-1})
 \le\frac{3}{a_{j_T}}\bigl(\sqrt{\EestObs_T+1}-1\bigr).
\]

It remains to replace the observed squares \(\EestObs_T\) by the
unobserved loss quadratic form. By Lemma~\ref{lem:qix:sandwich}, each
summand \(p_t\normt{G_t}{\widehat\theta_t}^2\) has conditional mean at
most \(d\normt{G_t}{\theta_t}^2\) and range at most \(\beta_t^2/\ell_{j_t}^2\).
An exponential supermartingale bound (Lemma~\ref{lem:upper-clock})
therefore gives, on one horizon-free event,
\begin{equation}
 \EestObs_T\le\frac32d\sum_{t=1}^T\normt{G_t}{\theta_t}^2
 +\frac{24}{\ell_{j(\beta_T)}}\beta_T^2.
 \label{eq:upper-clock-overview}
\end{equation}
Substituting \eqref{eq:upper-clock-overview} into the telescoped stability
sum and using \(a_{j_T}\ge a_j=\sqrt{\ell_j/\{48(j+1)\}}\) with
\(j=j(\beta_T)\) gives (Lemma~\ref{lem:summed-stability})
\[
 \sum_{t=1}^T\frac{p_t\normt{G_t}{\widehat\theta_t}^2}{\beta_t}
 \le\widetilde O\!\left(\beta_T+\sqrt{d\sum_{t=1}^T\normt{G_t}{\theta_t}^2}\right).
\]

We can now complete Proposition~\ref{prop:main-reduction}. On the
intersection of the high-probability events for sampling noise (Step~1),
epochwise variance and lifted noise (Step~4), and observed squares above,
substituting Step~1, Step~5, and the stability bound into
\eqref{eq:step2-intermediate} gives, simultaneously for every unit \(v\)
(and hence every \(v\in\ball\)),
\[
 \sum_{t=1}^T\iprod{a_t-v}{\theta_t}
 \le\widetilde O\!\left(\beta_T+\sqrt{d\sum_{t=1}^T\normt{G_t}{\theta_t}^2}\right).
\]
Appendix~\ref{sec:assembly} tracks the explicit finite-horizon constants
in \eqref{eq:regret-at-scale} and extends the sampling event over dyadic
horizons \(T_k=2^k\) to make the guarantee simultaneous for all \(T\ge1\),
proving Proposition~\ref{prop:main-reduction}.

\qixproofstage{7}{Bound the final scale and conclude Theorem~\ref{thm:worst-case}}
To turn Proposition~\ref{prop:main-reduction} into an explicit regret
bound, we must control the final regularisation scale \(\beta_T\). By
Algorithm~\ref{alg:radial-qix}, \(\beta_T\) is the maximum of three
sources: the initial scale \(\beta_0=\widetilde O(\sqrt d)\), the
observed-square candidate \(a_{j_t}\sqrt{\EestObs_{t-1}+1}\), and the
cubic floor \(\rho_{\kappa_{j_t}}(\twonorm{\widehat\Theta_{t-1}})\).

The observed-square candidate is already controlled by
\eqref{eq:upper-clock-overview}. On any round \(s\) where this candidate
raises the scale to \(\beta_s=a_{j_s}\sqrt{\EestObs_{s-1}+1}\), applying
\eqref{eq:upper-clock-overview} at prefix \(s-1\) (where
\(j(\beta_{s-1})=j_s\) and \(\beta_{s-1}\le\beta_s\)) and using
\(24a_{j_s}^2/\ell_{j_s}=1/\{2(j_s+1)\}\le1/2\) gives
\[
 \beta_s^2
 =a_{j_s}^2(\EestObs_{s-1}+1)
 \le a_{j_s}^2\Biggl(\frac32d\sum_{t=1}^{s-1}\normt{G_t}{\theta_t}^2+1\Biggr)
  +\frac12\beta_s^2.
\]
Absorbing \(\beta_s^2/2\) into the left-hand side (and noting
\(a_{j_s}\le a_0<\beta_0=\widetilde O(\sqrt d)\)) bounds this candidate by
\(\widetilde O\bigl(\sqrt{d\sum_{t=1}^T\normt{G_t}{\theta_t}^2}+\sqrt d\bigr)\)
(Lemma~\ref{lem:scale-envelope}).

What remains is the cubic floor
\(\rho_{\kappa_{j_t}}(\twonorm{\widehat\Theta_{t-1}})\), which can force
\(\beta_T\) upward whenever the state norm \(\twonorm{\widehat\Theta_{t-1}}\)
grows. This requires care because a unit loss aligned with \(-e_t\) adds
directly to \(\twonorm{\widehat\Theta_{t-1}}\) even though its local
squared size \(\normt{G_t}{\theta_t}^2=p_t\) is small near the boundary.
To bound state growth independently of regret, Appendix~\ref{sec:radial-growth}
tracks the \emph{effective radial potential}
\(\Phi_{\kappa_0}(h)=\int_0^h\{\sqrt{\kappa_0}/\rho_{\kappa_0}(u)\}\,du
 =\frac{3}{4\sqrt{\kappa_0}}\{\rho_{\kappa_0}(h)^2-\kappa_0\}\)
at \(h=\twonorm{\widehat\Theta_t}\), whose derivative
\(\Phi_{\kappa_0}'(\twonorm{\widehat\Theta_{t-1}})
 =\sqrt{\kappa_0}/\rho_{\kappa_0}(\twonorm{\widehat\Theta_{t-1}})
 \le\sqrt{p_t}\)
weights radial steps by the cubic-floor exploration rate, and applies
\citet{freedman1975} concentration to the weighted radial noise. With the
\emph{excess radial drift} \(\mathsf E_T\ge0\) defined in
Section~\ref{sec:main-result}, Lemma~\ref{lem:final-scale} gives
\[
 \beta_T\le\widetilde O\!\left(
 \sqrt{d\sum_{t=1}^T\bigl(\twonorm{\theta_t}^2-\twonorm{z_t}\iprod{e_t}{\theta_t}^2\bigr)+\sqrt d\,\mathsf E_T}
 +\sqrt d\right).
\]
Substituting this bound into Proposition~\ref{prop:main-reduction} with
\(v=u^\star\) yields the second-order bound of Theorem~\ref{thm:headline}.
In the worst case, each round satisfies \(\normt{G_t}{\theta_t}^2\le1\)
and contributes at most \(1/(4\sqrt{\kappa_0})\) to \(\mathsf E_T\), so
\(\sqrt d\,\mathsf E_T=O(T)\) (as \(\kappa_0\ge d\ell_0^2\)) and
\(\beta_T=\widetilde O(\sqrt{dT})\). Tracking the exact logarithmic
factors (Appendix~\ref{sec:headline-details}) yields
Theorem~\ref{thm:worst-case}.

\subsection{Concentration under predictable contractions}
\label{sec:contractive-concentration}
Step~4 needed a high-probability bound on the dual norm
\(\normt{H_{j,t}^{-1}}{S_t}\) of the cumulative lifted noise
\(S_t=\sum_{s\le t,\,s\in I_j}\bar\xi_s\) under a cover matrix \(H_{j,t}\)
that grows online. Whitening the partial sum by \(H_{j,t}^{-1/2}\) gives
the vector \(\mathcal Z_t=H_{j,t}^{-1/2}S_t\), whose Euclidean norm is the
desired dual norm. Because \(H_{j,t}\) can increase before round \(t\)'s
draw, \(\mathcal Z_t\) is not an ordinary martingale; instead it satisfies
the contractive recursion
\[
 \mathcal Z_t=K_t\mathcal Z_{t-1}+\zeta_t,
 \qquad
 K_t=H_{j,t}^{-1/2}H_{j,t-1}^{1/2},
 \qquad
 \zeta_t=H_{j,t}^{-1/2}\II{t\in I_j}\bar\xi_t.
\]
Here \(\zeta_t\) is a martingale difference (\(\E_t\zeta_t=0\)), and
\(K_t\) is a predictable linear contraction because
\(H_{j,t-1}\preceq H_{j,t}\) implies
\(K_tK_t^*=H_{j,t}^{-1/2}H_{j,t-1}H_{j,t}^{-1/2}\preceq I\).

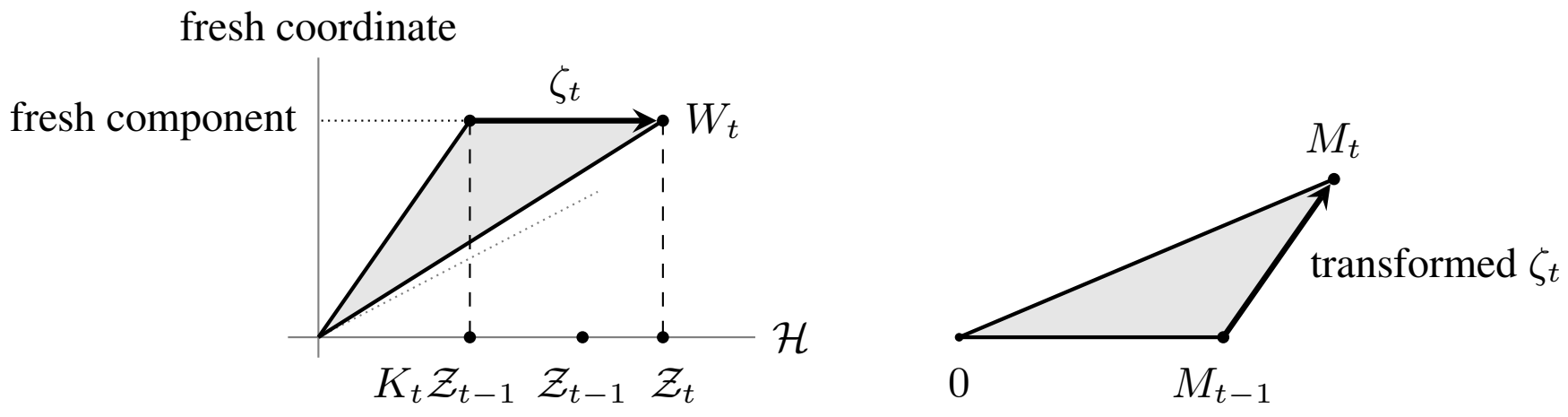
\begin{figure}[htbp]
\centering
\begin{tikzpicture}[>=stealth,font=\small,scale=0.75,baseline=(dila.base)]
 \fill[gray!20] (0,0) -- (1.491,2.130) -- (3.391,2.130) -- cycle;
 \draw[gray] (-0.3,0) -- (4.3,0);
 \draw[gray] (0,-0.2) -- (0,2.75);
 \node[right] at (4.3,0) {\(\mathcal H\)};
 \node[above] at (0,2.75) {fresh coordinate};
 \draw[gray,densely dotted] (0,0) -- (27.5:3.1);
 \draw[dashed] (1.491,2.130) -- (1.491,0);
 \draw[dashed] (3.391,2.130) -- (3.391,0);
 \draw[densely dotted] (1.491,2.130) -- (0,2.130);
 \draw[thick] (0,0) -- (1.491,2.130);
 \draw[thick] (0,0) -- (3.391,2.130);
 \draw[->,very thick] (1.491,2.130) -- (3.33,2.130);
 \node[above] at (2.44,2.130) {\(\zeta_t\)};
 \fill (2.6,0) circle (1.7pt);
 \fill (1.491,2.130) circle (1.7pt);
 \fill (3.391,2.130) circle (1.7pt);
 \fill (1.491,0) circle (1.7pt);
 \fill (3.391,0) circle (1.7pt);
 \node[below=3pt] at (1.25,0) {\(K_t\mathcal Z_{t-1}\)};
 \node[below=3pt] at (2.6,0) {\(\mathcal Z_{t-1}\)};
 \node[below=3pt] at (3.5,0) {\(\mathcal Z_t\)};
 \node[right=1pt] at (3.391,2.130) {\(W_t\)};
 \node[left=1pt] at (0,2.130) {\footnotesize fresh component};
 \node (dila) at (1.6,-1.15) {(a) apply the isometry, then add noise};
\end{tikzpicture}\hspace{0.7cm}
\begin{tikzpicture}[>=stealth,font=\small,scale=0.75,baseline=(dilb.base)]
 \fill[gray!20] (0,0) -- (2.6,0) -- (3.690,1.556) -- cycle;
 \draw[thick] (0,0) -- (2.6,0);
 \draw[thick] (0,0) -- (3.690,1.556);
 \draw[->,very thick] (2.6,0) -- (3.655,1.506);
 \fill (0,0) circle (1.2pt) node[below=3pt] {\(0\)};
 \fill (2.6,0) circle (1.7pt) node[below=3pt] {\(M_{t-1}\)};
 \fill (3.690,1.556) circle (1.7pt) node[above=1pt] {\(M_t\)};
 \node[right=3pt] at (3.15,0.7) {\footnotesize transformed \(\zeta_t\)};
 \node (dilb) at (2.2,-1.15) {(b) undo the isometry};
\end{tikzpicture}
\caption{A scalar contraction as a two-dimensional reflection. (a) The
Halmos isometry reflects \(\mathcal Z_{t-1}=W_{t-1}\) across the dotted
line, projecting onto \(K_t\mathcal Z_{t-1}\); adding \(\zeta_t\) gives
\(W_t\), projecting to \(\mathcal Z_t\). (b) Undoing the isometry preserves
increment length and gives \(\twonorm{M_t}=\twonorm{W_t}\ge\twonorm{\mathcal Z_t}\).}
\label{fig:dilation}
\end{figure}

To recover ordinary martingale concentration without paying a
log-determinant penalty, we embed the recursion in a larger Hilbert
space where each contraction becomes an isometry (Figure~\ref{fig:dilation}): we replace
each contraction \(K_t\) by its Halmos unitary dilation acting on
\(\mathcal H\) and a fresh orthogonal coordinate, producing a lifted
vector \(W_t\) that projects onto \(\mathcal Z_t\)
(Figure~\ref{fig:dilation}a). Undoing the cumulative unitary map then turns
\(W_t\) into an honest martingale \(M_t\) with the exact same increment
norms \(\twonorm{\zeta_t}\) and \(\twonorm{M_t}=\twonorm{W_t}\ge\twonorm{\mathcal Z_t}\)
(Figure~\ref{fig:dilation}b).

\begin{lemma}[Martingale dilation of predictable contractions]
\label{lem:contractive-dilation}
Let \(\mathcal H\) be a real separable Hilbert space, let \(K_t\) be
predictable (strongly measurable) linear contractions, and let \(\zeta_t\) be square-integrable
\(\mathcal H\)-valued martingale differences. Define
\(\mathcal Z_0=0\) and \(\mathcal Z_t=K_t\mathcal Z_{t-1}+\zeta_t\).
There is a martingale \(M_t\), with \(M_0=0\), in a separable Hilbert
space \(\mathcal K\), such that
\[
 \twonorm{\mathcal Z_t}\le\twonorm{M_t},\qquad
 \twonorm{M_t-M_{t-1}}=\twonorm{\zeta_t}.
\]
\end{lemma}

Because \(M_t\) is a Hilbert-space martingale whose increments have the
same norms and conditional second moments as \(\zeta_t\), any
Hilbert-space martingale inequality transfers verbatim to
\(\twonorm{\mathcal Z_t}\). Applying the Bennett inequality of
\citet[Theorem~3.4]{pinelis1994optimum} to \(M_t\) yields horizon-free
bounds uniform over all prefixes, depending only on the accumulated
conditional variance trace
\(\sigma_\tau^2=\sum_{t\le\tau}\E_t\twonorm{\zeta_t}^2\).

\Needspace{11\baselineskip}
\begin{proposition}[Bennett bound under predictable contractions]
\label{prop:contractive-classical}
Under the assumptions of Lemma~\ref{lem:contractive-dilation}, suppose
\(\twonorm{\zeta_t}\le R\) almost surely for a deterministic \(R>0\),
and let \(\ell>0\).
\begin{lemenum}
\item \label{prop:contractive-classical:fixed}
For any deterministic \(v>0\), with probability at least \(1-e^{-\ell}\),
every \(\tau\ge1\) with \(\sigma_\tau^2\le v\) satisfies
\begin{equation}
 \twonorm{\mathcal Z_\tau}
 \le\sqrt{2v(\ell+\log2)}+\frac{R(\ell+\log2)}3.
 \label{eq:contractive-fixed-v}
\end{equation}
\item \label{prop:contractive-classical:peeled}
With
\(\ell'_\tau=\ell+\log\{\pi^2(1+\ceil{\log_2(1\vee\sigma_\tau^2/R^2)})^2/3\}\),
with probability at least \(1-e^{-\ell}\), every \(\tau\ge1\) satisfies
\begin{equation}
 \twonorm{\mathcal Z_\tau}
 \le\sqrt{4(\sigma_\tau^2+R^2)\ell'_\tau}+\frac{R\ell'_\tau}3.
 \label{eq:contractive-peeled}
\end{equation}
\end{lemenum}
\end{proposition}

Appendix~\ref{sec:vector-tools} proves Lemma~\ref{lem:contractive-dilation}
and Proposition~\ref{prop:contractive-classical}, and
Lemma~\ref{lem:lifted-uniform} applies Part~(a) to the moving-cover
recursion of Step~4.

\FloatBarrier
\Needspace{12\baselineskip}
\section{Upper and lower bounds in the loss quadratic form}
\label{sec:main-result}
The worst-case proof of Section~\ref{sec:proof-strategy} ends by
substituting \(\sum_{t=1}^T\normt{G_t}{\theta_t}^2\le T\) and a worst-case
bound on the final scale \(\beta_T\). Keeping the actual loss quadratic
form yields a second-order upper bound with an excess radial drift
residual, exposes a \(\widetilde\Theta(T^{1/3})\) exploration cost on
aligned losses, and motivates a general lower bound showing that no
nonanticipating learner can achieve a polylogarithmic guarantee in the
loss quadratic form alone at high confidence. Further consequences of
Theorem~\ref{thm:headline}---for the cancellation slack and for random
loss vectors---are derived in Appendix~\ref{sec:consequences}.

\subsection{Second-order upper bound}
Near the boundary, a unit loss \(\theta_t=-e_t\) has small local squared
size \(\normt{G_t}{\theta_t}^2=p_t\), yet points along \(\widehat\Theta_{t-1}\)
and can grow its norm, forcing the cubic floor to raise the scale.
Evaluating the linearised conditional mean increment of the effective
radial potential
\(\Phi_{\kappa_0}(h)=\frac{3}{4\sqrt{\kappa_0}}\{(\kappa_0^{3/2}+2\kappa_0h)^{2/3}-\kappa_0\}\)
(Appendix~\ref{sec:radial-growth}) weights the outward radial step
\(\iprod{e_t}{\widehat\Theta_{t-1}-\E_t\widehat\Theta_t}\) by
\(\Phi_{\kappa_0}'(\twonorm{\widehat\Theta_{t-1}})
 =\sqrt{\kappa_0}/(\kappa_0^{3/2}+2\kappa_0\twonorm{\widehat\Theta_{t-1}})^{1/3}
 \le\sqrt{p_t}\).
Define the largest interval drift \(\mathsf M_T\) (\(\mathsf M_T:=0\) for
\(T\le2\)) and the \emph{excess radial drift} \(\mathsf E_T\) above the
quadratic-form allowance:
\begin{equation}
 \begin{aligned}
 \mathsf M_T&:=\max\left\{0,
   \max_{2\le k\le \tau\le T-1}
   \sum_{t=k}^\tau\frac{\sqrt{\kappa_0}}{(\kappa_0^{3/2}+2\kappa_0\twonorm{\widehat\Theta_{t-1}})^{1/3}}\,
   \iprod{e_t}{\widehat\Theta_{t-1}-\E_t\widehat\Theta_t}\right\},\\
 \mathsf E_T&:=\Biggl[\mathsf M_T-\sqrt{\kappa_0}\Biggl(1+\sum_{t=1}^T\bigl(\twonorm{\theta_t}^2-\twonorm{z_t}\iprod{e_t}{\theta_t}^2\bigr)\Biggr)\Biggr]_+.
 \end{aligned}
 \label{eq:intro-net-drawup}
\end{equation}

\Needspace{7\baselineskip}
\begin{theorem}[Second-order regret bound]
\label{thm:headline}
With probability at least \(1-\delta\), simultaneously for all \(T\ge1\),
Algorithm~\ref{alg:radial-qix} satisfies
\begin{equation}
 \Reg_T=\widetilde O\!\left(\sqrt{d\sum_{t=1}^T\bigl(\twonorm{\theta_t}^2-\twonorm{z_t}\iprod{e_t}{\theta_t}^2\bigr)+\sqrt d\,\mathsf E_T}+\sqrt d\right).
 \label{eq:intro-readable-rate}
\end{equation}
\end{theorem}
When \(\mathsf E_T=0\), the rate is
\(\widetilde O\bigl(\sqrt{d\sum_{t\le T}\normt{G_t}{\theta_t}^2}+\sqrt d\bigr)\).
Sufficient tangential loss pays for radial drift:
\(\mathsf E_T=0\) whenever
\(\sum_{t=1}^T(\twonorm{\theta_t}^2-\iprod{e_t}{\theta_t}^2)\ge T/(4\kappa_0)\)
(Corollary~\ref{cor:headline-relaxations},
Appendix~\ref{sec:ball-headline-relaxations}), which also forces
\(\sum_{t=1}^T\normt{G_t}{\theta_t}^2\ge T/(4\kappa_0)\) and keeps the bound at least
order \(\sqrt T\). In general, \(\mathsf E_T\le\sqrt{T\sum_{t\le T}\normt{G_t}{\theta_t}^2}\)
gives a bound in the loss quadratic form alone,
\begin{equation}
 \Reg_T=\widetilde O\Biggl(\sqrt{d\sum_{t=1}^T\normt{G_t}{\theta_t}^2}
   +\Bigl(dT\sum_{t=1}^T\normt{G_t}{\theta_t}^2\Bigr)^{1/4}+\sqrt d\Biggr),
 \label{eq:v-only}
\end{equation}
whose fourth-root term dominates when \(\sum_{t=1}^T\normt{G_t}{\theta_t}^2\le T/d\).

\paragraph{Worst-case rate.}
Since a round's weighted drift is at most \(\normt{G_t}{\theta_t}\) while
its allowance is \(\sqrt{\kappa_0}\normt{G_t}{\theta_t}^2\), only the
\(T_{\mathrm{low}}\) rounds \(2\le t\le T-1\) with
\(\normt{G_t}{\theta_t}^2<\kappa_0^{-1}=O(d^{-1}\ell_0^{-2})\) contribute to
\(\mathsf E_T\), each at most \(1/(4\sqrt{\kappa_0})\), giving
\(\mathsf E_T\le T_{\mathrm{low}}/(4\sqrt{\kappa_0})\) and the \emph{count form}
\begin{equation}
 \Reg_T=\widetilde O\!\left(\sqrt{d\sum_{t=1}^T\bigl(\twonorm{\theta_t}^2-\twonorm{z_t}\iprod{e_t}{\theta_t}^2\bigr)+T_{\mathrm{low}}}+\sqrt d\right)
 \label{eq:count-form}
\end{equation}
and hence \(\widetilde O(\sqrt{dT})\).
Appendix~\ref{sec:headline-details} gives the explicit bound
\eqref{eq:headline-explicit} and deduces Theorem~\ref{thm:worst-case}
from Proposition~\ref{prop:fixed-horizon}.

\Needspace{8\baselineskip}
\subsection{Aligned-loss separation}
\label{sec:aligned-losses}
On a constant nonzero loss, exact feedback lets a learner probe the \(d\)
coordinates and incur \(O(d)\) regret. On aligned losses
\(\theta_t=-\gamma_tu\) (\(\twonorm{u}=1\), \(\gamma_t>0\)), however, even
when \(e_t=u\), QIX incurs conditional expected regret \(\gamma_tp_t\) from
exploratory draws. When \(\twonorm{\widehat\Theta_{t-1}}\asymp t\) and the
cubic floor \eqref{eq:overview-rho-definition} sets \(\beta_t\), we have
\(\beta_t\asymp(dt)^{1/3}\) and \(p_t\asymp d^{1/3}t^{-2/3}\), summing to
order \((dT)^{1/3}\). Put \(\mathcal L_T=\log(40Tk_T^2/\delta)\), where
\(k_T=1\vee\ceil{\log_2T}\).

\begin{proposition}[Separation on aligned losses]
\label{prop:fixed-ray-separation}
Let \(\theta_t=-\gamma_tu\) with \(\twonorm{u}=1\) and \(\gamma_t\in[\gamma_0,1]\), where \(\gamma_0\in(0,1]\). With probability at
least \(1-\delta\), simultaneously for all \(T\ge1\), Algorithm~\ref{alg:radial-qix} satisfies
\begin{equation}
 \frac{c \gamma_0 dT}{\{\mathcal L_T^{3/2}d+\mathcal L_T^{4/3}(dT)^{1/3}\}^2}-2\log(32/\delta)
 \le\Reg_T\le C_0\bigl\{\mathcal L_T^3d+\mathcal L_T^{5/2}(dT)^{1/3}\bigr\},
 \label{eq:ray-regret-bounds}
\end{equation}
where \(c,C_0>0\) are absolute constants.
\end{proposition}
For fixed \(d,\delta,\gamma_0\), the regret is \(\widetilde\Theta(T^{1/3})\)
(proved in Appendix~\ref{sec:fixed-ray} alongside Corollary~\ref{cor:fixed-ray-count}).
On the same event, Theorem~\ref{thm:headline} gives
\(\widetilde O(d+(dT)^{1/3})\), whereas the count form \eqref{eq:count-form}
is \(\widetilde\Theta(\sqrt T)\).

\Needspace{10\baselineskip}
\subsection{Lower bound for quadratic-form guarantees}
\label{sec:lower-bound}
Start with losses of varying magnitude along one ray (optimal action \(e\))
and add a perpendicular perturbation of unknown sign on a hidden interval.
Playing \(e\) reveals nothing about the sign; playing near \(e\) reveals
little because scalar feedback mixes the radial and perpendicular terms;
moving away from \(e\) costs regret on the unperturbed sequence. Because
the two signs tilt the best hindsight action oppositely, a learner that
seldom leaves \(e\) incurs regret even when its loss quadratic form is small.

While Proposition~\ref{prop:fixed-ray-separation} quantified QIX's own
exploration rule, Theorem~\ref{thm:lower-bound} applies to any
nonanticipating learner (cf.\ \citealp{gerchinovitz2016} for finite-armed
bandits). Write \(x_t=(1-p_t)e_t\) for the learner's conditional mean
action before round \(t\) (\(\twonorm{x_t}=1-p_t\), unit direction \(e_t\)),
and \(G_t=I_d-\twonorm{x_t}e_te_t\transpose\) for its local metric \eqref{eq:q-clock}.

\begin{theorem}[Lower bound for quadratic-form guarantees]
\label{thm:lower-bound}
Let \(d\ge2\), \(T\ge1\), and let \(\mathsf F\) and \(\mathsf B\ge1\) be deterministic. If a
nonanticipating learner satisfies, on every deterministic loss sequence in
the unit ball,
\begin{equation}
 \bbP\Biggl\{\Reg_T\le\mathsf F+\mathsf B\sqrt{\sum_{t=1}^T\bigl(\twonorm{\theta_t}^2-\twonorm{x_t}\iprod{e_t}{\theta_t}^2\bigr)}\Biggr\}
 \ge1-\frac1{16T},
 \label{eq:lower-assumption}
\end{equation}
then, for an absolute constant \(c>0\),
\begin{equation}
 \mathsf F+\mathsf B^2+1
 \ge\frac{cT^{1/3}}{\mathsf B^{2/3}},
 \qquad\text{hence}\qquad
 \mathsf F+\mathsf B^2+1
 =\Omega\bigl(T^{1/4}\bigr).
 \label{eq:lower-conclusion}
\end{equation}
\end{theorem}
Thus at confidence \(1-1/(16T)\), either \(\mathsf F\) or \(\mathsf B\)
must grow with \(T\). Appendix~\ref{sec:lower-bound-proof} builds a
sequence whose cumulative loss norm is at least \(t/4\) at every prefix
\(t\), using varying radial norms in \([1/4,1/2]\) to mask the
perpendicular perturbation. Meanwhile, substituting
\(\mathsf E_T\le\sqrt{Tv}\) (with \(v=\sum_{t\le T}\normt{G_t}{\theta_t}^2\))
into Theorem~\ref{thm:headline} and applying
\((dTv)^{1/4}\le\tfrac12(dT)^{1/3}+\tfrac12(dT)^{1/6}\sqrt v\) gives
\(\mathsf F+\mathsf B^2=\widetilde O(d+(dT)^{1/3})\), leaving a gap between
exponents \(1/4\) and \(1/3\) at fixed \(d\).

\FloatBarrier
\Needspace{8\baselineskip}
\section{Discussion}
\label{sec:discussion}
Algorithm~\ref{alg:radial-qix} achieves anytime, comparator-uniform
\(\widetilde O(\sqrt{dT})\) high-probability regret on \(\ball\) in \(O(d)\)
time and memory per round, matching \citet{bubeck2012minimax} with
sub-Gaussian \(\sqrt{\log(e/\delta)}\) confidence scaling. Matching
exploration to the inverse barrier Hessian controls directional noise by
each comparator's gap, while the sphere's chord identity turns contraction
credit into a quadratic form in the lifted comparator, letting one
vector-martingale bound control the entire ball.

The main open question is how much exploration is needed when losses
persist along one ray. QIX's cubic floor introduces the excess radial drift
\(\mathsf E_T\) and incurs \(\widetilde\Theta(T^{1/3})\) regret on aligned
losses (Proposition~\ref{prop:fixed-ray-separation}), whereas
full-information methods attain constant or logarithmic regret
\citep{huang2017following,mayo2026smallloss}. Theorem~\ref{thm:lower-bound}
rules out polylogarithmic guarantees in \(\sum_{t\le T}\normt{G_t}{\theta_t}^2\)
alone, but it remains open whether high-probability \(\widetilde O(\sqrt{dT})\)
regret can coexist with fast rates on aligned losses.

Two further extensions remain open. On smooth, strongly convex sets
\citep{kerdreux2021linear}, the gap still lower-bounds squared distance to
the leader, leaving open a cover without the sphere's exact quadratic
identity. With constant-variance additive observation noise, \(d\sqrt T\)
regret is necessary on \(\ball\times\{1\}\)
\citep[Theorem~2]{shamir2015complexity}, and even noiseless affine feedback
\(\iprod{a_t}{\theta_t}+c_t\) maps to \(\ball\times\{1\}\); exact
zero-centred linear feedback is essential for QIX's boundary variance
scaling.

\ifdefined\InThesis\else
\section*{AI disclosure}
OpenAI's Codex, Anthropic's Claude, and Google's Gemini Neon assisted with proof auditing, revisions, experiments,
numerical verification, literature checks and editing, including suggested mathematical arguments. The author has checked all content, and bears full responsibility for the correctness, novelty and relevance of the results, including full and proper attribution of prior work used.
\fi

\ifdefined\InThesis\else
\clearpage
\begingroup
\urlstyle{same}
\bibliography{adaptive_qix_rate_insurance}
\endgroup
\clearpage
\fi
\ifdefined\qixappendixguide\qixappendixguide\fi
\appendix

\ifdefined\InThesis\else
\counterwithin{theorem}{section}
\counterwithin{equation}{section}
\fi

\section{Setup, geometry, and FTRL}
\label{sec:calibration}
\label{sec:local}

This appendix specifies the probability space and exact tuning of
Algorithm~\ref{alg:radial-qix}, establishes the deterministic geometry of
the radial barrier and QIX sampling law (proving Lemma~\ref{lem:qix}), and
derives the deterministic FTRL decomposition and stability bounds (proving
Lemma~\ref{lem:contracted-ftrl}). Algorithm~\ref{alg:radial-qix} depends
only on \((d,\delta)\); horizon-dependent grids appear only in the analysis.

\subsection{Probability space}
\label{sec:notation}

We use the notation of Sections~\ref{sec:setting}--\ref{sec:main-result}
and define local quantities where they arise.
All randomness lives on one probability space carrying i.i.d.\ draws
\((\Upsilon_t,U_t)_{t\ge1}\), with \(\Upsilon_t\sim\mathrm{Unif}[0,1]\)
independent of \(U_t\sim\mathrm{Unif}(\sphere^{d-1})\), and an independent
element \(\omega\) for the adversary's randomisation.
Set \(\mathcal F_0=\sigma(\omega)\) and
\(\mathcal F_t=\sigma(\omega,(\Upsilon_s,U_s)_{s\le t})\).
A process is \emph{predictable} if its round-\(t\) value is
\(\mathcal F_{t-1}\)-measurable and \emph{adapted} if it is
\(\mathcal F_t\)-measurable.

Algorithm~\ref{alg:radial-qix} uses its fresh draw through
\(Z_t=\II{\Upsilon_t\le p_t}\) and \(U_t\), which are conditionally
independent given \(\mathcal F_{t-1}\) with
\(Z_t\sim\mathrm{Bernoulli}(p_t)\) and \(U_t\sim\mathrm{Unif}(\sphere^{d-1})\);
the conditional moments in \eqref{eq:experiment-moment} and
Lemma~\ref{lem:qix} are taken in this sense.
Because the adversary is nonanticipating, \(\theta_t\), every pre-sampling
quantity in Algorithm~\ref{alg:radial-qix} (including the post-test cover,
\(\epsilon_t\), and \(\nu_t\)), and the unobserved quantities
\(\normt{G_t}{\theta_t}^2\) and \(\iprod{e_t}{\theta_t}\) are predictable.
Actions, feedback, estimates, and updated states are adapted, so
\(\xi_t=\widehat\theta_t-\E_t\widehat\theta_t\) is a martingale difference
sequence.

Throughout the appendices, \(P_{\parallel,t}=e_te_t\transpose\) and
\(P_{\perp,t}=I_d-P_{\parallel,t}\) denote the radial and tangential
projections (so \(G_t=P_{\perp,t}+p_tP_{\parallel,t}\)),
\(\Qcurv_\tau=\sum_{t=1}^\tau\normt{G_t}{\theta_t}^2\)
abbreviates the cumulative loss quadratic form \eqref{eq:q-clock}, and
\(C_0>0\) denotes an absolute constant that may change between occurrences.

\subsection{Exact tuning}
\label{sec:public-tuning}

Table~\ref{tab:tuning} collects the tuning schedule computed from
\(d\ge2\) and \(\delta\in(0,1)\), anchored by the base floor
\begin{equation}
 \kappa_0
 =\ell_0^2\,\frac{12d^2(d+3)}{(d+4)(7d+6)}.
 \label{eq:default-tuning}
\end{equation}

\begin{table}[ht]
\centering
\small
\caption{Tuning schedule and derived scale quantities used by
Algorithm~\ref{alg:radial-qix}, computed from \(d\ge2\) and \(\delta\in(0,1)\).}
\label{tab:tuning}
\renewcommand{\arraystretch}{1.45}
\begin{tabular}{@{}lll@{}}
\toprule
Quantity & Value & Role in Algorithm~\ref{alg:radial-qix}\\
\midrule
\(\delta_0\) & \(\delta/8\) & confidence level of the tuning\\
\(\ell_j\) & \(\log\{\pi^2(j+1)^2/(6\delta_0)\}\)
 & confidence log for epoch \(j\ge0\)\\
\(q_j,\,\kappa_j\) & \(q_j=\max\{1,(j+1)/d\}(\ell_j/\ell_0)^2,\; \kappa_j=q_j\kappa_0\)
 & cubic floor \(\rho_{\kappa_j}\), with \(\kappa_0\) in \eqref{eq:default-tuning}\\
\(a_j\) & \(\sqrt{\ell_j/\{48(j+1)\}}\)
 & observed-square weight \(a_{j_t}\sqrt{\EestObs_{t-1}+1}\)\\
\(\beta_0,\,B_j\) & \(\beta_0=\sqrt{\kappa_0},\; B_j=2^j\beta_0\)
 & base scale and dyadic threshold\\
\midrule
\(j(\beta),\,I_j\) & \(j(\beta)=\floor{\log_2(\beta/\beta_0)}\) (\(\beta\ge\beta_0\)),\; \(I_j=\{t:j(\beta_t)=j\}\)
 & tuning index \(j_t=j(\beta_{t-1})\); epoch \(B_j\le\beta_t<2B_j\)\\
\(\nu_t\) & \(d^2\ell_{j_t}^2/(4\beta_t^2)\)
 & QIX denominator in \eqref{eq:qix-estimator}\\
\bottomrule
\end{tabular}
\end{table}

\emph{Monotonicity and base scales.}
Since \(\delta_0=\delta/8<1/8\), we have
\(\ell_0=\log\{4\pi^2/(3\delta)\}>2.577\), so
\(\ell_j=\ell_0+2\log(j+1)\) strictly increases in \(j\).
On \([1,\infty)\), \(x\mapsto(\ell_0+2\log x)/x\) has derivative
\((2-\ell_0-2\log x)/x^2<0\), so \(a_j=\sqrt{\ell_j/\{48(j+1)\}}\) strictly
decreases in \(j\) and satisfies
\begin{equation}
 a_j^2\cdot\frac{24}{\ell_j}
 =\frac{1}{2(j+1)}\le\frac12
 \qquad(j\ge0).
 \label{eq:clock-absorption}
\end{equation}
Cross-multiplying in \eqref{eq:default-tuning} shows that
\(d\ell_0^2\le\kappa_0\le\tfrac{12}{7}d\ell_0^2\) (since
\((d-2)(5d+12)\ge0\) and \(21d\le34d+24\) for \(d\ge2\)); multiplying by
\(q_j=\max\{1,(j+1)/d\}(\ell_j/\ell_0)^2\) gives, for every \(j\ge0\) with
\(j_-:=[j-2]_+\),
\begin{equation}
 \max\{d,j+1\}\ell_j^2\le\kappa_j\le\tfrac{12}{7}\max\{d,j+1\}\ell_j^2,
 \qquad
 \ell_j\le2\ell_{j_-},
 \label{eq:kappa-bounds}
\end{equation}
where the second inequality holds because
\(\ell_j/\ell_{j_-}\le1+2\log3/\ell_0<2\). Moreover,
\begin{equation}
 a_0<\ell_0<\beta_0=\sqrt{\kappa_0},
 \qquad
 \ell_j\le B_j\quad(j\ge0),
 \label{eq:base-scales}
\end{equation}
since \(a_0^2=\ell_0/48<\ell_0^2\), \(\kappa_0\ge2\ell_0^2\), and
\(B_j/\ell_j>2^j\ell_0/\ell_j\ge1\) for all \(j\ge0\) and \(\ell_0>2.577\).
Because \(\twonorm{\widehat\Theta_0}=\EestObs_0=0\), \(a_0<\beta_0\), and
\(\rho_{\kappa_0}(0)=\beta_0\), the first scale is \(\beta_1=\beta_0\)
(\(j_1=0\)). Since \(\beta_t\) is nondecreasing, every round falls in an
epoch \(I_j\) (\(j\ge0\)) with \(j_t=j(\beta_{t-1})\le j\) and hence
\(a_{j_t}\ge a_j\).

\subsection{Barrier and spherical geometry}

We first record the spherical polynomial moments used in the sampling and
QIX calculations, then establish the barrier geometry, QIX sampling law,
and radial self-concordance.

\begin{lemma}[Polynomial moments on the sphere]
\label{lem:spherical-polynomial-moments}
Let \(d\ge1\) be an integer, let \(U\) be uniform on \(\sphere^{d-1}\),
let \(v\in\R^d\), and let \(A\) be a real symmetric \(d\times d\) matrix.
\begin{lemenum}
\item \(\E[UU\transpose]=I_d/d\).
\item \(\E[\iprod Uv^2UU\transpose]
 =\{\twonorm v^2I_d+2vv\transpose\}/\{d(d+2)\}\); in particular,
 \(\E\iprod Uv^4=3\twonorm v^4/\{d(d+2)\}\).
\item \(\E[\iprod Uv^4U\transpose AU]
 =3\{\twonorm v^4\tr A+4\twonorm v^2v\transpose Av\}/\{d(d+2)(d+4)\}\).
\end{lemenum}
\end{lemma}

\begin{proof}
Write \(Z=RU\sim N(0,I_d)\) with \(R=\twonorm Z\) independent of \(U\);
since \(R^2\sim\chi_d^2\), we have \(\E R^2=d\), \(\E R^4=d(d+2)\), and
\(\E R^6=d(d+2)(d+4)\). Dividing \(\E[ZZ\transpose]=I_d\) by \(\E R^2\)
proves~(a). For (b) and~(c) with \(v\ne0\), rotate so that
\(v=\twonorm v\,e_1\). Symmetry makes \(\E[Z_1^{2k}ZZ\transpose]\) diagonal
with first entry \(\E Z_1^{2k+2}\) and other diagonal entries
\(\E Z_1^{2k}\E Z_i^2\): for \(k=1\) (\(3\) and \(1\)) this yields
\(\E[\iprod Zv^2ZZ\transpose]=\twonorm v^2I_d+2vv\transpose\), and for
\(k=2\) (\(15\) and \(3\)) it yields
\(\E[\iprod Zv^4Z\transpose AZ]=3\twonorm v^4\tr A+12\twonorm v^2v\transpose Av\).
Dividing by \(\E R^4\) and \(\E R^6\) gives (b) and~(c).
\end{proof}

The barrier and conjugate in part~(a) below are those of
\citet[Theorem~6 and its proof]{bubeck2012minimax}.

Fix \(e\in\sphere^{d-1}\), \(h\ge0\), \(\beta>0\), \(\theta\in\ball\), and
\(\nu\ge0\).  Extend
\(F(w)=-\twonorm w-\log(1-\twonorm w)\) by \(+\infty\) outside
\(\operatorname{int}\ball\); put
\(P_\parallel=ee\transpose\), \(P_\perp=I_d-P_\parallel\),
\(\Psi_\beta=(\beta F)^*\).
The exploration probability is \(p=\beta/(h+\beta)\), the mean action is
\(z=(1-p)e\), and the sampling metric is \(G=P_\perp+pP_\parallel\).
Let \(U\sim\operatorname{Unif}(\sphere^{d-1})\) and, independently,
\(Z\sim\operatorname{Bernoulli}(p)\); put \(a=(1-Z)e+ZG^{1/2}U\) and
\(Y=\iprod a\theta\), and define the quadratic implicit-exploration family
\[
 \widehat\theta^{\nu}=(1-Z)eY+
 \frac{dZY}{p+\nu Y^2}G^{1/2}U
\]
(with \(\nu=0\) giving the unshrunk estimator).

\begin{proposition}[Barrier geometry and QIX sampling law]
\label{prop:qix-boundary-release}
\begin{lemenum}
\item \emph{Barrier derivatives and Hessian.}
The barrier \(F\) is \(C^2\) on \(\operatorname{int}\ball\) with
\(\nabla F(0)=0\), \(\nabla^2F(0)=I_d\), and at \(z=(1-p)e\),
\(\nabla F(z)=z/p\), \(\nabla^2F(z)=p^{-1}P_\perp+p^{-2}P_\parallel\), and
\begin{equation}
 \{\nabla^2F(z)\}^{-1}=pP_\perp+p^2P_\parallel=pG.
 \label{eq:inverse-hessian}
\end{equation}
Its conjugate is
\(\Psi_\beta(y)=\twonorm y-\beta\log(1+\twonorm y/\beta)\); at \(y=he\),
\(z=\nabla\Psi_\beta(he)\) is the unique minimiser of
\(w\mapsto-\iprod{he}{w}+\beta F(w)\).
\item \emph{Sampling moments.}
The law of \(a\) is supported on \(\ball\), with
\(\E a=z\), \(d\,\E[Zaa\transpose]=pG\), and
\(\normt{G}{\theta}^2=d\,\E[Y^2\mid Z=1]\).
\item \emph{QIX estimator and bias.}
The unshrunk estimator is unbiased, \(\E \widehat\theta^0=\theta\); moreover,
\begin{equation}
 \theta-\E \widehat\theta^{\nu}
 =\lambda G\theta
 =\frac{\lambda}{p}\{\nabla^2F(z)\}^{-1}\theta,
 \label{eq:qix-natural-gradient-alignment}
\end{equation}
where \(\lambda=0\) if \(\theta=0\) or \(\nu=0\), and otherwise
\begin{equation}
 \lambda=\frac{d\nu}{\normt{G}{\theta}^2}\,\E_U\!\left[
 \frac{\iprod{U}{G^{1/2}\theta}^4}{p+\nu\iprod{U}{G^{1/2}\theta}^2}\right]\in[0,1].
 \label{eq:qix-lambda}
\end{equation}
\end{lemenum}
\end{proposition}
Thus \(d\) times the exploratory second moment equals the inverse barrier
Hessian at the mean action. QIX stabilisation introduces bias in the same
inverse-Hessian direction, preserving the geometry that
determined the exploration.

\begin{proofof}{Proposition~\ref{prop:qix-boundary-release}.}
\emph{(a) Barrier mean and Hessian.}
Radial conjugacy at \(s=\twonorm y\) gives
\(F^*(y)=\sup_{0\le r<1}\{sr+r+\log(1-r)\} =s-\log(1+s)\),
uniquely maximised at \(r=s/(1+s)\).  Scaling gives
\(\Psi_\beta\), whose gradient is
\(\nabla\Psi_\beta(y)=y/(\beta+\twonorm y)\); thus
\(\nabla\Psi_\beta(he)=he/(\beta+h)=(1-p)e\).
For \(f(r)=-r-\log(1-r)\), the barrier's tangential and radial
Hessian eigenvalues satisfy, as \(r\downarrow0\),
\[
 \frac{f'(r)}r=\frac1{1-r}\to1,
 \qquad f''(r)=\frac1{(1-r)^2}\to1.
\]
Thus the radial Hessian extends continuously to the origin, with
\(\nabla F(0)=0\) and \(\nabla^2F(0)=I_d\).
At \(r=1-p\), \(\nabla F(z)=f'(r)e=z/p\), the tangential Hessian
eigenvalue is \(p^{-1}\), and the radial eigenvalue is \(p^{-2}\), giving
\eqref{eq:inverse-hessian} including at \(p=1\). Strict convexity and
boundary divergence of \(F\) make \(z=\nabla\Psi_\beta(he)\) the unique
minimiser of \(w\mapsto-\iprod{he}{w}+\beta F(w)\), since
\(\beta\nabla F(z)=\beta(1-p)e/p=he\).

\emph{(b) Sampling moments.}
Spherical symmetry gives \(\E U=0\) and \(\E UU\transpose=I_d/d\), so
\(\E a=(1-p)e=z\), \(d\,\E[Zaa\transpose]=pG\), and
\(d\,\E[Y^2\mid Z=1]=\normt{G}{\theta}^2\). Since \(G\preceq I_d\), the law
is supported on \(\ball\). The resolution \((1-p)P_\parallel+G=I_d\) gives
\(\E \widehat\theta^0=(1-p)P_\parallel\theta+G\theta=\theta\),
so the unshrunk estimator is unbiased.

\emph{(c) QIX estimator and bias.}
Conditioning on \(Z\) gives
\[
 \E \widehat\theta^{\nu}
 =(1-p)P_\parallel\theta
 +dp\,G^{1/2}\E_U
 \frac{U\iprod{U}{G^{1/2}\theta}}
      {p+\nu\iprod{U}{G^{1/2}\theta}^2}.
\]
Put \(v=G^{1/2}\theta\). Subtracting the conditional mean above from
\(\E\widehat\theta^0=\theta\) gives
\[
 \theta-\E\widehat\theta^\nu
 =dG^{1/2}\E_U\left[
 U\iprod Uv\left(1-\frac p{p+\nu\iprod Uv^2}\right)\right]
 =d\nu G^{1/2}\E_U
       \frac{U\iprod Uv^3}{p+\nu\iprod Uv^2}.
\]
For \(\nu>0\) and \(v\ne0\), rotational symmetry makes the expectation
parallel to \(v\); taking its inner product with \(v\) yields
\eqref{eq:qix-natural-gradient-alignment} with \(\lambda\) as in
\eqref{eq:qix-lambda}, which satisfies
\(0\le\lambda\le\frac{d}{\twonorm v^2}\E_U\iprod{U}{v}^2=1\).
If \(v=0\) or \(\nu=0\), \(\lambda=0\in[0,1]\).
\end{proofof}

\begin{lemma}[Bregman bound for the radial barrier]
\label{lem:radial-barrier-bregman}
For every \(y,z\) in the open unit ball,
\begin{equation}
 D_F(y,z)
 \ge\omega\bigl(\normt{\nabla^2F(z)}{y-z}\bigr),
 \qquad
 \omega(r)=r-\log(1+r).
 \label{eq:radial-barrier-bregman}
\end{equation}
\end{lemma}

\begin{proof}
We follow \citet[Section~9.6.3]{boyd2004convex} and verify continuity
through the origin. If \(y=z\), both sides vanish. Otherwise put
\(w=y-z\) and \(\phi(t)=F(z+tw)\), so \(\phi''(t)>0\) on \([0,1]\).
Away from any crossing of the origin, write \(z+tw=(1-p)e\) and
\(w=\alpha e+w_\perp\), and set \(A=\alpha/p\) and
\(C=\twonorm{w_\perp}/\sqrt p\). By
Proposition~\ref{prop:qix-boundary-release}(a),
\(\phi''(t)=A^2+C^2\) and \(\phi'''(t)=2A^3+3AC^2\). With
\(c=|A|/\sqrt{A^2+C^2}\in[0,1]\),
\(|2A^3+3AC^2|/(A^2+C^2)^{3/2}=3c-c^3\le2\), so
\(q(t)=\phi''(t)^{-1/2}\) satisfies \(|q'(t)|\le1\) wherever \(\phi'''\)
exists. Continuity of \(\nabla^2F\) at \(0\)
(Proposition~\ref{prop:qix-boundary-release}(a)) makes \(q\) continuous
and hence globally \(1\)-Lipschitz on \([0,1]\). With
\(r=\sqrt{\phi''(0)}=\normt{\nabla^2F(z)}{w}\), we have
\(q(t)\le1/r+t\) and \(\phi''(t)\ge r^2/(1+rt)^2\); substituting
\(u=1+rt\) in Taylor's formula gives
\[
 D_F(y,z)
 =\int_0^1(1-t)\phi''(t)\,dt
 \ge\int_1^{1+r}\frac{r+1-u}{u^2}\,du
 =r-\log(1+r)=\omega(r). \qedhere
\]
\end{proof}

\subsection{Cubic floor, QIX moments, and scale jumps}
\label{sec:qix-moments}

\begin{lemma}[Cubic-floor properties]
\label{lem:cubic-support}
For \(\kappa\ge1\) and \(h\ge0\), the floor
\(\rho_\kappa(h)=(\kappa^{3/2}+2\kappa h)^{1/3}\) is increasing in \(h\),
satisfies \(\rho_\kappa(0)=\sqrt\kappa\),
\begin{equation}
 \max\bigl\{\sqrt\kappa,\,(2\kappa h)^{1/3}\bigr\}
 \le\rho_\kappa(h)
 \le\sqrt\kappa+(2\kappa h)^{1/3},
 \label{eq:rho-bounds}
\end{equation}
and obeys \(\rho_\kappa(h)^3\ge\kappa(\rho_\kappa(h)+h)\).
Algorithm~\ref{alg:radial-qix} satisfies, pathwise,
\begin{equation}
 \beta_t^3\ge\kappa_{j_t}(\beta_t+\twonorm{\widehat\Theta_{t-1}})
 \ge\kappa_0(\beta_t+\twonorm{\widehat\Theta_{t-1}}),
 \qquad
 \beta_t^2p_t\ge\kappa_{j_t}\ge\kappa_0.
 \label{eq:cubic-support}
\end{equation}
\end{lemma}

\begin{proof}
Monotonicity, \(\rho_\kappa(0)=\sqrt\kappa\), and \eqref{eq:rho-bounds}
follow immediately from the definition (using \((a+b)^{1/3}\le a^{1/3}+b^{1/3}\)
for the upper bound). For the cubic inequality, expanding
\((\sqrt\kappa+h)^3=\kappa^{3/2}+3\kappa h+3\sqrt\kappa\,h^2+h^3\ge\kappa^{3/2}+2\kappa h\)
and taking cube roots gives \(\rho_\kappa(h)\le\sqrt\kappa+h\), whence
\(\kappa(\rho_\kappa(h)+h)\le\kappa^{3/2}+2\kappa h=\rho_\kappa(h)^3\).

On \([\sqrt\kappa,\infty)\), \(r(b)=b^3-\kappa b-\kappa h\) has
derivative \(r'(b)=3b^2-\kappa\ge2\kappa>0\). Since the scale rule gives
\(\beta_t\ge\rho_{\kappa_{j_t}}(\twonorm{\widehat\Theta_{t-1}})\ge\sqrt{\kappa_{j_t}}\),
monotonicity gives
\(r(\beta_t)\ge r(\rho_{\kappa_{j_t}}(\twonorm{\widehat\Theta_{t-1}}))\ge0\),
proving the first bound in \eqref{eq:cubic-support} (with
\(\kappa_{j_t}\ge\kappa_0\)). Dividing by
\(\beta_t+\twonorm{\widehat\Theta_{t-1}}>0\) and using
\(p_t=\beta_t/(\twonorm{\widehat\Theta_{t-1}}+\beta_t)\) gives
\(\beta_t^2p_t\ge\kappa_{j_t}\ge\kappa_0\).
\end{proof}

Applying Proposition~\ref{prop:qix-boundary-release} conditionally at
\((e_t,p_t,G_t,\theta_t,\nu_t)\) gives the predictable bias coefficient
\(\lambda_t\in[0,1]\) of \eqref{eq:qix-lambda}. Using
\(p_t+\nu_t\iprod{U}{G_t^{1/2}\theta_t}^2\ge p_t\) in \eqref{eq:qix-lambda},
Lemma~\ref{lem:spherical-polynomial-moments}(b), and
\(\nu_t=d^2\ell_{j_t}^2/(4\beta_t^2)\) yields the pathwise bound (trivial
when \(\theta_t=0\))
\begin{equation}
 \lambda_t
 \le\frac{d\nu_t}{p_t\normt{G_t}{\theta_t}^2}\E_U\iprod{U}{G_t^{1/2}\theta_t}^4
 =\frac{3\nu_t\normt{G_t}{\theta_t}^2}{p_t(d+2)}
 =\frac{3d^2\ell_{j_t}^2\normt{G_t}{\theta_t}^2}{4\beta_t^2p_t(d+2)}.
 \label{eq:qix-lambda-bound}
\end{equation}
Resolving the bias into radial and tangential components gives
\(\E_t\widehat{\theta}_t=(1-\lambda_t)P_{\perp,t}\theta_t+(1-\lambda_tp_t)P_{\parallel,t}\theta_t\);
near the boundary, radial shrinkage is tiny whereas tangential shrinkage
can be larger, leaving the radial drift to account for in
Appendix~\ref{sec:radial-growth}.

\begin{proofof}{Lemma~\ref{lem:qix}.}
Write \(\chi_t=\iprod{e_t}{\theta_t}\) within this proof.

\emph{Bias.}
Proposition~\ref{prop:qix-boundary-release}(c), applied conditionally at
\((e_t,p_t,G_t,\theta_t,\nu_t)\), gives Lemma~\ref{lem:qix:bias} with
\(\lambda_t\in[0,1]\).

\emph{Covariance.}
For the unshrunk estimator \(\widehat{\theta}_t^0\) of
Proposition~\ref{prop:qix-boundary-release}, \(\E_t\widehat{\theta}_t^0=\theta_t\).
Writing \(\theta_t=w_t+\chi_te_t\) with \(w_t\perp e_t\) and
\(\alpha_d=2d/(d+2)\ge1\ge p_t\), Lemma~\ref{lem:spherical-polynomial-moments}(a)--(b)
shows that \(\bigl(\tfrac{3d}{d+2}-p_t\bigr)\normt{G_t}{\theta_t}^2G_t/p_t
 -\operatorname{Cov}_t(\widehat{\theta}_t^0)\) is a nonnegative multiple of
the identity on \(\{w_t,e_t\}^\perp\), and on \(\operatorname{span}\{w_t,e_t\}\)
has matrix \(\bigl(\begin{smallmatrix}A&B\\B&C\end{smallmatrix}\bigr)\) with
\(A=(\alpha_d-p_t)\chi_t^2\ge0\), \(B=-(\alpha_d-1)\twonorm{w_t}\chi_t\),
\(C=(\alpha_d-p_t)\twonorm{w_t}^2+p_t(1-p_t)\chi_t^2\ge0\), and
\(AC-B^2=\twonorm{w_t}^2\chi_t^2\{(\alpha_d-p_t)^2-(\alpha_d-1)^2\}
 +(\alpha_d-p_t)p_t(1-p_t)\chi_t^4\ge0\). Thus
\(\operatorname{Cov}_t(\widehat{\theta}_t^0)\preceq(3d/(d+2)-p_t)\normt{G_t}{\theta_t}^2G_t/p_t\).

QIX shrinks exploratory draws by \(p_t/(p_t+\nu_tY_t^2)\in[0,1]\), so
\(\E_t[\widehat\theta_t\widehat\theta_t\transpose]\preceq\E_t[\widehat\theta_t^0(\widehat\theta_t^0)\transpose]\).
By Lemma~\ref{lem:qix:bias},
\[
 \operatorname{Cov}_t(\widehat\theta_t)-\operatorname{Cov}_t(\widehat\theta_t^0)
 \preceq\theta_t\theta_t\transpose-(\E_t\widehat\theta_t)(\E_t\widehat\theta_t)\transpose
 \preceq\lambda_t(G_t\theta_t\theta_t\transpose+\theta_t\theta_t\transpose G_t).
\]
Since \(\normt{G_t^{-1}}{\theta_t}^2=\twonorm{w_t}^2+\chi_t^2/p_t\le\normt{G_t}{\theta_t}^2/p_t^2\),
Cauchy--Schwarz gives \(|\iprod{v}{G_t\theta_t}|\le\normt{G_t}{v}\normt{G_t}{\theta_t}\)
and \(|\iprod{v}{\theta_t}|\le\normt{G_t}{v}\normt{G_t}{\theta_t}/p_t\), so
\(\lambda_t(G_t\theta_t\theta_t\transpose+\theta_t\theta_t\transpose G_t)
 \preceq2\lambda_t\normt{G_t}{\theta_t}^2G_t/p_t\). Since \(\lambda_t\le1\)
and \(3d/(d+2)<3\), this yields
\(\operatorname{Cov}_t(\widehat\theta_t)\preceq5\normt{G_t}{\theta_t}^2G_t/p_t\),
proving Lemma~\ref{lem:qix:cov}.

\emph{Range.}
On an exploratory draw, \eqref{eq:qix-estimator},
\((p_t-\nu_tY_t^2)^2\ge0\), and \(\nu_t=d^2\ell_{j_t}^2/(4\beta_t^2)\) give
\(\normt{p_tG_t^{-1}}{\widehat\theta_t}^2=d^2p_tY_t^2/(p_t+\nu_tY_t^2)^2
 \le d^2/(4\nu_t)=\beta_t^2/\ell_{j_t}^2\). On a boundary draw,
\(\normt{p_tG_t^{-1}}{\widehat\theta_t}=|\chi_t|\le1\le\beta_t/\ell_{j_t}\)
(by \eqref{eq:base-scales}). Since \(0\preceq I_d-\lambda_tG_t\preceq I_d\) and
\(p_tG_t^{-1}\preceq I_d\), Lemma~\ref{lem:qix:bias} gives
\(\normt{p_tG_t^{-1}}{\E_t\widehat\theta_t}\le1\), hence
\(\normt{p_tG_t^{-1}}{\xi_t}\le\beta_t/\ell_{j_t}+1\). Finally,
\(\{\nabla^2F(z_t)\}^{-1}=p_tG_t\preceq p_tG_t^{-1}\) gives
\(\sqrt{p_t}\normt{G_t}{\widehat\theta_t}\le\beta_t/\ell_{j_t}\), proving
Lemmas~\ref{lem:qix:range} and~\ref{lem:qix:local}.

\emph{Expected squared local norm.}
Lemma~\ref{lem:spherical-polynomial-moments}(b), paired with \(G_t^2\), gives
\[
 \E_t[p_t\normt{G_t}{\widehat\theta_t^0}^2]
 =(1-p_t)p_t^2\chi_t^2
 +\frac d{d+2}\bigl\{
 \normt{G_t}{\theta_t}^2\tr(G_t^2)+2\theta_t\transpose G_t^3\theta_t
 \bigr\}.
\]
The coefficients of \(\twonorm{w_t}^2\) and \(p_t\chi_t^2\) lie between
\(d(d-1)/(d+2)\) and \(d\), so
\[
 \frac{d(d-1)}{d+2}\normt{G_t}{\theta_t}^2
 \le\E_t[p_t\normt{G_t}{\widehat\theta_t^0}^2]
 \le d\normt{G_t}{\theta_t}^2.
\]
Only the exploratory branch is shrunk, and
\(1-\{p_t/(p_t+\nu_tY_t^2)\}^2\le2\nu_tY_t^2/p_t\). Applying
Lemma~\ref{lem:spherical-polynomial-moments}(c) with \(G_t^{1/2}\theta_t\)
and \(G_t^2\), and substituting \(\nu_t=d^2\ell_{j_t}^2/(4\beta_t^2)\), gives
\begin{align*}
 \E_t(p_t\normt{G_t}{\widehat\theta_t^0}^2-p_t\normt{G_t}{\widehat\theta_t}^2)
 &\le\frac{2d^2\nu_t}{p_t}\E_U\bigl[Y_t^4U\transpose G_t^2U\bigr]\\
 &=\frac{3d^3\ell_{j_t}^2\normt{G_t}{\theta_t}^2
 \{\normt{G_t}{\theta_t}^2\tr(G_t^2)+4\theta_t\transpose G_t^3\theta_t\}}
 {2\beta_t^2p_t(d+2)(d+4)}.
\end{align*}

\emph{Uniform lower bound after shrinkage: pure orientations.}
We compare the shrinkage bound with the unshrunk margin above
\((d/8)\normt{G_t}{\theta_t}^2\), writing \(\ell=\ell_{j_t}\).
For a unit tangent loss, the unshrunk margin is
\(d(7d+6+8p_t^2)/\{8(d+2)\}\), while at \(\beta_t^2p_t\ge\kappa\) the
shrinkage bound is \(3d^3\ell^2(d+3+p_t^2)/\{2\kappa(d+2)(d+4)\}\).
Thus it suffices that \(\kappa\ge\kappa_\perp(p_t)\), where
\[
 \kappa_\perp(p_t)
 =\frac{12d^2\ell^2(d+3+p_t^2)}
 {(d+4)(7d+6+8p_t^2)}.
\]
Since \((7d+6)-8(d+3)=-d-18<0\), \(\kappa_\perp(p_t)\) decreases in \(p_t^2\),
with maximum
\(\kappa_\perp(0)=12d^2\ell^2(d+3)/\{(d+4)(7d+6)\}=(\ell/\ell_0)^2\kappa_0\).
For a unit radial loss (discarding one factor \(p_t\le1\)), it suffices that
\(\kappa\) exceed
\[
 \frac{3d^3\ell^2(d-1+5p_t^2)}
 {2(d+2)(d+4)
 \bigl\{p_t(1-p_t)+\frac d{d+2}(d-1+3p_t^2)-\frac d8\bigr\}}.
\]
This expression is at most \(\kappa_\perp(0)\): clearing positive
denominators yields
\(12d(d-2)+8(d+3)(d+2)p_t -(19d^2-2d+48)p_t^2\ge0\), which is concave in
\(p_t\) and takes nonnegative values \(12d(d-2)\) and \(d(d+18)\) at the
endpoints \(p_t\in\{0,1\}\) for \(d\ge2\).

\emph{Arbitrary losses.}
For a general unit loss, put
\(x=\twonorm{w_t}^2\), so \(\chi_t^2=1-x\).  The unshrunk margin \(D(x)\)
above \((d/8)\normt{G_t}{\theta_t}^2\) is affine and positive on \([0,1]\),
whereas the numerator
\(N(x)=\normt{G_t}{\theta_t}^2\{\normt{G_t}{\theta_t}^2\tr(G_t^2)+4\theta_t\transpose G_t^3\theta_t\}\)
is convex in \(x\) (a positive sum of a squared affine function and a
product of two affine functions with nonnegative slopes). By convexity and
positivity,
\(N(x)\le(1-x)N(0)+xN(1)\le D(x)\max\{N(0)/D(0),N(1)/D(1)\}\), so the
shrinkage-to-margin ratio is bounded by its endpoint values.
Scaling \(\twonorm{\theta_t}\) below one only
decreases this ratio. Hence the pure orientations
cover every \(\theta_t\in\ball\). Lemma~\ref{lem:cubic-support}
and \(\kappa_{j_t}=\max\{1,(j_t+1)/d\}\kappa_\perp(0)\ge\kappa_\perp(0)\) give
\(\E_t[p_t\normt{G_t}{\widehat\theta_t}^2]\ge d\normt{G_t}{\theta_t}^2/8\),
and the upper bound follows from the unshrunk second moment, proving
Lemma~\ref{lem:qix:sandwich}.
\end{proofof}

\begin{lemma}[Scale-ratio bound]
\label{lem:rate-jumps}
Pathwise, \(\beta_{t+1}\le3\beta_t\) for every \(t\ge0\). In particular,
on epoch \(I_j\), \(j_-\le j_t\le j\), so \(\ell_{j_-}\le\ell_{j_t}\le\ell_j\).
\end{lemma}

\begin{proof}
We proceed by induction on \(t\ge0\). For \(t=0\),
\(\beta_1=\beta_0\le3\beta_0\). Assume \(t\ge1\) and
\(\beta_t\le3\beta_{t-1}\); then
\(j_{t+1}=j(\beta_t)\le j(\beta_{t-1})+2=j_t+2\). We bound each of the
three candidates in the rule for \(\beta_{t+1}\) by \(3\beta_t\) (the
inherited scale \(\beta_t\) is immediate).

\emph{Second-order candidate.}
Monotonicity gives \(a_{j_{t+1}}\le a_{j_t}\). The round-\(t\) rule gives
\(a_{j_t}\sqrt{\EestObs_{t-1}+1}\le\beta_t\), and
\(p_t\normt{G_t}{\widehat\theta_t}^2\le\beta_t^2/\ell_{j_t}^2\) by
Lemma~\ref{lem:qix:local}, so
\[
 a_{j_{t+1}}\sqrt{\EestObs_t+1}
 \le a_{j_t}\sqrt{\EestObs_{t-1}+1+p_t\normt{G_t}{\widehat\theta_t}^2}
 \le\beta_t\sqrt{1+(a_0/\ell_0)^2}<2\beta_t.
\]

\emph{Cubic-floor candidate.}
Since \(p_tG_t^{-1}\succeq p_tI_d\), Lemma~\ref{lem:qix:range} and
\(\beta_t^2p_t\ge\kappa_{j_t}\) (by \eqref{eq:cubic-support}) give
\(\twonorm{\widehat\theta_t}\le\beta_t/(\ell_{j_t}\sqrt{p_t})\le\beta_t^2/(\ell_{j_t}\sqrt{\kappa_{j_t}})\),
so \(\twonorm{\widehat\Theta_t}\le\twonorm{\widehat\Theta_{t-1}}+\beta_t^2/(\ell_{j_t}\sqrt{\kappa_{j_t}})\).
By definition of the floor, \(\beta_t\ge\rho_{\kappa_{j_t}}(\twonorm{\widehat\Theta_{t-1}})\)
means \(2\kappa_{j_t}\twonorm{\widehat\Theta_{t-1}}\le\beta_t^3-\kappa_{j_t}^{3/2}\).
Since \(j_{t+1}\le j_t+2\) and \(d\ge2\),
\(\max\{d,j_{t+1}+1\}/\max\{d,j_t+1\}\le2\) and
\(R:=\kappa_{j_{t+1}}/\kappa_{j_t}\le2(1+2\log3/\ell_0)^2<7\).
With \(x=\beta_t/\sqrt{\kappa_{j_t}}\ge1\) and \(\ell_{j_t}\ge\ell_0>2.577\),
\[
 \frac{\rho_{\kappa_{j_{t+1}}}(\twonorm{\widehat\Theta_t})^3}{\beta_t^3}
 =\frac{\kappa_{j_{t+1}}^{3/2}+2\kappa_{j_{t+1}}\twonorm{\widehat\Theta_t}}{\beta_t^3}
 \le\frac{R^{3/2}}{x^3}+R\Bigl(1-\frac1{x^3}+\frac{2}{\ell_{j_t}x}\Bigr)
 \le R^{3/2}+\frac{2R}{\ell_0}
 <7^{3/2}+\frac{14}{2.577}<24<3^3,
\]
where the second inequality holds because \(R^{3/2}>R\) makes the
expression decreasing in \(x\ge1\). Taking cube roots yields
\(\rho_{\kappa_{j_{t+1}}}(\twonorm{\widehat\Theta_t})\le3\beta_t\).
This completes the induction; the epoch consequence follows immediately
from \(B_j\le\beta_t<2B_j\) and \(\beta_{t-1}\ge\beta_t/3\).
\end{proof}

\subsection{FTRL on effective losses, stability, and regularisation}
\label{sec:decomposition}

With \(g_t=\widehat\Theta_t-\widehat\Theta_{t-1}\),
\(z_t=\nabla\Psi_{\beta_t}(-\widehat\Theta_{t-1})\) from
Proposition~\ref{prop:qix-boundary-release}(a), and the Bregman divergence
\(D_F(y,z)=F(y)-F(z)-\iprod{\nabla F(z)}{y-z}\) (and similarly
\(D_{\Psi_\beta}\)), evenness of \(\Psi_\beta\) gives
\begin{equation}
 \mathrm{stab}_t=D_{\Psi_{\beta_t}}(-\widehat\Theta_t,-\widehat\Theta_{t-1})
 =\iprod{z_t}{g_t}+\twonorm{\widehat\Theta_t}-\twonorm{\widehat\Theta_{t-1}}
 -\beta_t\log\frac{\beta_t+\twonorm{\widehat\Theta_t}}{\beta_t+\twonorm{\widehat\Theta_{t-1}}}.
 \label{eq:stability-explicit}
\end{equation}
For fixed \(h\ge0\), \(b\log(1+h/b)\) is nondecreasing in \(b>0\) (its
derivative is \(\log(1+h/b)-h/(b+h)\ge0\)), so \(\Psi_\beta(y)\) is
nonincreasing in \(\beta\). Instead of the vacuous Fenchel--Young bound at
unit comparators (where \(F(v)=+\infty\)), we use
\(\twonorm{\widehat\Theta_T}+\iprod v{\widehat\Theta_T}=\Gamma_T(v)\ge0\).

\begin{lemma}[FTRL on effective losses]
\label{lem:effective-ftrl}
For every unit comparator \(v\),
\begin{equation}
 \sum_{t=1}^T\iprod{z_t-v}{g_t}
 =\sum_{t=1}^T\beta_t\log
 \frac{\beta_t+\twonorm{\widehat\Theta_t}}{\beta_t+\twonorm{\widehat\Theta_{t-1}}}
 +\sum_{t=1}^T\mathrm{stab}_t-\Gamma_T(v).
 \label{eq:effective-ftrl}
\end{equation}
If \(\beta_t\) is nondecreasing, the first term is at most \(\beta_T\log(1+\twonorm{\widehat\Theta_T}/\beta_T)\).
\end{lemma}

\begin{proof}
The proof is the potential argument of \citet[Lemma~2]{ALST14}.
Since \(\Psi_\beta\) is even, \eqref{eq:stability-explicit} rearranges to
the standard FTRL step
\(\iprod{z_t}{g_t}=\Psi_{\beta_t}(\widehat\Theta_{t-1})-\Psi_{\beta_t}(\widehat\Theta_t)+\mathrm{stab}_t\).
Summed from \(\widehat\Theta_0=0\), the norm parts of the potentials telescope to
\(-\twonorm{\widehat\Theta_T}\) and their logarithmic parts give the logarithmic sum.
The comparator contributes \(-\sum_t\iprod v{g_t}=-\iprod v{\widehat\Theta_T}\), and
\(-\twonorm{\widehat\Theta_T}-\iprod v{\widehat\Theta_T}=-\Gamma_T(v)\). This proves
\eqref{eq:effective-ftrl}.

For the logarithmic sum, regroup the potentials by state:
\[
 \sum_{t=1}^T\{\Psi_{\beta_t}(\widehat\Theta_{t-1})-\Psi_{\beta_t}(\widehat\Theta_t)\}
 =-\Psi_{\beta_T}(\widehat\Theta_T)
 +\sum_{t=1}^{T-1}\{\Psi_{\beta_{t+1}}(\widehat\Theta_t)-\Psi_{\beta_t}(\widehat\Theta_t)\}
 \le-\Psi_{\beta_T}(\widehat\Theta_T),
\]
using \(\Psi_{\beta_1}(\widehat\Theta_0)=0\) and the monotonicity of \(\Psi_\beta\) in
\(\beta\). The left side is \(-\twonorm{\widehat\Theta_T}\) plus the logarithmic sum,
and
\(-\Psi_{\beta_T}(\widehat\Theta_T)=-\twonorm{\widehat\Theta_T}+\beta_T\log(1+\twonorm{\widehat\Theta_T}/\beta_T)\).
\end{proof}

Since \(\widehat\theta_t=g_t+\epsilon_t\widehat\Theta_{t-1}\),
\(\widehat\Theta_{t-1}=-\twonorm{\widehat\Theta_{t-1}}e_t\),
\(z_t=(1-p_t)e_t\), and \(p_t\twonorm{\widehat\Theta_{t-1}}=\beta_t(1-p_t)\),
\begin{equation}
 \epsilon_t\iprod{z_t-v}{\widehat\Theta_{t-1}}
 =\epsilon_t\twonorm{\widehat\Theta_{t-1}}\{\iprod v{e_t}-1+p_t\}
 =-\epsilon_t\Gamma_{t-1}(v)+\epsilon_t\beta_t(1-p_t),
 \label{eq:contraction-ledger}
\end{equation}
matching the linear bias term of \citet[Theorem~1]{zimmert2022bias} with
bias \(\epsilon_t\widehat\Theta_{t-1}=-\epsilon_t\beta_t\nabla F(z_t)\).

\begin{proofof}{Lemma~\ref{lem:contracted-ftrl}.}
Sum \eqref{eq:contraction-ledger} over \(t\) and add it to
\eqref{eq:effective-ftrl}.
\end{proofof}

\paragraph{Stability of FTRL on effective losses.}

By Fenchel duality between \(\Psi_{\beta_t}\) and \(\beta_tF\), the
stability cost has the variational form
\[
 \mathrm{stab}_t=\sup_{\twonorm z<1}
 \left\{\iprod{z_t-z}{g_t}-\beta_tD_F(z,z_t)\right\}.
\]
The radial barrier is self-concordant, so
\(D_F(z,z_t)\ge\omega(\normt{\nabla^2F(z_t)}{z-z_t})\) with
\(\omega(r)=r-\log(1+r)\) (Lemma~\ref{lem:radial-barrier-bregman}). Bounding the inner product by the
dual local norm and using the conjugate \(\omega_*(s)=-s-\log(1-s)\) of
\(\omega\) gives the usual local-norm bound for barrier FTRL:
\begin{equation}
 \mathrm{stab}_t\le\beta_t\,\omega_*\!\left(
 \frac{\normt{\nabla^2F(z_t)^{-1}}{g_t}}{\beta_t}\right)
 \qquad\text{when }\normt{\nabla^2F(z_t)^{-1}}{g_t}<\beta_t.
 \label{eq:local-norm-stability}
\end{equation}

The estimate has local norm at most \(\beta_t/\ell_{j_t}\le\beta_t/\ell_0\),
with \(1/\ell_0<0.4\), and a contraction adds at most \(\beta_t/4\), so the
combined local norm stays below \(\beta_t\).

\begin{lemma}[One-round FTRL stability]
\label{lem:ftrl-stability}
Pathwise,
\begin{equation}
 \mathrm{stab}_t
 \le\frac{p_t\normt{G_t}{\widehat\theta_t}^2}{\beta_t}
 +\frac34\II{\epsilon_t>0}\beta_t(1-p_t).
 \label{eq:ftrl-stability}
\end{equation}
\end{lemma}

\begin{proof}
By \eqref{eq:inverse-hessian}, Lemma~\ref{lem:qix:local}, and
\(\ell_{j_t}\ge\ell_0\),
\[
 s_t:=\frac{\normt{\nabla^2F(z_t)^{-1}}{\widehat{\theta}_t}}{\beta_t}
 =\frac{\sqrt{p_t}\normt{G_t}{\widehat\theta_t}}{\beta_t}
 \le\frac1{\ell_{j_t}}\le\frac1{\ell_0}.
\]
Since \(\widehat\Theta_{t-1}=-\twonorm{\widehat\Theta_{t-1}}e_t\),
\(G_te_t=p_te_t\), and \(p_t\twonorm{\widehat\Theta_{t-1}}/\beta_t=1-p_t\)
by \eqref{eq:natural-depth}, the contraction displacement has local norm
\[
 \frac{\normt{\nabla^2F(z_t)^{-1}}{\epsilon_t\widehat\Theta_{t-1}}}{\beta_t}
 =\frac{\epsilon_t\twonorm{\widehat\Theta_{t-1}}\sqrt{e_t\transpose(p_tG_t)e_t}}{\beta_t}
 =\frac{\epsilon_tp_t\twonorm{\widehat\Theta_{t-1}}}{\beta_t}
 =\epsilon_t(1-p_t).
\]
Off a trigger, \(g_t=\widehat\theta_t\), and
\eqref{eq:local-norm-stability} bounds the stability cost by
\(\beta_t\omega_*(s_t)\). Since \(\omega_*(s)=\sum_{k\ge2}s^k/k\) and
\(0\le s_t\le1/\ell_0<0.4\),
\[
 \beta_t\omega_*(s_t)
 =\beta_ts_t^2\sum_{k\ge2}\frac{s_t^{k-2}}k
 \le\beta_ts_t^2\left(\frac12+\frac{0.4}{3\cdot0.6}\right)
 \le\beta_ts_t^2
 =\frac{p_t\normt{G_t}{\widehat\theta_t}^2}{\beta_t}.
\]

On a trigger, \(g_t=\widehat\theta_t-\epsilon_t\widehat\Theta_{t-1}\), and the
triangle inequality increases the argument of \(\omega_*\) by at most
\(\epsilon_t(1-p_t)\). Since \(\omega_*'(s)=s/(1-s)\) is nondecreasing and
\(s_t+\epsilon_t(1-p_t)\le1/\ell_0+1/4<0.65\), the mean-value theorem gives
\[
 \beta_t\bigl\{\omega_*(s_t+\epsilon_t(1-p_t))-\omega_*(s_t)\bigr\}
 \le\beta_t\epsilon_t(1-p_t)\omega_*'(0.65)
 \le\frac{\beta_t(1-p_t)}4\,\frac{0.65}{0.35}
 <\frac34\beta_t(1-p_t).
\]
Adding the two bounds proves \eqref{eq:ftrl-stability}.
\end{proof}

Adding \eqref{eq:ftrl-stability} to Lemma~\ref{lem:contracted-ftrl},
bounding the logarithmic sum by Lemma~\ref{lem:effective-ftrl}, and
dropping \(-\Gamma_T(v)\le0\) gives the pathwise FTRL bound: for every
unit comparator \(v\),
\begin{equation}
 \sum_{t=1}^T\iprod{z_t-v}{\widehat{\theta}_t}
 \le\beta_T\log\Bigl(1+\frac{\twonorm{\widehat\Theta_T}}{\beta_T}\Bigr)
 +\sum_{t=1}^T\frac{p_t\normt{G_t}{\widehat\theta_t}^2}{\beta_t}
 -\sum_{t=1}^T\epsilon_t\Gamma_{t-1}(v)
 +\sum_{t:\,\epsilon_t>0}\beta_t(1-p_t).
 \label{eq:ftrl-pathwise}
\end{equation}
Each contraction costs at most \(\frac34\beta_t(1-p_t)\) in stability and
\(\epsilon_t\beta_t(1-p_t)=\frac14\beta_t(1-p_t)\) in
\eqref{eq:contraction-ledger}, a total of \(\beta_t(1-p_t)\).

\Needspace{8\baselineskip}
\paragraph{Terminal regularisation cost.}

The FTRL identity leaves a logarithmic term in the terminal state,
which we bound through \(\beta_T\).

\begin{lemma}[Terminal regularisation cost]
\label{lem:terminal-cost}
Pathwise, the terminal state satisfies
\(\twonorm{\widehat\Theta_T}\le\beta_T^3/\kappa_0-(1-1/\ell_0)\beta_T\).
The logarithmic contribution is bounded by
\begin{equation}
 \sum_{t=1}^T\beta_t\log
 \frac{\beta_t+\twonorm{\widehat\Theta_t}}{\beta_t+\twonorm{\widehat\Theta_{t-1}}}
 \le\beta_T\log\left(1+\frac{\twonorm{\widehat\Theta_T}}{\beta_T}\right)
 \le\beta_T\log(\beta_T^2/\kappa_0+1/\ell_0).
 \label{eq:terminal-cost}
\end{equation}
\end{lemma}

\begin{proof}
\emph{Terminal state.}
The cubic-floor condition \eqref{eq:cubic-support} at round \(T\) gives
\(\twonorm{\widehat\Theta_{T-1}}\le\beta_T^3/\kappa_0-\beta_T\).
On an exploratory draw write \(\widehat{\theta}_T=kG_T^{1/2}U_T\), set
\(r=r_T=\twonorm{\widehat\Theta_{T-1}}/\beta_T=(1-p_T)/p_T\), and let
\(x=|\iprod{e_T}{U_T}|\). The QIX range and \(\ell_{j_T}\ge\ell_0\) give
\(|k|\le\beta_T/(\ell_0\sqrt{p_T})\). Since \(0\le\epsilon_T\le1/4\),
the radial component of the updated state has absolute value at most
\(\twonorm{\widehat\Theta_{T-1}}+|k|\sqrt{p_T}x\); its orthogonal component has squared norm
\(k^2(1-x^2)\). Dividing by \(\beta_T^2\), and using
\(\abs k\sqrt{p_T}/\beta_T\le1/\ell_0\) and
\(k^2/\beta_T^2\le1/(\ell_0^2p_T)=(1+r)/\ell_0^2\), gives
\[
 \frac{\twonorm{\widehat\Theta_T}^2}{\beta_T^2}
 \le (r+x/\ell_0)^2+\ell_0^{-2}(1+r)(1-x^2)
 \le (r+1/\ell_0)^2,
\]
because
\((r+1/\ell_0)^2-(r+x/\ell_0)^2-\ell_0^{-2}(1+r)(1-x^2)
 =\frac{r}{\ell_0}(1-x)\{2-(1+x)/\ell_0\}\ge0\)
for \(x\in[0,1]\) and \(\ell_0>2\). On a boundary draw the update
is radial and \(|Y_T|\le1\le\beta_T/\ell_0\), so the same conclusion holds.
In either branch, taking square roots and applying the round-\(T\) cubic
floor gives
\[
 \twonorm{\widehat\Theta_T}
 \le\twonorm{\widehat\Theta_{T-1}}+\frac{\beta_T}{\ell_0}
 \le\frac{\beta_T^3}{\kappa_0}-\Bigl(1-\frac1{\ell_0}\Bigr)\beta_T.
\]

\emph{Logarithmic contribution.}
Substituting the terminal state bound into Lemma~\ref{lem:effective-ftrl}
gives \(1+\twonorm{\widehat\Theta_T}/\beta_T\le\beta_T^2/\kappa_0+1/\ell_0\),
proving \eqref{eq:terminal-cost}.
\end{proof}

\section{Contractive Hilbert-space concentration and martingale tools}
\label{sec:tools}
\label{sec:concentration}

We collect the concentration tools used throughout the analysis: a scalar
Freedman bound for masked increments on a deterministic grid, and the
Bennett-type bound for Hilbert-space recursions with predictable
contractions (Lemma~\ref{lem:contractive-dilation} and
Proposition~\ref{prop:contractive-classical} of
Section~\ref{sec:contractive-concentration}), derived from
\citet[Theorem~3.4]{pinelis1994optimum} for \((2,D)\)-smooth spaces with
\(D=1\).

\subsection{Scalar Bennett and masked Freedman inequalities}

Write \(\mathfrak h(u)=(1+u)\log(1+u)-u\) for the Bennett rate function
\citep{bennett1962probability}. It is strictly increasing on
\([0,\infty)\); write \(\mathfrak h^{-1}\) for its inverse.

\begin{lemma}[Elementary Bennett inverse bound]
\label{lem:bennett-comparisons}
For every \(z\ge0\),
\begin{equation}
 \mathfrak h^{-1}(z)\le\sqrt{2z}+z/3.
 \label{eq:bennett-inverse-bound}
\end{equation}
\end{lemma}

\begin{proof}
Put \(a=\sqrt{2z}\) and
\(g(a)=\log(1+a+a^2/6)-3a/(a+3)\). Direct differentiation gives
\[
 g(0)=0,\qquad
 g'(a)=\frac{a^2(2a+9)}{(a^2+6a+6)(a+3)^2}\ge0
 \quad(a\ge0).
\]
Consequently,
\[
 \frac d{da}\left\{\mathfrak h(a+a^2/6)-\frac{a^2}{2}\right\}
 =(1+a/3)g(a)\ge0.
\]
The expression in braces is zero at \(a=0\), so
\(\mathfrak h(\sqrt{2z}+z/3)\ge z\). Monotonicity of
\(\mathfrak h\) proves \eqref{eq:bennett-inverse-bound}.
\end{proof}


\phantomsection
\label{sec:masked-concentration}

The sampling-noise and radial proofs use the following masked concentration
bound.
A predictable mask bounds the cumulative variance proxy by a deterministic level.
We use the exponential argument of \citet{bennett1962probability} and
\citet[Lemma~1.3(a) and Theorem~1.6]{freedman1975} for martingales (see also
\citealp{fan2012large}), together with Ville's maximal inequality
\citep{ville1939etude}. A union bound over a deterministic geometric grid
\citep[cf.][]{howard2021time} then lets us select its least level covering
the realised proxy.
Let \((X_t)_{t\le T}\) be adapted with \(\E_tX_t=0\), let \(k_t\in\{0,1\}\) be
predictable with \(|k_tX_t|\le R\) almost surely for some \(R>0\), and let
\(\overline\sigma_t^2\) be predictable with \(\overline\sigma_t^2\ge\E_tX_t^2\)
whenever \(k_t=1\). For a deterministic variance bound \(v>0\), define the
predictable mask
\[
 m_t^{(v)}
 =k_t\,\II*{\sum_{s<t}m_s^{(v)}\overline\sigma_s^2
 +\overline\sigma_t^2\le v}.
\]

\begin{lemma}[Masked Freedman inequality]
\label{lem:masked-freedman}
For every \(\ell>0\), with probability at least \(1-e^{-\ell}\),
simultaneously for every prefix \(\tau\le T\),
\begin{equation}
 \sum_{t\le\tau}m_t^{(v)}X_t
 \le\sqrt{2v\ell}+\frac R3\ell;
 \label{eq:masked-freedman}
\end{equation}
the two-sided bound, for \(\pm X_t\), fails with probability at most
\(2e^{-\ell}\). Whenever \(\sum_{t\le\tau}k_t\overline\sigma_t^2\le v\),
every kept round through \(\tau\) has \(m_t^{(v)}=k_t\), so
\eqref{eq:masked-freedman} holds for \(\sum_{t\le\tau}k_tX_t\).
\end{lemma}

\begin{proof}
\emph{Step 1: control the masked exponential process.}
Predictability of \(k_t\) makes \((k_tX_t)\) a martingale-difference
sequence.  The mask is predictable, so the selected increments
\(m_t^{(v)}X_t\) remain centred, have range
\(R\), and total predictable variance at most \(v\), since
\(m_t^{(v)}\le k_t\) and the hypotheses hold on kept rounds.  Write
\(S_t=\sum_{s\le t}m_s^{(v)}X_s\) and
\(V_t=\sum_{s\le t}m_s^{(v)}\overline\sigma_s^2\le v\).
For every fixed \(\lambda\ge0\), the conditional Bennett bound
\citep[Lemma~1.3(a)]{freedman1975} makes
\[
 Z_t(\lambda)
 =\exp\!\left\{\lambda S_t
 -\frac{e^{R\lambda}-1-R\lambda}{R^2}V_t\right\}
\]
a nonnegative supermartingale with \(Z_0(\lambda)=1\).

\emph{Step 2: invert the maximal tail bound.}
For deterministic \(x>0\), Ville's inequality and \(V_t\le v\) give
\[
 \bbP\{\exists\tau\le T:S_\tau\ge x\}
 \le\inf_{\lambda>0}
 \exp\!\left\{-\lambda x+
 \frac v{R^2}(e^{R\lambda}-1-R\lambda)\right\}
 =\exp\!\left\{-\frac v{R^2}\mathfrak h(Rx/v)\right\},
\]
where the infimum is attained at \(R\lambda=\log(1+Rx/v)\).
Now take \(x=(v/R)\mathfrak h^{-1}(R^2\ell/v)\), so that the exponent
equals \(-\ell\); Lemma~\ref{lem:bennett-comparisons} with
\(z=R^2\ell/v\) gives \(x\le\sqrt{2v\ell}+R\ell/3\), proving
\eqref{eq:masked-freedman}; applying the same bound to \(-X_t\) gives
the two-sided claim.

\emph{Step 3: remove the mask below \(v\).}
If \(\sum_{t\le\tau}k_t\overline\sigma_t^2\le v\), induction shows that
\(m_t^{(v)}=k_t\) for all \(t\le\tau\).
\end{proof}

\subsection{Vector concentration under predictable contractions}
\label{sec:vector-tools}

Since contractions do not increase the norm, the supermartingale
argument behind \citet[Theorem~3.4]{pinelis1994optimum} also applies to
the recursion of Lemma~\ref{lem:contractive-dilation} directly. The
dilation instead makes every norm bound for Hilbert-space martingales
available unchanged, because it preserves increment norms and
conditional covariances up to a predictable unitary map.
Unlike self-normalised Hilbert-space Bennett inequalities
\citep{janz2026selfnormalised}, which normalise an ordinary martingale by
its own predictable covariance process at the cost of a Fredholm
log-determinant term, Proposition~\ref{prop:contractive-classical} allows
arbitrary predictable contractions (in Lemma~\ref{lem:lifted-uniform},
built from trigger-gap covers rather than the noise covariance) and pays
only a trace-variance term without a log-determinant penalty
(Figure~\ref{fig:dilation}).

\begin{proofof}{Lemma~\ref{lem:contractive-dilation}.}
We work in the Hilbert direct sum
\(\mathcal K=\bigoplus_{r\ge0}\mathcal H_r\), which is separable, where
each \(\mathcal H_r\) is a copy of \(\mathcal H\), and write
\(\iota_r\) for its canonical embedding.

\emph{Step 1: replace each contraction by a unitary map.}
For a contraction \(K\), set
\(D_K=(I-K^*K)^{1/2}\) and \(D_{K^*}=(I-KK^*)^{1/2}\).
The Halmos unitary dilation \citep{halmos1950normal} is
\begin{equation}
 \mathsf U_K=
 \begin{pmatrix}K&D_{K^*}\\D_K&-K^*\end{pmatrix}.
\end{equation}
To check unitarity, first note that
\(K(I-K^*K)^n=(I-KK^*)^nK\) for every integer \(n\ge0\).
Polynomial approximation to the square root on \([0,1]\) gives
\(KD_K=D_{K^*}K\), and therefore
\[
 \mathsf U_K\mathsf U_K^*
 =\begin{pmatrix}
 KK^*+D_{K^*}^2&KD_K-D_{K^*}K\\
 D_KK^*-K^*D_{K^*}&D_K^2+K^*K
 \end{pmatrix}=I.
\]
The same calculation with \(K^*\) gives
\(\mathsf U_K^*\mathsf U_K=I\).

Let \(\widetilde U_t\) act as \(\mathsf U_{K_t}\) on
\(\mathcal H_0\oplus\mathcal H_t\) and as the identity elsewhere. Put
\begin{equation}
 W_0=0,\qquad W_t=\widetilde U_tW_{t-1}+\iota_0\zeta_t.
\end{equation}
No earlier update uses coordinate \(\mathcal H_t\), so its component
in \(W_{t-1}\) is zero. Induction now gives
\begin{equation}
 \iota_0^*W_t=K_t\iota_0^*W_{t-1}+\zeta_t=\mathcal Z_t.
\end{equation}

\emph{Step 2: undo the unitary updates.}
Set \(\mathcal U_0=I\), \(\mathcal U_t=\widetilde U_t\cdots\widetilde U_1\), and
\(M_t=\mathcal U_t^*W_t\). Since
\(\mathcal U_t^*\widetilde U_t=\mathcal U_{t-1}^*\),
\[
 M_t=M_{t-1}+\Delta_t,\qquad
 \Delta_t=\mathcal U_t^*\iota_0\zeta_t.
\]
Every factor in \(\mathcal U_t\) is strongly \(\mathcal F_{t-1}\)-measurable:
separability of \(\mathcal H\) makes \(K_t^*\) strongly predictable, and
uniform polynomial approximation on \([0,1]\) makes \(D_{K_t}\) and
\(D_{K_t^*}\) strongly predictable.
Thus \(\E_t\Delta_t=\mathcal U_t^*\iota_0\E_t\zeta_t=0\), proving the
martingale property. Unitarity and the isometry \(\iota_0\) give
\(\twonorm{\Delta_t}=\twonorm{\zeta_t}\), and orthogonal projection and
unitarity give
\[
 \twonorm{\mathcal Z_t}
 =\twonorm{\iota_0^*W_t}
 \le\twonorm{W_t}
 =\twonorm{\mathcal U_t^*W_t}
 =\twonorm{M_t}. \qedhere
\]
\end{proofof}

\paragraph{Bennett bound.}
Proposition~\ref{prop:contractive-classical} uses only the covariance
trace and makes the variance levels and their probability allocation
explicit.

\begin{proofof}{Proposition~\ref{prop:contractive-classical}.}
Apply Lemma~\ref{lem:contractive-dilation}. The increments
\(\Delta_t=M_t-M_{t-1}\) of its martingale are bounded by \(R\) and satisfy
\[
 \sum_{t\le\tau}\E_t\twonorm{\Delta_t}^2=\sigma_\tau^2.
\]

\emph{Step 1: impose a deterministic variance bound.}
Fix \(v>0\), and define
\[
 I_t^{(v)}=\II{\sigma_t^2\le v},\qquad
 M_\tau^{(v)}=\sum_{t\le\tau}I_t^{(v)}\Delta_t.
\]
The indicator is predictable: \(\sigma_t^2\) includes the current
conditional second moment, which is known before the draw. Hence
\(M^{(v)}\) is a martingale with increments bounded by \(R\), and
\[
 \sum_{t\le T}\E_t\twonorm{I_t^{(v)}\Delta_t}^2
 =\sum_{t\le T}I_t^{(v)}(\sigma_t^2-\sigma_{t-1}^2)\le v.
\]
Indeed, \(\sigma_t^2\) is nondecreasing, so the retained increments
form the prefix before its first crossing of \(v\). If
\(\sigma_\tau^2\le v\), every increment up to \(\tau\) is retained
and \(M_\tau^{(v)}=M_\tau\).

\emph{Step 2: apply the Hilbert-space maximal inequality (Part~(a)).}
Hilbert spaces have smoothness constant \(D=1\).
The Bennett inequality of \citet[Theorem~3.4]{pinelis1994optimum}, with increment bound \(R\)
and variance bound \(v\), gives, for every \(T\ge1\),
\[
 \bbP\left\{\max_{\tau\le T}\twonorm{M_\tau^{(v)}}>r\right\}
 \le2\exp\left\{-\frac v{R^2}\mathfrak h(Rr/v)\right\}.
\]
By monotone convergence as \(T\to\infty\), the same bound holds for
\(\bbP\{\sup_{\tau\ge1}\twonorm{M_\tau^{(v)}}>r\}\).
For \(x>0\), setting \(r=(v/R)\mathfrak h^{-1}(R^2x/v)\) makes the
exponent equal to \(-x\), so the probability is at most \(2e^{-x}\); by
Lemma~\ref{lem:bennett-comparisons} with \(z=R^2x/v\),
\(r\le\sqrt{2vx}+Rx/3\). Taking \(x=\ell+\log2\) gives probability at
most \(e^{-\ell}\) that \(\sup_{\tau\ge1}\twonorm{M_\tau^{(v)}}\)
exceeds the right-hand side of \eqref{eq:contractive-fixed-v}. Whenever
\(\sigma_\tau^2\le v\), Lemma~\ref{lem:contractive-dilation} and Step~1
give \(\twonorm{\mathcal Z_\tau}\le\twonorm{M_\tau}=\twonorm{M_\tau^{(v)}}\),
proving~(a).

\emph{Step 3: stitch variance levels on one event (Part~(b)).}
Following the geometric-grid stitching method \citep[Section~3.1]{howard2021time},
use the infinite grid \(v_r=2^rR^2\) for integers \(r\ge0\), and apply
Step~2 at level \(v_r\) with \(x=\ell+\log\{\pi^2(r+1)^2/3\}\). Since
\(2e^{-x}=\frac{6}{\pi^2(r+1)^2}e^{-\ell}\), a union bound over all
\(r\ge0\) costs at most \(\sum_{r\ge0}\frac{6}{\pi^2(r+1)^2}e^{-\ell}=e^{-\ell}\).
On the resulting event, fix any \(\tau\ge1\) and let
\(r=\ceil{\log_2(1\vee\sigma_\tau^2/R^2)}\ge0\) be the least grid index
with \(\sigma_\tau^2\le v_r\). Then \(v_r\le2(\sigma_\tau^2+R^2)\) and
\(\ell'_\tau=\ell+\log\{\pi^2(r+1)^2/3\}\), so Lemma~\ref{lem:contractive-dilation}
and Steps~1--2 give
\[
 \twonorm{\mathcal Z_\tau}
 \le\twonorm{M_\tau}
 =\twonorm{M_\tau^{(v_r)}}
 \le\sqrt{2v_r\ell'_\tau}+\frac{R\ell'_\tau}3
 \le\sqrt{4(\sigma_\tau^2+R^2)\ell'_\tau}+\frac{R\ell'_\tau}3.
\]
This proves~(b). For \(\tau\le T\), \(\sigma_\tau^2\le TR^2\) gives
\(r\le\ceil{\log_2T}\), so
\(\ell'_\tau\le\ell+\log\{\pi^2(1+\ceil{\log_2T})^2/3\}\).
\end{proofof}

\section{Estimation error after contraction credit}
\label{sec:bandit-error}

We prove Lemma~\ref{lem:directional-geometry}, establish the epochwise
reverse-moment and lifted-noise concentration events
(\(\mathcal E_{\mathrm{rev}}\) and \(\mathcal E_{\mathrm{vec}}\)), and
bound the estimation error after contraction credit uniformly over unit
comparators (Lemma~\ref{lem:bandit-correction}).

\subsection{Directional gap and estimator bias}

Recall \(\Gamma_{t-1}(v)=\twonorm{\widehat\Theta_{t-1}}\{1-\iprod v{e_t}\}\).

\begin{proofof}{Lemma~\ref{lem:directional-geometry}.}
\emph{The geometric bound \eqref{eq:stock-geometry}.}
Write \(c=\iprod v{e_t}\) and \(\eta=1-c\).
The tangential component of \(z_t-v\) has squared norm
\(1-c^2=\eta(2-\eta)\) and its radial component is \(\eta-p_t\), so
\(\normt{G_t}{z_t-v}^2=\eta(2-\eta)+p_t(\eta-p_t)^2\). With
\(p_tr_t=1-p_t\), expanding gives
\[
 p_t(1+2r_t\eta)-\normt{G_t}{z_t-v}^2
 =(1-p_t)\{(\eta-p_t)^2+p_t\}\ge0.
\]
Dividing by \(p_t\) yields \eqref{eq:stock-geometry}, since
\(r_t\eta=\twonorm{\widehat\Theta_{t-1}}\{1-\iprod v{e_t}\}/\beta_t=\Gamma_{t-1}(v)/\beta_t\).

\emph{The bias bound \eqref{eq:bias-bound}.}
Multiplying \eqref{eq:qix-lambda-bound} by
\(\sqrt{p_t}\,\normt{G_t}{\theta_t}\), and using
\(\beta_t\sqrt{p_t}\ge\sqrt{\kappa_{j_t}}\) from \eqref{eq:cubic-support},
\(\kappa_{j_t}\ge(j_t+1)\ell_{j_t}^2\) from \eqref{eq:kappa-bounds},
\(3d/\{4(d+2)\}<1\), and
\(\normt{G_t}{\theta_t}\le1\),
\[
 \lambda_t\sqrt{p_t}\,\normt{G_t}{\theta_t}
 \le\frac{3d^2\ell_{j_t}^2\normt{G_t}{\theta_t}^3}{4\beta_t^2\sqrt{p_t}\,(d+2)}
 \le\frac{d\ell_{j_t}\normt{G_t}{\theta_t}^2}{\sqrt{j_t+1}\,\beta_t}.
\]
Cauchy--Schwarz in the \(G_t\) metric and \eqref{eq:stock-geometry} then give
\[
 \lambda_t\abs{\iprod{z_t-v}{G_t\theta_t}}
 \le\lambda_t\sqrt{p_t}\,\normt{G_t}{\theta_t}\frac{\normt{G_t}{z_t-v}}{\sqrt{p_t}}
 \le\frac{d\ell_{j_t}\normt{G_t}{\theta_t}^2}{\sqrt{j_t+1}\,\beta_t}
       \sqrt{1+2\Gamma_{t-1}(v)/\beta_t}. \qedhere
\]
\end{proofof}

\subsection{Covariance trigger for comparator-uniform concentration}

Adapting \citet[Algorithm~2 and Lemma~B.12]{lee2020bias}, with
\(r_t=\twonorm{\widehat\Theta_{t-1}}/\beta_t\), define
\begin{equation}
 A_t=\frac{r_t}{2}
 \begin{pmatrix}I_d&e_t\\ e_t\transpose&1\end{pmatrix},
 \label{eq:cover-matrix}
\end{equation}
starting the cover \(H_j\) at \(I_{d+1}\) in each epoch
\(I_j=\{t:B_j\le\beta_t<2B_j\}\), adding \(A_t\) and
contracting whenever \(A_t\npreceq H_j\).
Let \(\mathcal S_j\) be the trigger times of epoch \(j\), let
\(H_j=I_{d+1}+\sum_{s\in\mathcal S_j}A_s\) denote its terminal cover once
the epoch ends, and lift a unit vector \(v\) to \(\bar v=(-v,1)\).

\begin{lemma}[Scalar trigger and dimension-free count]
\label{lem:trigger}
Pathwise, the matrix trigger in Algorithm~\ref{alg:radial-qix} is equivalent to its
scalar test, and to
\(\bar v\transpose A_t\bar v>\bar v\transpose H_j\bar v\)
for some unit \(v\). Each epoch has fewer than \(10(j+1)\) triggers.
\end{lemma}

\begin{proof}
\emph{The scalar trigger test.}
After any set of triggers, write
\(\alpha=1+\tfrac12\sum_{s\in\mathcal S_j}r_s\) and
\(b=\tfrac12\sum_{s\in\mathcal S_j}r_se_s\), so the cover has block form
\(H_j=\bigl(\begin{smallmatrix}\alpha I_d&b\\b\transpose&\alpha\end{smallmatrix}\bigr)\).
Set \(D=\alpha^2-\twonorm b^2\).  The triangle inequality gives
\(\twonorm b\le\frac12\sum_{s\in\mathcal S_j}r_s<\alpha\), so \(D>0\) (initially
\(D=1\)).  For \(r\ge0\) and a unit vector \(e\), let \(A(r,e)\) be the
matrix \eqref{eq:cover-matrix} with \((r,e)\) in place of
\((r_t,e_t)\), so \(A_t=A(r_t,e_t)\).  Direct subtraction shows
\begin{equation}
 A(r,e)\preceq H_j
 \quad\Longleftrightarrow\quad
 r\{\alpha-\iprod be\}\le D.
 \label{eq:exact-trigger}
\end{equation}
Indeed,
\[
 H_j-A(r,e)=
 \begin{pmatrix}
 (\alpha-r/2)I_d&b-re/2\\
 (b-re/2)\transpose&\alpha-r/2
 \end{pmatrix},
\]
whose smallest eigenvalue is \(\alpha-r/2-\twonorm{b-re/2}\), so it is
positive semidefinite exactly when \(\alpha-r/2\ge\twonorm{b-re/2}\).
Its quadratic form at \(\bar v\) is
\(2(\alpha-r/2)-2\iprod{b-re/2}{v}\), with minimum
\(2\{\alpha-r/2-\twonorm{b-re/2}\}\) over unit \(v\), so the same
condition is
\(\bar v\transpose A(r,e)\bar v\le\bar v\transpose H_j\bar v\)
for every unit \(v\).
Squaring both sides and cancelling \(r^2/4\),
\[
 \alpha^2-\alpha r\ge\twonorm b^2-r\iprod be
 \quad\Longleftrightarrow\quad
 r\{\alpha-\iprod be\}\le\alpha^2-\twonorm b^2=D,
\]
which is \eqref{eq:exact-trigger}.  The squaring is reversible because
\(\iprod be\le\twonorm b<\alpha\) makes the right-hand condition force
\(r\le D/(\alpha-\iprod be)\le(\alpha^2-\twonorm b^2)/(\alpha-\twonorm b)=\alpha+\twonorm b\le2\alpha\),
hence \(\alpha-r/2\ge0\).

\emph{The number of triggers.}
On a trigger, the failure of \eqref{eq:exact-trigger} gives
\[
 D^+
 =\Bigl(\alpha+\frac r2\Bigr)^2-\twonorm{b+re/2}^2
 =D+r\{\alpha-\iprod be\}
 >2D.
\]
By the cubic floor \eqref{eq:cubic-support} and
\(\beta_t<2B_j=2^{j+1}\beta_0\) (with \(\beta_0^2=\kappa_0\)), every round in epoch \(j\)
satisfies \(r_t\le\beta_t^2/\kappa_0-1<4^{j+1}-1\).
Put \(k=j+1\ge1\) and \(m=|\mathcal S_j|\). If \(m\ge1\), then
\(D^+>2D\) with \(D_0=1\) and \(D\le\alpha^2\) give
\[
 2^m<D_m\le\alpha_m^2
 =\Bigl(1+\frac12\sum_{s\in\mathcal S_j}r_s\Bigr)^2
 <\Bigl(1+\frac{m(4^k-1)}2\Bigr)^2,
\]
so \(2^{m/2}<1+m(4^k-1)/2\). On the other hand, if \(m=10k+s\) with
\(s\ge0\), then \(1+m(4^k-1)/2<\tfrac m2 4^k\) (since \(m\ge10\)), while
\(8^k\ge8k\) and \(2^{s/2}\ge1+s/4\) (since \(\tfrac12\log2>\tfrac14\)) yield
\[
 2^{m/2-2k}=8^k 2^{s/2}\ge8k\Bigl(1+\frac s4\Bigr)=8k+2ks>5k+\frac s2=\frac m2,
\]
contradicting \(2^{m/2}<\tfrac m2 4^k\). Hence \(|\mathcal S_j|<10(j+1)\).
\end{proof}

\begin{lemma}[Lifted gap and terminal cover]
\label{lem:trigger-cover}
For every unit \(v\), the following hold:
\begin{lemenum}
\item \(\bar v\transpose A_t\bar v
 =\Gamma_{t-1}(v)/\beta_t\).
 \label{lem:trigger-cover:gap}
\item \(m_j(v)=\bar v\transpose H_j\bar v
 =2+\sum_{s\in\mathcal S_j}\Gamma_{s-1}(v)/\beta_s\ge2\).
 \label{lem:trigger-cover:mass}
\item \(A_t\preceq H_j\) for \(t\in I_j\).
 \label{lem:trigger-cover:cover}
\end{lemenum}
\end{lemma}

\begin{proof}
After each round's test,
\(A_t\preceq H_j\), giving (c). For unit
\(v\) and \(\bar v=(-v,1)\),
\[
 \bar v\transpose A_t\bar v
 =\frac{r_t}2\{\twonorm v^2-2\iprod v{e_t}+1\}
 =r_t\{1-\iprod v{e_t}\}=\frac{\Gamma_{t-1}(v)}{\beta_t},
\]
proving (a). Summing over \(\mathcal S_j\) and adding
\(\bar v\transpose I_{d+1}\bar v=2\) gives (b), since \(\iprod{v}{e_s}\le1\).
\end{proof}

\begin{lemma}[Contraction credit]
\label{lem:trigger-credit}
For every unit \(v\), contractions supply credit proportional to the
mass \(m_j(v)\) in Lemma~\ref{lem:trigger-cover}:
\begin{equation}
 -\sum_{s\in\mathcal S_j}\epsilon_s\Gamma_{s-1}(v)
 \le-\frac{B_j}4\{m_j(v)-2\}.
 \label{eq:strict-credit}
\end{equation}
\end{lemma}

\begin{proof}
Compare \citet[Lemma~B.13]{lee2020bias} and
\citet[Appendix~F, Step~4]{zimmert2022bias}. On a trigger,
\(\epsilon_s=1/4\) and \(\beta_s\ge B_j\), so
\[
 \sum_{s\in\mathcal S_j}\epsilon_s\Gamma_{s-1}(v)
 =\frac14\sum_{s\in\mathcal S_j}\beta_s\frac{\Gamma_{s-1}(v)}{\beta_s}
 \ge\frac{B_j}4\sum_{s\in\mathcal S_j}\frac{\Gamma_{s-1}(v)}{\beta_s}
 =\frac{B_j}4\{m_j(v)-2\},
\]
which is \eqref{eq:strict-credit} after negation.
\end{proof}

\subsection{Lifting the directional error}
\label{sec:lifted}

For each epoch \(j\ge0\), write
\(H_{j,\tau}\) for the cover after the predictable trigger
test at round \(\tau\), frozen at \(I_{d+1}\) before the epoch and
at its terminal value afterwards (and constant at \(I_{d+1}\) if
\(I_j=\varnothing\)), so \(H_j=H_{j,T}\).
The cover process is predictable and nondecreasing in the Loewner order.

For \(t\in I_j\), let \(\Pi_t\colon\R^{d+1}\to\R^d\) be the map
\(\Pi_t(x,s)=x+(1-p_t)se_t\). Its lifted noise is
\(\bar\xi_t=-\Pi_t\transpose\xi_t\), with metric
\(C_t=\Pi_t\transpose G_t\Pi_t/p_t\).
Then \(-\iprod{z_t-v}{\xi_t}=\iprod{\bar v}{\bar\xi_t}\) for
\(\bar v=(-v,1)\): this lift combines the directional estimator noise
with its mean-action component. The concentration step,
Lemma~\ref{lem:lifted-uniform} below, uses
the following domination bound.

\begin{lemma}[Cover domination]
\label{lem:cover-domination}
For every nonempty epoch \(j\) and \(t\in I_j\),
\(C_t\preceq2H_{j,t}\).
\end{lemma}

\begin{proof}
We prove \(2H_{j,t}-C_t\succeq0\) by a scalar
Schur-complement calculation.

\emph{Cover constraints and the scalar target.}
For \(y=(x,s)\), write \(x=x_\perp+ce_t\), so that
\(\Pi_ty=x_\perp+\{c+(1-p_t)s\}e_t\).  Since
\(G_t/p_t=p_t^{-1}P_{\perp,t}+P_{\parallel,t}\) and \(p_t^{-1}=1+r_t\),
\[
 y\transpose C_ty
 =\frac{\normt{G_t}{\Pi_ty}^2}{p_t}
 =(1+r_t)\twonorm{x_\perp}^2+\{c+(1-p_t)s\}^2.
\]
Suppress \(t,j\), write
\(H=H_{j,t}=\bigl(\begin{smallmatrix}\alpha I_d&b\\b\transpose&\alpha\end{smallmatrix}\bigr)\)
as in Lemma~\ref{lem:trigger}, and set \(r=r_t\), \(p=(1+r)^{-1}\), and
\(q=1-p\). Resolve \(b\) into its radial component
and squared transverse norm \(w^2=\twonorm{b-\iprod{b}{e_t}e_t}^2\),
and set \(\gamma=2\alpha-1\) and \(u=\alpha-\iprod{b}{e_t}\).
Since \(\twonorm{b}\le\alpha-1\),
\begin{equation}
 1\le u\le\gamma,\qquad
 w^2\le(u-1)(\gamma-u).
 \label{eq:cover-basic-constraints}
\end{equation}
Moreover, the post-test cover relation
\(A_t\preceq H\), equivalently
\eqref{eq:exact-trigger}, gives
\begin{equation}
 w^2\le u(\gamma+1-u-r),\qquad r\le\gamma.
 \label{eq:cover-trigger-constraints}
\end{equation}
On the span of \(e_t\), \(b-\iprod{b}{e_t}e_t\), and the lift coordinate, the matrix
\(2H-C_t\) is
\[
 \begin{pmatrix}
 \gamma&0&\gamma+1-2u-q\\
 0&\gamma-r&2w\\
 \gamma+1-2u-q&2w&\gamma+1-q^2
 \end{pmatrix};
\]
on the remaining directions it equals \(\gamma-r\) times the identity.  For
\(0<r<\gamma\), the first two diagonal entries are positive, so the
Schur complement reduces positive semidefiniteness to
\[
 F(u,w^2):=
 \gamma(\gamma-r)(\gamma+1-q^2)
 -(\gamma-r)(\gamma+1-2u-q)^2-4\gamma w^2
 \ge0.
\]
The bound in \eqref{eq:cover-basic-constraints} minus that in
\eqref{eq:cover-trigger-constraints} is \(ru-\gamma\), which
determines the case split.

\emph{Case \(ru\le\gamma\): accumulated-cover constraint.}
Substituting \eqref{eq:cover-basic-constraints} yields a convex lower
bound \(F_0\) with \(F_0''(u)=8r\) and minimiser
\(u_0=\{\gamma(r+2)+1\}/\{2(r+1)\}\), where
\[
 F_0(u_0)
 =\gamma(\gamma-1)+\frac{\gamma\{\gamma(2-r^2)-r\}}{r+1}.
\]
The feasible upper endpoint is \(\gamma\) for \(r\le1\) and
\(\gamma/r\) for \(r\ge1\); the lower endpoint never binds, because
\(\gamma\ge1\) and \(r\le\gamma\) give
\(2(r+1)(u_0-1)=(\gamma-1)(r+1)+(\gamma-r)\ge0\).
If \(u_0\) is feasible, then \(u_0\le\gamma\) (for \(r\le1\)) gives
\(\gamma r\ge1\) and \(r^2+r/\gamma\le2r^2\le2\), while \(u_0\le\gamma/r\)
(for \(r\ge1\)) gives \(r^2+r/\gamma\le2\) directly; in either case
\(F_0(u_0)\ge\gamma(\gamma-1)\ge0\).
Otherwise the minimum lies at the upper endpoint, where
\[
 F_0(\gamma)
 =\frac{(\gamma-r)(4\gamma r+3\gamma-1)}{(1+r)^2}\ge0,
 \qquad
 F_0(\gamma/r)
 =\frac{(\gamma-r)(3\gamma r+4\gamma-r)}{r(1+r)^2}\ge0.
\]

\emph{Case \(ru\ge\gamma\): current-round constraint.}
Substituting \eqref{eq:cover-trigger-constraints} yields a lower bound
\(F_1\) with
\(F_1'(u)=\frac{4r}{1+r}\{2(r+1)u-(\gamma+1)\}>0\),
since \(2(r+1)u-(\gamma+1)\ge\gamma+2\gamma/r-1>0\) for
\(u\ge\gamma/r\) and \(\gamma\ge1\). Since the two cover bounds
coincide at \(ru=\gamma\),
\(F_1(u)\ge F_1(\gamma/r)=F_0(\gamma/r)\ge0\).

For the boundary cases, \(r=0\) gives
\(C_t=\operatorname{diag}(I_d,0)\preceq2H\), since \(H\succeq I_{d+1}\).
If \(r=\gamma\), \eqref{eq:cover-trigger-constraints} and
\(u\ge1\) force \(u=1\) and \(w=0\), and the remaining
\(2\)-by-\(2\) block is positive semidefinite as the limit \(r\uparrow\gamma\).
\end{proof}

\subsection{Reverse-moment and lifted-noise concentration}
\label{sec:vector-event}

For each epoch \(j\ge0\), put
\(\Qcurv^{(j)}(\tau)=\sum_{t\in I_j,\,t\le\tau}\normt{G_t}{\theta_t}^2\),
recall \(j_-=[j-2]_+\), and define
\(K_j=a_j^{-2}+\ell_{j_-}^{-2}-1/(4B_j^2)>0\) (since
\(B_j\ge\beta_0>a_0\ge a_j\) by \eqref{eq:base-scales}). Let \(x_j>0\) be
the unique positive solution of
\(e^{x_j}=1+x_j+\ell_j/(\ell_{j_-}^2K_j)\), and set
\begin{equation}
 L_j=32K_je^{x_j}.
 \label{eq:exact-shell-data}
\end{equation}

\begin{lemma}[Reverse-moment event \(\mathcal E_{\mathrm{rev}}\)]
\label{lem:shell-reverse-clock}
For every \(j\ge0\), \(L_j\le C_0(j+1)/\ell_j\). Moreover, with
probability at least \(1-\delta_0\), simultaneously for all \(j\ge0\) and
all \(\tau\ge1\), \(d\Qcurv^{(j)}(\tau)\le B_j^2L_j\).
\end{lemma}

\begin{proof}
For the size of \(L_j\), \(a_j^{-2}=48(j+1)/\ell_j\) and
\(\ell_j\le2\ell_{j_-}\) (by \eqref{eq:kappa-bounds}) give
\(K_j\le a_j^{-2}+\ell_{j_-}^{-2}\le C_0(j+1)/\ell_j\). Since
\(e^x\ge2x\) for \(x\ge0\), \(e^x=1+x+z\) implies \(e^x\le2(1+z)\);
taking \(z=\ell_j/(\ell_{j_-}^2K_j)\) yields
\(L_j\le64(K_j+\ell_j/\ell_{j_-}^2)\le C_0(j+1)/\ell_j\).

For the concentration bound, fix \(j\ge0\) and put
\(R=4B_j^2/\ell_{j_-}^2\) and
\(X_t=\II{t\in I_j}\,p_t\normt{G_t}{\widehat\theta_t}^2\). By
Lemma~\ref{lem:rate-jumps} and Lemma~\ref{lem:qix:local},
\(0\le X_t<R\) on every round and the indicator \(\II{t\in I_j}\) is
predictable. For \(x>0\), convexity gives
\(e^{-xX_t/R}\le1-(1-e^{-x})X_t/R\), yielding a nonnegative exponential
supermartingale. Ville's inequality at level
\(\rho_j=6\delta_0/\{\pi^2(j+1)^2\}\) (so \(\log(1/\rho_j)=\ell_j\)) and a
union bound over \(j\ge0\) give, with probability at least \(1-\delta_0\),
simultaneously for all \(j\ge0\) and \(\tau\ge1\),
\[
 \sum_{t\le\tau}\E_tX_t
 \le\frac{x_j}{1-e^{-x_j}}\sum_{t\le\tau}X_t
 +\frac{R\ell_j}{1-e^{-x_j}}.
\]
Fix \(\tau\ge1\). If \(I_j\cap\{1,\ldots,\tau\}=\varnothing\), then
\(\Qcurv^{(j)}(\tau)=0\). Otherwise let
\(s=\max\{t\in I_j:t\le\tau\}\); then \(a_{j_s}\ge a_j\),
\(\ell_{j_s}\ge\ell_{j_-}\), and \(\beta_s<2B_j\), so the scale rule and
Lemma~\ref{lem:qix:local} give
\(\sum_{t\le\tau}X_t\le\EestObs_s\le\beta_s^2/a_{j_s}^2-1+\beta_s^2/\ell_{j_s}^2<4B_j^2K_j\).
Combining this with
\(\sum_{t\le\tau}\E_tX_t\ge d\Qcurv^{(j)}(\tau)/8\)
(Lemma~\ref{lem:qix:sandwich}) and the defining equation of \(x_j\) yields
\[
 d\Qcurv^{(j)}(\tau)
 <32B_j^2\frac{x_jK_j+\ell_j/\ell_{j_-}^2}{1-e^{-x_j}}
 =32B_j^2K_j\frac{e^{x_j}-1}{1-e^{-x_j}}
 =32B_j^2K_je^{x_j}=B_j^2L_j. \qedhere
\]
\end{proof}

Next, Lemma~\ref{lem:cover-domination} supplies the range and conditional
variance of the lifted noise \(\bar\xi_t\) in the evolving cover norm, and
Proposition~\ref{prop:contractive-classical:fixed}
(Appendix~\ref{sec:vector-tools}) gives a horizon-free bound up to the
predictable variance cap \(10B_j^2L_j\) implied by
\(\mathcal E_{\mathrm{rev}}\). With epoch range bound
\(R_j=3\sqrt2\,B_j/\ell_{j_-}\), set
\begin{equation}
 \varrho_j
 =\sqrt{20B_j^2L_j(\ell_j+\log2)}
   +\frac{R_j}{3}(\ell_j+\log2).
 \label{eq:evolving-bennett-radius}
\end{equation}

\begin{lemma}[Vector event \(\mathcal E_{\mathrm{vec}}\)]
\label{lem:lifted-uniform}
With probability at least \(1-\delta_0\), simultaneously for every
epoch \(j\ge0\) and every prefix \(\tau\ge1\) with
\(d\Qcurv^{(j)}(\tau)\le B_j^2L_j\),
\begin{equation}
 \norm[\bigg]{
 \sum_{\substack{t\in I_j\\t\le\tau}}\bar\xi_t}_{H_{j,\tau}^{-1}}
 \le \varrho_j.
 \label{eq:classical-vector-event}
\end{equation}
In particular, on \(\mathcal E_{\mathrm{rev}}\cap\mathcal E_{\mathrm{vec}}\),
for every unit \(v\), every \(j\ge0\), and every \(\tau\ge1\),
\begin{equation}
 -\sum_{\substack{t\in I_j\\t\le\tau}}\iprod{z_t-v}{\xi_t}
 \le \varrho_j\normt{H_{j,\tau}}{\bar v}.
 \label{eq:local-vector-uniform}
\end{equation}
\end{lemma}

\begin{proof}
\emph{Step 1: bound the noise after the linear transformation.}
Fix \(t\in I_j\) and write \(H=H_{j,t}\), which is predictable.
Lemma~\ref{lem:cover-domination} gives
\(H^{-1/2}C_tH^{-1/2}\preceq2I_{d+1}\), and \(\operatorname{rank}C_t\le d\)
because \(\Pi_t\) maps into \(\R^d\). Thus
\(\opnorm{H^{-1/2}\Pi_t\transpose G_t^{1/2}/\sqrt{p_t}}\le\sqrt2\). Since
\(\ell_{j_-}\le\ell_{j_t}\le\ell_j\le B_j\) (by Lemma~\ref{lem:rate-jumps}
and \eqref{eq:base-scales}), Lemma~\ref{lem:qix:range} and
\(\beta_t<2B_j\) give
\begin{equation}
 \normt{H^{-1}}{\bar\xi_t}
 =\twonorm{H^{-1/2}\Pi_t\transpose\xi_t}
 \le\opnorm{H^{-1/2}\Pi_t\transpose G_t^{1/2}/\sqrt{p_t}}
    \twonorm{\sqrt{p_t}\,G_t^{-1/2}\xi_t}
 \le\sqrt2\,(\beta_t/\ell_{j_t}+1)
 <R_j.
 \label{eq:vector-range}
\end{equation}
Conjugating Lemma~\ref{lem:qix:cov} by \(H^{-1/2}\Pi_t\transpose\) gives
\[
 \operatorname{Cov}_t\bigl(H^{-1/2}\bar\xi_t\bigr)
 \preceq5\normt{G_t}{\theta_t}^2\,H^{-1/2}C_tH^{-1/2};
\]
the right-hand side has rank at most \(d\) and eigenvalues at most
\(10\normt{G_t}{\theta_t}^2\), so
\begin{equation}
 \E_t\normt{H^{-1}}{\bar\xi_t}^2
 \le10d\normt{G_t}{\theta_t}^2.
 \label{eq:vector-trace}
\end{equation}

\emph{Step 2: write the moving-cover recursion.}
Fix \(j\ge0\). Both \(\II{t\in I_j}\) and \(H_{j,t}\)
(Appendix~\ref{sec:lifted}) are predictable, and \(H_{j,t}\) is
nondecreasing in the Loewner order with \(H_{j,0}=I_{d+1}\). Whitening the
cumulative lifted noise \(S_t^{(j)}=\sum_{k\le t,\,k\in I_j}\bar\xi_k\) in
the current cover norm gives \(\mathcal Z_t^{(j)}=H_{j,t}^{-1/2}S_t^{(j)}\).
With \(K_t^{(j)}=H_{j,t}^{-1/2}H_{j,t-1}^{1/2}\) (a predictable
contraction, since
\(K_t^{(j)}(K_t^{(j)})^*=H_{j,t}^{-1/2}H_{j,t-1}H_{j,t}^{-1/2}\preceq I\)),
\begin{equation}
 \mathcal Z_t^{(j)}
 =K_t^{(j)}\mathcal Z_{t-1}^{(j)}+\zeta_t^{(j)},
 \qquad
 \zeta_t^{(j)}
 =H_{j,t}^{-1/2}\II{t\in I_j}\bar\xi_t.
 \label{eq:evolving-cover-recursion}
\end{equation}
Then \(\E_t\zeta_t^{(j)}=0\), and \eqref{eq:vector-range}--\eqref{eq:vector-trace}
give \(\twonorm{\zeta_t^{(j)}}\le R_j\) and
\(\sum_{t\le\tau}\E_t\twonorm{\zeta_t^{(j)}}^2\le10d\Qcurv^{(j)}(\tau)\)
for every \(\tau\ge1\).

\emph{Step 3: apply the Bennett bound and conclude.}
Apply Proposition~\ref{prop:contractive-classical:fixed} to
\eqref{eq:evolving-cover-recursion} with \(R=R_j\), \(v=10B_j^2L_j\), and
\(\ell=\ell_j\). Whenever \(d\Qcurv^{(j)}(\tau)\le B_j^2L_j\),
\eqref{eq:classical-vector-event} holds simultaneously over all such
\(\tau\ge1\) with failure probability at most
\(e^{-\ell_j}=6\delta_0/\{\pi^2(j+1)^2\}\); a union bound over \(j\ge0\)
gives total failure probability at most \(\delta_0\). On
\(\mathcal E_{\mathrm{rev}}\cap\mathcal E_{\mathrm{vec}}\),
Cauchy--Schwarz in the prefix cover gives \eqref{eq:local-vector-uniform}
simultaneously for all \(v\in\sphere^{d-1}\).
\end{proof}

\subsection{Error after contraction credit}

Noise and bias grow as the square root of the comparator's cover mass,
whereas credit grows linearly. Combining Lemmas~\ref{lem:shell-reverse-clock}
and~\ref{lem:lifted-uniform} yields the epochwise estimate and then the
summed estimation-error bound.

\begin{proposition}[Epochwise directional bound]
\label{prop:epoch-error}
On \(\mathcal E_{\mathrm{vec}}\cap\mathcal E_{\mathrm{rev}}\), for every
unit \(v\), each nonempty epoch satisfies
\begin{equation}
 \sum_{t\in I_j}\{\iprod{z_t-v}{\theta_t-\widehat{\theta}_t}-\epsilon_t\Gamma_{t-1}(v)\}
 \le\frac{B_j}2+\frac{(\varrho_j+3B_j\ell_jL_j/\sqrt{j+1})^2}{B_j}.
 \label{eq:epoch-error}
\end{equation}
\end{proposition}

\begin{proof}
Fix a nonempty epoch \(j\) and a unit \(v\), and write \(m=m_j(v)\), so
\(m\ge2\) by Lemma~\ref{lem:trigger-cover:mass}.
Lemma~\ref{lem:lifted-uniform} gives
\(-\sum_{t\in I_j}\iprod{z_t-v}{\xi_t}\le\varrho_j\sqrt m\).
For \(t\in I_j\), \(j_-\le j_t\le j\) gives \(\ell_{j_t}\le\ell_j\) and
\(j_t+1\ge j_-+1\ge(j+1)/3\). By Lemma~\ref{lem:trigger-cover},
\(\Gamma_{t-1}(v)/\beta_t\le m\); hence \eqref{eq:bias-bound},
\(\beta_t\ge B_j\), the reverse-moment event, and \(\sqrt{15/2}<3\) yield
\begin{align*}
 \sum_{t\in I_j}\lambda_t\iprod{z_t-v}{G_t\theta_t}
 &\le\frac{\sqrt3\,d\ell_j\Qcurv^{(j)}(T)}{\sqrt{j+1}\,B_j}\sqrt{1+2m}
 &&\text{(by \eqref{eq:bias-bound})}\\
 &\le\frac{\sqrt3\,B_j\ell_jL_j}{\sqrt{j+1}}\sqrt{1+2m}
 &&\text{(Lemma~\ref{lem:shell-reverse-clock})}\\
 &\le\frac{3B_j\ell_jL_j}{\sqrt{j+1}}\sqrt m.
 &&\text{(\(m\ge2,\ \sqrt{15/2}<3\))}
\end{align*}
Lemma~\ref{lem:trigger-credit} pays
\(B_j(m-2)/4\) for contractions. Thus the error after
credit is bounded by a concave quadratic in \(\sqrt m\); completing its square
bounds it independently of the comparator's cover mass:
\begin{align*}
 \sum_{t\in I_j}\{\iprod{z_t-v}{\theta_t-\widehat{\theta}_t}-\epsilon_t\Gamma_{t-1}(v)\}
 &\le\Bigl(\varrho_j+\frac{3B_j\ell_jL_j}{\sqrt{j+1}}\Bigr)\sqrt m
      -\frac{B_j}4(m-2)\\
 &\le\frac{B_j}2
  +\frac{(\varrho_j+3B_j\ell_jL_j/\sqrt{j+1})^2}{B_j}. \qedhere
\end{align*}
\end{proof}

\begin{lemma}[Estimation error after contraction credit]
\label{lem:bandit-correction}
On \(\mathcal E_{\mathrm{rev}}\cap\mathcal E_{\mathrm{vec}}\), for every
\(T\ge1\) and every unit comparator \(v\), with \(j=j(\beta_T)\),
\[
 \sum_{t=1}^T\left\{\iprod{z_t-v}{\theta_t-\widehat\theta_t}
       -\epsilon_t\Gamma_{t-1}(v)\right\}
 +\sum_{t=1}^T\II{\epsilon_t>0}\beta_t(1-p_t)
 \le C_0(j+1)\beta_T.
\]
\end{lemma}

\begin{proof}
Apply Proposition~\ref{prop:epoch-error} with the same comparator \(v\).
By Lemma~\ref{lem:shell-reverse-clock} and \eqref{eq:kappa-bounds}, every
epoch \(j\ge0\) satisfies \(L_j\le C_0(j+1)/\ell_j\),
\(R_j=3\sqrt2\,B_j/\ell_{j_-}\le C_0B_j/\ell_j\), and
\(\ell_j+\log2\le2\ell_j\) (as \(\ell_j\ge\ell_0>2.577\)). Therefore
\eqref{eq:evolving-bennett-radius} gives
\(\varrho_j\le C_0B_j\sqrt{j+1}\) and
\(3B_j\ell_jL_j/\sqrt{j+1}\le C_0B_j\sqrt{j+1}\), so each nonempty
epoch's estimation error after credit is at most \(C_0(j+1)B_j\). Every
nonempty epoch has \(B_j\le\beta_T\), so \(j\le j(\beta_T)\) and
\(\sum_{j:I_j\ne\varnothing}B_j<2\beta_T\), yielding
\(C_0\{j(\beta_T)+1\}\beta_T\).

For the positive contraction costs, each trigger in epoch \(j\) has
\(\beta_s(1-p_s)<2B_j\); by Lemma~\ref{lem:trigger},
\[
 \sum_{j:I_j\ne\varnothing}\sum_{s\in\mathcal S_j}\beta_s(1-p_s)
 \le2\sum_{j:I_j\ne\varnothing}|\mathcal S_j|B_j
 \le20\sum_{j:I_j\ne\varnothing}(j+1)B_j
 \le40\{j(\beta_T)+1\}\beta_T.
\]
Adding the two bounds completes the proof.
\end{proof}

\section{Observed squares, radial growth, and scale control}
\label{sec:radial-growth}
\label{sec:observed-squares}

We establish the observed-square event \(\mathcal E_{\mathrm{up}}\) and
summed stability bound (Lemma~\ref{lem:summed-stability}), then control
cubic-floor scale increases via an effective radial potential and the
weighted radial event \(\mathcal E_{\mathrm{rad}}\) to prove the final
scale bound (Lemma~\ref{lem:final-scale}).

\subsection{Observed-square concentration and adaptive stability}
\label{sec:upper-event}

\begin{lemma}[Observed-square event \(\mathcal E_{\mathrm{up}}\)]
\label{lem:upper-clock}
With probability at least \(1-\nobreak\delta_0\), simultaneously for every
prefix \(\tau\ge1\),
\begin{equation}
 \EestObs_\tau\le\frac32d\Qcurv_\tau+
 \frac{24}{\ell_{j(\beta_\tau)}}\beta_\tau^2.
 \label{eq:implicit-clock}
\end{equation}
\end{lemma}

\begin{proof}
For \(j\ge0\), put \(M_j=4B_j^2/\ell_{j_-}^2\). By
Lemma~\ref{lem:rate-jumps} and Lemma~\ref{lem:qix:local},
\(p_t\normt{G_t}{\widehat\theta_t}^2\le\beta_t^2/\ell_{j_t}^2\le M_j\) on
\(\{t\in I_j\}\). Predictability of \(\II{t\in I_j}\) and convexity of the
exponential on \([0,M_j]\) make
\[
 \exp\left\{
 \lambda\sum_{t\le\tau}\II{t\in I_j}p_t\normt{G_t}{\widehat\theta_t}^2
 -\frac{e^{\lambda M_j}-1}{M_j}
  \sum_{t\le\tau}\II{t\in I_j}\E_t[p_t\normt{G_t}{\widehat\theta_t}^2]
 \right\}
\]
a nonnegative supermartingale for \(\lambda\ge0\). Take
\(\lambda=3/(4M_j)\) (so \((e^{\lambda M_j}-1)/(\lambda M_j)<3/2\)) and
apply Ville's inequality at level \(6\delta_0/\{\pi^2(j+1)^2\}\) for each
\(j\ge0\); a union bound fails with probability at most \(\delta_0\). On
that event, for every \(\tau\ge1\) and \(j\ge0\),
\[
 \sum_{t\le\tau}\II{t\in I_j}p_t\normt{G_t}{\widehat\theta_t}^2
 \le\frac32\sum_{t\le\tau}\II{t\in I_j}\E_t[p_t\normt{G_t}{\widehat\theta_t}^2]
 +\frac43M_j\ell_j.
\]
Put \(k=j(\beta_\tau)\). Summing over \(j\le k\) and using
Lemma~\ref{lem:qix:sandwich} gives
\(\EestObs_\tau\le\tfrac32d\Qcurv_\tau+\tfrac43\sum_{j\le k}M_j\ell_j\).
Let \(g_k:=(\ell_k/B_k^2)\sum_{j\le k}M_j\ell_j\). Then \(g_0=4\) and
\(g_k=(\ell_k/(4\ell_{k-1}))g_{k-1}+4(\ell_k/\ell_{[k-2]_+})^2\) for
\(k\ge1\). Since \(\ell_0>2.577\), direct checks give
\(\ell_1/\ell_0<1.54\), \(\ell_2/\ell_0<1.86\), \(\ell_2/\ell_1<1.21\),
so \(g_1<11.1\) and \(g_2<17.2<18\); for \(k\ge3\),
\(\ell_k/\ell_{k-1}\le\ell_3/\ell_2<4/3\) and
\(\ell_k/\ell_{k-2}\le\ell_3/\ell_1<\sqrt2\), so induction gives
\(g_k\le g_{k-1}/3+8<18\) for all \(k\ge0\). Since \(B_k^2\le\beta_\tau^2\),
\(\tfrac43\sum_{j\le k}M_j\ell_j\le24\beta_\tau^2/\ell_k\), proving
\eqref{eq:implicit-clock}.
\end{proof}

\begin{lemma}[Lagged adaptive stability sum]
\label{lem:lagged-clock}
Pathwise, for every \(\tau\ge1\),
\[
 \sum_{t=1}^\tau\frac{p_t\normt{G_t}{\widehat\theta_t}^2}{\beta_t}
 \le\frac{3}{a_{j_\tau}}\bigl(\sqrt{\EestObs_\tau+1}-1\bigr).
\]
\end{lemma}

\begin{proof}
Put \(w_t=\sqrt{\EestObs_t+1}\) (\(w_0=1\)), so
\(p_t\normt{G_t}{\widehat\theta_t}^2/\beta_t=(w_t-w_{t-1})(w_t+w_{t-1})/\beta_t\).
For \(t\le\tau\), \(a_{j_t}\ge a_{j_\tau}\), so the scale rule and
Lemma~\ref{lem:qix:local} give \(w_{t-1}\le\beta_t/a_{j_\tau}\) and
\(w_t^2\le\beta_t^2/a_{j_\tau}^2+\beta_t^2/\ell_{j_t}^2\). Since
\(a_{j_\tau}/\ell_{j_t}\le a_0/\ell_0<1\) by \eqref{eq:base-scales},
\((w_t+w_{t-1})/\beta_t<(1+\sqrt2)/a_{j_\tau}<3/a_{j_\tau}\); telescoping
in \(w_t-w_{t-1}\ge0\) gives the bound.
\end{proof}

Call a round \(1\le t\le T\) \emph{cubic-floor-active} if the cubic floor
strictly exceeds the other scale candidates,
\begin{equation}
 \rho_{\kappa_{j_t}}(\twonorm{\widehat\Theta_{t-1}})>
 \max\bigl\{\beta_{t-1},a_{j_t}\sqrt{\EestObs_{t-1}+1}\bigr\},
 \label{eq:support-binding}
\end{equation}
and let \(\rho_{\mathrm{bind},T}\) be the maximum of \(\beta_0\) and
\(\rho_{\kappa_{j_t}}(\twonorm{\widehat\Theta_{t-1}})\) over all
cubic-floor-active rounds \(t\le T\).

\begin{lemma}[Scale envelope]
\label{lem:scale-envelope}
On \(\mathcal E_{\mathrm{up}}\) (Lemma~\ref{lem:upper-clock}),
\begin{equation}
 \beta_T\le\max\Bigl\{\rho_{\mathrm{bind},T},\;
 \sqrt6\,a_{j(\beta_T)}\sqrt{\tfrac32d\Qcurv_{T-1}+1}\Bigr\}.
 \label{eq:max-clock-envelope}
\end{equation}
\end{lemma}

\begin{proof}
Expanding the monotone scale rule gives
\(\beta_T=\max\{\rho_{\mathrm{bind},T},\,a_{j_s}\sqrt{\EestObs_{s-1}+1}\}\),
where \(s\) is the last round where the second-order candidate strictly
increased the scale (if any). If \(\rho_{\mathrm{bind},T}\) attains the
maximum, \eqref{eq:max-clock-envelope} is immediate. Otherwise
\(\beta_T=\beta_s=a_{j_s}\sqrt{\EestObs_{s-1}+1}\) with \(s\ge2\) (as
\(a_0<\beta_0\)). Applying \eqref{eq:implicit-clock} at prefix \(s-1\)
(where \(j(\beta_{s-1})=j_s\) and \(\beta_{s-1}\le\beta_T\)) and
multiplying by \(a_{j_s}^2\), using \(24a_{j_s}^2/\ell_{j_s}\le1/2\)
(by \eqref{eq:clock-absorption}), gives
\(\tfrac12\beta_T^2\le a_{j_s}^2(\tfrac32d\Qcurv_{T-1}+1)\). By
Lemma~\ref{lem:rate-jumps}, \(j_s\ge[j(\beta_T)-2]_+\), so
\(j_s+1\ge\{j(\beta_T)+1\}/3\), \(\ell_{j_s}\le\ell_{j(\beta_T)}\), and
hence \(a_{j_s}^2\le3a_{j(\beta_T)}^2\), yielding
\eqref{eq:max-clock-envelope}.
\end{proof}

\begin{lemma}[Summed stability cost]
\label{lem:summed-stability}
On \(\mathcal E_{\mathrm{up}}\), for every \(T\ge1\), with \(j=j(\beta_T)\),
\[
 \sum_{t=1}^T\frac{p_t\normt{G_t}{\widehat\theta_t}^2}{\beta_t}
 \le C_0\bigl\{a_j^{-1}\sqrt{d(1+\Qcurv_T)}
 +(j+1)\beta_T\bigr\}.
\]
\end{lemma}

\begin{proof}
Put \(j=j(\beta_T)\). Lemma~\ref{lem:lagged-clock} (with \(j_T\le j\))
and \eqref{eq:implicit-clock} give
\[
 \sum_{t=1}^T\frac{p_t\normt{G_t}{\widehat\theta_t}^2}{\beta_t}
 \le\frac{3}{a_j}\sqrt{\tfrac32d\Qcurv_T+\frac{24}{\ell_j}\beta_T^2+1}.
\]
Since \(a_j^{-1}\sqrt{24/\ell_j}=\sqrt{1152(j+1)}/\ell_j\le C_0(j+1)\)
(\(\ell_j>1\)) and \(d\ge2\), splitting the square root gives the claim.
\end{proof}

\subsection{Effective radial potential, range, and weighted radial noise}

The radial estimate is \(\iprod{e_t}{\widehat\theta_t}\); its centred noise is
\(\iprod{e_t}{\xi_t}\), where \(\xi_t=\widehat\theta_t-\E_t\widehat\theta_t\).
Empty sums are zero.

\paragraph{Effective radial potential.}
Near the boundary, a radial loss \(\theta_t=-e_t\) increases the raw state
norm \(h=\twonorm{\widehat\Theta_{t-1}}\) by order \(1\), whereas the
local loss norm
\(\normt{G_t}{\theta_t}=\sqrt{\twonorm{P_{\perp,t}\theta_t}^2+p_t\iprod{e_t}{\theta_t}^2}\)
weights radial components only by \(\sqrt{p_t}\). To compare radial growth
directly with \(\normt{G_t}{\theta_t}\), we rescale radial increments by
the state-only lower bound on \(\sqrt{p_t}\) implied by the cubic floor
\(\rho_{\kappa_0}(h)=(\kappa_0^{3/2}+2\kappa_0h)^{1/3}\): since
\(\beta_t\ge\rho_{\kappa_0}(h)\), \(b\mapsto b/(h+b)\) is nondecreasing,
and \(\rho_{\kappa_0}(h)^3\ge\kappa_0(\rho_{\kappa_0}(h)+h)\)
(Lemma~\ref{lem:cubic-support}),
\[
 p_t=\frac{\beta_t}{h+\beta_t}
 \ge\frac{\rho_{\kappa_0}(h)}{h+\rho_{\kappa_0}(h)}
 \ge\frac{\kappa_0}{\rho_{\kappa_0}(h)^2},
 \qquad\text{hence}\qquad
 \sqrt{p_t}\ge\frac{\sqrt{\kappa_0}}{\rho_{\kappa_0}(h)}.
\]
Integrating this exploration weight from \(0\) defines the
\emph{effective radial potential}
\begin{equation}
 \Phi_{\kappa_0}(h)
 :=\int_0^h\frac{\sqrt{\kappa_0}}{(\kappa_0^{3/2}+2\kappa_0u)^{1/3}}\,du
 =\frac{3}{4\sqrt{\kappa_0}}\bigl\{\rho_{\kappa_0}(h)^2-\kappa_0\bigr\}.
 \label{eq:overview-potential}
\end{equation}
Put \(\varpi_t=\Phi_{\kappa_0}'(\twonorm{\widehat\Theta_{t-1}})
 =\sqrt{\kappa_0}/\rho_{\kappa_0}(\twonorm{\widehat\Theta_{t-1}})\in(0,1]\)
for the predictable radial weights. Since
\(\beta_t\ge\rho_{\kappa_0}(\twonorm{\widehat\Theta_{t-1}})\), we have
\(\varpi_t\beta_t\ge\beta_0\).

\begin{lemma}[Geometry of the effective radial potential]
\label{lem:growth-Phi-geometry}
The effective radial potential \(\Phi_{\kappa_0}\) is increasing and
concave with \(\Phi_{\kappa_0}(0)=0\) and derivative
\(\Phi_{\kappa_0}'(h)=\sqrt{\kappa_0}/\rho_{\kappa_0}(h)\). On every
round,
\begin{equation}
 \varpi_t^2\le p_t,
 \qquad
 \bigl[\varpi_t\iprod{e_t}{\widehat\Theta_{t-1}-\E_t\widehat\Theta_t}\bigr]_+
 \le\varpi_t(1-\lambda_tp_t)[-\iprod{e_t}{\theta_t}]_+\le\normt{G_t}{\theta_t}.
 \label{eq:growth-Phi-depth}
\end{equation}
Hence every interval \(I\) satisfies
\begin{equation}
 \sum_{t\in I}\varpi_t\iprod{e_t}{\widehat\Theta_{t-1}-\E_t\widehat\Theta_t}
 \le\sum_{t\in I}\normt{G_t}{\theta_t}
 \le\sqrt{|I|\sum_{t\in I}\normt{G_t}{\theta_t}^2}.
 \label{eq:growth-Phi-signed-mean}
\end{equation}
\end{lemma}

\begin{proof}
\emph{Derivative and concavity.}
From \eqref{eq:overview-potential}, \(\Phi_{\kappa_0}(0)=0\) and
\(\Phi_{\kappa_0}'(h)=\sqrt{\kappa_0}/\rho_{\kappa_0}(h)\). Since
\(\rho_{\kappa_0}(h)\ge\sqrt{\kappa_0}\) is positive and increasing in
\(h\), \(\Phi_{\kappa_0}'\) is positive and nonincreasing, so
\(\Phi_{\kappa_0}\) is increasing and concave.

\emph{Comparison with exploration.}
Evaluating the cubic-floor lower bound above at
\(h=\twonorm{\widehat\Theta_{t-1}}\) gives
\(p_t\ge\kappa_0/\rho_{\kappa_0}(\twonorm{\widehat\Theta_{t-1}})^2=\varpi_t^2\),
proving the first bound in \eqref{eq:growth-Phi-depth}.

\emph{Signed drift.}
Lemma~\ref{lem:qix:bias} gives
\(\E_t\iprod{e_t}{\widehat\theta_t}=(1-\lambda_tp_t)\iprod{e_t}{\theta_t}\) with \(0\le1-\lambda_tp_t\le1\).
Using \eqref{eq:state-update} with \(\epsilon_t\ge0\), \(\varpi_t\le\sqrt{p_t}\), and
\(\normt{G_t}{\theta_t}^2=\twonorm{P_{\perp,t}\theta_t}^2+p_t\iprod{e_t}{\theta_t}^2\),
\begin{align*}
 \varpi_t\iprod{e_t}{\widehat\Theta_{t-1}-\E_t\widehat\Theta_t}
 &\le-\varpi_t\E_t\iprod{e_t}{\widehat\theta_t}
  =-\varpi_t(1-\lambda_tp_t)\iprod{e_t}{\theta_t}\\
 &\le\varpi_t(1-\lambda_tp_t)[-\iprod{e_t}{\theta_t}]_+
  \le\sqrt{p_t}\,|\iprod{e_t}{\theta_t}|
  \le\normt{G_t}{\theta_t}.
\end{align*}
Taking positive parts proves the second bound in \eqref{eq:growth-Phi-depth};
summation and Cauchy--Schwarz give \eqref{eq:growth-Phi-signed-mean}.
\end{proof}

\paragraph{Radial range and variance.}

Projecting onto the leader direction gives a sharper range bound than
the full Euclidean norm of the estimator. Its variance is controlled by
\(\normt{G_t}{\theta_t}^2\).

\begin{lemma}[Radial range and variance]
\label{lem:growth-radial-range}
Pathwise, the radial estimate satisfies \(|\iprod{e_t}{\widehat\theta_t}|\le\beta_t/\ell_{j_t}\le\beta_t/\ell_0\),
and its centred noise satisfies \(|\iprod{e_t}{\xi_t}|\le\beta_t/\ell_0+1\).
The conditional variance is bounded by the loss quadratic form:
\begin{equation}
 \E_t\iprod{e_t}{\xi_t}^2\le3\normt{G_t}{\theta_t}^2.
 \label{eq:growth-radial-range}
\end{equation}
\end{lemma}

\begin{proof}
\emph{Range.}
Since \(p_tG_t^{-1}=p_tP_{\perp,t}+P_{\parallel,t}\),
Lemma~\ref{lem:qix:range} and \(\ell_{j_t}\ge\ell_0\) give
\(\abs{\iprod{e_t}{\widehat\theta_t}}\le\normt{p_tG_t^{-1}}{\widehat{\theta}_t}\le\beta_t/\ell_{j_t}\le\beta_t/\ell_0\).
Projecting the bias identity onto \(e_t\) gives
\(\E_t\iprod{e_t}{\widehat\theta_t}=(1-\lambda_tp_t)\iprod{e_t}{\theta_t}\), whose absolute value is at
most \(1\). Hence \(\abs{\iprod{e_t}{\xi_t}}\le\beta_t/\ell_0+1\).

\emph{Variance.}
Write \(\chi_t=\iprod{e_t}{\theta_t}\) and \(v=G_t^{1/2}\theta_t\), and let
\[
 A=\frac{d\sqrt{p_t}\,\iprod{U_t}{e_t}\iprod{U_t}{v}}
          {p_t+\nu_t\iprod{U_t}{v}^2},
 \qquad
 A^\circ=\frac d{\sqrt{p_t}}\iprod{U_t}{e_t}\iprod{U_t}{v}
\]
be the radial coordinate and its unshrunk counterpart on an exploratory draw.
Centring at the boundary-branch value \(\chi_t\),
\[
 \E_t\iprod{e_t}{\xi_t}^2
 \le\E_t(\iprod{e_t}{\widehat\theta_t}-\chi_t)^2
 =p_t\E_U(A-\chi_t)^2.
\]
The bias identity
\((1-p_t)\chi_t+p_t\E_UA=(1-\lambda_tp_t)\chi_t\)
gives \(\E_UA=(1-\lambda_t)\chi_t\). Since \(|A|\le|A^\circ|\),
Lemma~\ref{lem:spherical-polynomial-moments}(b) paired with
\(e_te_t\transpose\) gives
\[
 p_t\E_UA^2
 \le p_t\E_U(A^\circ)^2
 =\frac d{d+2}\{\normt{G_t}{\theta_t}^2+2p_t\chi_t^2\}.
\]
Writing \(\theta_t=w_t+\chi_te_t\), with \(w_t\perp e_t\), we obtain
\begin{align*}
 \E_t\iprod{e_t}{\xi_t}^2
 &\le p_t\E_UA^2-2p_t\chi_t\E_UA+p_t\chi_t^2\\
 &\le\frac d{d+2}(\normt{G_t}{\theta_t}^2+2p_t\chi_t^2)
             +(-1+2\lambda_t)p_t\chi_t^2\\
 &=\frac d{d+2}\twonorm{w_t}^2
 +\left\{\frac{2(d-1)}{d+2}+2\lambda_t\right\}
 p_t\chi_t^2.
\end{align*}
Finally, \eqref{eq:qix-lambda-bound}, \(\beta_t^2p_t\ge\kappa_{j_t}\)
from \eqref{eq:cubic-support}, \(\normt{G_t}{\theta_t}\le1\), and
\(\kappa_{j_t}/\ell_{j_t}^2=\max\{1,(j_t+1)/d\}\kappa_0/\ell_0^2\ge\kappa_0/\ell_0^2\) imply
\[
 \lambda_t
 \le\frac{3d^2\ell_{j_t}^2}{4\kappa_{j_t}(d+2)}
 \le\frac{3d^2\ell_0^2}{4\kappa_0(d+2)}.
\]
Since \(2(d-1)/(d+2)\ge d/(d+2)\) for \(d\ge2\) and
\(\kappa_0/\ell_0^2=12d^2(d+3)/\{(d+4)(7d+6)\}\), both coefficients are at most
\(2(d-1)/(d+2)+(d+4)(7d+6)/\{8(d+2)(d+3)\}<2+7/8<3\). Using
\(\normt{G_t}{\theta_t}^2=\twonorm{w_t}^2+p_t\chi_t^2\) proves
\eqref{eq:growth-radial-range}.
\end{proof}

\paragraph{Weighted radial noise.}
\phantomsection
\label{sec:radial-event}
Controlling weighted radial noise on a path-selected suffix uses a union
bound over deterministic starts (at most \(T\)) and deterministic dyadic
range and variance grids of sizes
\begin{equation}
 J_{\mathrm{rad}}=1+\ceil*{T\log_23},
 \qquad
 L_{\mathrm{rad}}=1+\ceil{2\log_2T}.
 \label{eq:growth-logs}
\end{equation}
For \(T\ge1\) and an analysis level \(\bar\delta\in(0,1/8]\), put
\(A:=\log(T/\bar\delta)\ge\log8>1\) and
\(\ell_\Phi:=\log\{TJ_{\mathrm{rad}}L_{\mathrm{rad}}/(2\bar\delta)\}\).
Since \(\log_23<2\), \(J_{\mathrm{rad}}\le2T+1\) and
\(L_{\mathrm{rad}}\le C_0\log(eT)\), so
\begin{equation}
 \ell_\Phi\le\log(C_0T^3/\bar\delta)\le C_0A.
 \label{eq:ell-phi-bound}
\end{equation}

\begin{lemma}[Weighted radial event \(\mathcal E_{\mathrm{rad}}\)]
\label{lem:growth-martingale}
With probability at least \(1-4\bar\delta\), simultaneously for every integer
interval \(\varnothing\ne I\subseteq\{2,\ldots,T-1\}\),
\begin{equation}
 \begin{split}
 \left|\sum_{r\in I}\varpi_r\iprod{e_r}{\xi_r}\right|
 \le{}&\max\left\{
 2\sqrt{3\ell_\Phi
             \sum_{r\in I}\varpi_r^2\normt{G_r}{\theta_r}^2},\;
 2\Bigl(\max_{r\in I}\varpi_r\beta_r\Bigr)
       \sqrt{\frac{6\ell_\Phi}{T}}
 \right\}\\
 &+\frac13\ell_\Phi
 +\frac{2\ell_\Phi}{3\ell_0}\max_{r\in I}\varpi_r\beta_r.
 \end{split}
 \label{eq:growth-interval-martingale}
\end{equation}
\end{lemma}

\begin{proof}
Assume \(T\ge3\) (otherwise vacuous). By Lemma~\ref{lem:rate-jumps},
\(\beta_0\le\varpi_t\beta_t\le\beta_t<3^T\beta_0\) on \(2\le t\le T-1\),
so the grid \(b=B_j\) (\(0\le j<J_{\mathrm{rad}}\)) covers every
realised value of \(\varpi_t\beta_t\). For each deterministic start
\(m\in\{1,\ldots,T-2\}\), range level \(b=B_j\), and variance endpoint
\(v_{b,k}=2^kb^2/T\) (\(0\le k<L_{\mathrm{rad}}\)), put
\(X_r=\varpi_r\iprod{e_r}{\xi_r}\),
\(\overline\sigma_r^2=3\varpi_r^2\normt{G_r}{\theta_r}^2\), and
\(k_r=\II{r>m,\ \varpi_r\beta_r\le b}\). By
Lemma~\ref{lem:growth-radial-range}, \(\ell_{j_r}\ge\ell_0\), and
\(\varpi_r\le1\), we have \(|k_rX_r|\le1+b/\ell_0\) and
\(\E_rX_r^2\le\overline\sigma_r^2\). Applying the two-sided form of
Lemma~\ref{lem:masked-freedman} with range \(1+b/\ell_0\), variance cap
\(3v_{b,k}\), and level \(\ell_\Phi\) shows that, except with probability
at most \(2e^{-\ell_\Phi}\), every \(\tau\in\{m+1,\ldots,T-1\}\) with
\(\sum_{r=m+1}^\tau k_r\overline\sigma_r^2\le3v_{b,k}\) satisfies
\begin{equation}
 \left|\sum_{r=m+1}^{\tau}k_r\varpi_r\iprod{e_r}{\xi_r}\right|
 \le\sqrt{6v_{b,k}\ell_\Phi}
      +\frac13(1+b/\ell_0)\ell_\Phi.
 \label{eq:growth-masked}
\end{equation}
A union bound over at most \(TJ_{\mathrm{rad}}L_{\mathrm{rad}}\) triples
costs at most \(2TJ_{\mathrm{rad}}L_{\mathrm{rad}}e^{-\ell_\Phi}=4\bar\delta\).

On this event, fix \(I=\{m+1,\ldots,\tau\}\), and put
\(W=\sum_{r\in I}\varpi_r^2\normt{G_r}{\theta_r}^2\) and
\(M=\max_{r\in I}\varpi_r\beta_r\). Pick the least \(b\ge M\) (so \(b<2M\)
and \(k_r=1\) on \(I\)). Since \(W\le|I|<T\le Tb^2\le v_{b,L_{\mathrm{rad}}-1}\),
pick the least \(v_{b,k}\ge W\); then \(v_{b,k}\le\max\{2W,b^2/T\}\), so
\eqref{eq:growth-masked} applies and \(b<2M\) yields
\eqref{eq:growth-interval-martingale}.
\end{proof}

\subsection{Potential recursion after the last low-norm round}

Fix a trial scale \(s\). We examine the part of the trajectory after
the state norm last lay below \(s\), while all earlier scales remain
below \(s\). The entry step has bounded overshoot; afterwards, the
norm is large enough to charge rotation to observed squares.

The rotation residual
\(\mathfrak R_t=\twonorm{\widehat\Theta_t}-(1-\epsilon_t)\twonorm{\widehat\Theta_{t-1}}+\iprod{e_t}{\widehat{\theta}_t}\)
measures the norm increase beyond the radial component of the update.

The next lemma holds pathwise at any fixed reference level \(s\) for any
increasing concave potential.

\begin{lemma}[Potential recursion on a high-norm suffix]
\label{lem:growth-suffix}
Fix a round \(t\) and a number \(s\) such that
\(\beta_r<s\) for every \(r<t\), and suppose \(\twonorm{\widehat\Theta_{t-1}}\ge s\).  Let \(k\le t\)
be one plus the last index \(r<t\) with \(\twonorm{\widehat\Theta_{r-1}}<s\), assuming this set is
nonempty, and put \(I=\{k,\ldots,t-1\}\).
The entrance overshoot is bounded pathwise by
\begin{equation}
 \twonorm{\widehat\Theta_{k-1}}\le2s.
 \label{eq:growth-entry-overshoot}
\end{equation}
On the suffix, observed squares bound the rotation residual:
\begin{equation}
 0\le\mathfrak R_r\le4\frac{p_r\normt{G_r}{\widehat\theta_r}^2}{\beta_r}
 \qquad(r\in I).
 \label{eq:growth-rotation-clock}
\end{equation}
Moreover, for every concave, increasing, differentiable
\(\Phi\colon[0,\infty)\to[0,\infty)\), writing \(\varpi_r=\Phi'(\twonorm{\widehat\Theta_{r-1}})\) and
\(\omega=\Phi'(s)\),
\begin{equation}
 \Phi(\twonorm{\widehat\Theta_{t-1}})\le\Phi(\twonorm{\widehat\Theta_{k-1}})
 -\sum_{r\in I}\varpi_r\{\epsilon_r\twonorm{\widehat\Theta_{r-1}}+\E_r\iprod{e_r}{\widehat\theta_r}\}
 -\sum_{r\in I}\varpi_r\iprod{e_r}{\xi_r}
 +4\omega\sum_{r\in I}\frac{p_r\normt{G_r}{\widehat\theta_r}^2}{\beta_r}.
 \label{eq:growth-potential-recursion}
\end{equation}
\end{lemma}
In every application below the set contains \(r=1\), since \(\twonorm{\widehat\Theta_0}=0\).

\begin{proof}
\emph{Step 1: entrance overshoot.}
Put \(r_0=k-1\), so \(\twonorm{\widehat\Theta_{r_0-1}},\beta_{r_0}<s\).
By Lemma~\ref{lem:qix:range}, \(\ell_{j_r}\ge\ell_0\), and
\(p_rG_r^{-1}\succeq p_rI_d\),
\(\twonorm{\widehat\theta_r}\le\beta_r/(\ell_0\sqrt{p_r})
 =\ell_0^{-1}\sqrt{\beta_r(\twonorm{\widehat\Theta_{r-1}}+\beta_r)}\).
Since \(0\le\epsilon_{r_0}\le1/4\) and \(1/\ell_0<0.4\),
\(\twonorm{\widehat\Theta_{k-1}}
 \le\twonorm{\widehat\Theta_{r_0-1}}+\twonorm{\widehat\theta_{r_0}}
 <(1+\sqrt2/\ell_0)s<2s\),
proving \eqref{eq:growth-entry-overshoot}. If \(k=t\), the suffix is empty
and the remaining claims are immediate; assume \(k<t\).

\emph{Step 2: tangential remainder.}
Write \(\widehat\theta_r=\iprod{e_r}{\widehat\theta_r}e_r+u_r\) with
\(u_r\perp e_r\). From \eqref{eq:state-update} and
\(\widehat\Theta_{r-1}=-\twonorm{\widehat\Theta_{r-1}}e_r\),
\begin{equation}
 \twonorm{\widehat\Theta_r}
 =\sqrt{\{(1-\epsilon_r)\twonorm{\widehat\Theta_{r-1}}-\iprod{e_r}{\widehat\theta_r}\}^2+\twonorm{u_r}^2}.
 \label{eq:growth-rotation}
\end{equation}
For \(r\in I\), \(\twonorm{\widehat\Theta_{r-1}}\ge s>\beta_r\), so
Lemma~\ref{lem:growth-radial-range}, \(\epsilon_r\le1/4\), and
\(1/\ell_0<0.4\) give
\((1-\epsilon_r)\twonorm{\widehat\Theta_{r-1}}-\iprod{e_r}{\widehat\theta_r}
 >(3/4-1/\ell_0)\twonorm{\widehat\Theta_{r-1}}>0\).
Rationalising \eqref{eq:growth-rotation} yields
\(0\le\mathfrak R_r\le\twonorm{u_r}^2/\{(3/2-2/\ell_0)\twonorm{\widehat\Theta_{r-1}}\}\).
Since \(p_r\twonorm{u_r}^2\le p_r\normt{G_r}{\widehat\theta_r}^2\),
\(p_r\twonorm{\widehat\Theta_{r-1}}=\beta_r(1-p_r)\), and \(1-p_r>1/2\),
we obtain
\(\mathfrak R_r\le2(3/2-2/\ell_0)^{-1}p_r\normt{G_r}{\widehat\theta_r}^2/\beta_r
 \le4p_r\normt{G_r}{\widehat\theta_r}^2/\beta_r\),
proving \eqref{eq:growth-rotation-clock}.

\emph{Step 3: telescope the potential.}
Concavity of \(\Phi\) gives
\[
 \Phi(\twonorm{\widehat\Theta_r})-\Phi(\twonorm{\widehat\Theta_{r-1}})
 \le\varpi_r(\twonorm{\widehat\Theta_r}-\twonorm{\widehat\Theta_{r-1}})
 =-\varpi_r\{\epsilon_r\twonorm{\widehat\Theta_{r-1}}+\E_r\iprod{e_r}{\widehat\theta_r}\}
   -\varpi_r\iprod{e_r}{\xi_r}+\varpi_r\mathfrak R_r.
\]
On \(I\), \(\twonorm{\widehat\Theta_{r-1}}\ge s\) implies
\(0\le\varpi_r\le\omega\), so Step~2 bounds \(\varpi_r\mathfrak R_r\) by
\(4\omega p_r\normt{G_r}{\widehat\theta_r}^2/\beta_r\). Summing over
\(r\in I\) yields \eqref{eq:growth-potential-recursion}.
\end{proof}

\paragraph{Inequalities at update rounds.}

An active cubic-floor update meets the hypotheses of
Lemma~\ref{lem:growth-suffix}, except in the low-norm case, where its
scale is already bounded. The scale rule also bounds the observed
squares accumulated before that update.

\begin{lemma}[Bounds at cubic-floor updates]
\label{lem:growth-binding-scaffold}
Fix a cubic-floor-active round \(t\) and set
\(s=\allowbreak\rho_{\kappa_{j_t}}(\twonorm{\widehat\Theta_{t-1}})\).
Then \(\beta_r<s\) for all \(r<t\), \(\beta_t=s\),
\(\twonorm{\widehat\Theta_{t-1}}=(s^3-\kappa_{j_t}^{3/2})/(2\kappa_{j_t})\), and
\begin{equation}
 \sum_{r=1}^{t-1}\frac{p_r\normt{G_r}{\widehat\theta_r}^2}{\beta_r}
 \le\frac{3s}{a_{j_t}^2}.
 \label{eq:growth-lagged-at-binding}
\end{equation}
Whenever \(s^2\ge4\kappa_{j_t}\), we have
\(\twonorm{\widehat\Theta_{t-1}}\ge7s^3/(16\kappa_{j_t})>s\) and
Lemma~\ref{lem:growth-suffix} applies at \((t,s)\) with \(k\ge2\) and
\(I\subseteq\{2,\ldots,T-1\}\).
\end{lemma}

\begin{proof}
\emph{Floor activity.}
The definition in \eqref{eq:support-binding} gives
\(s>\max\{\beta_{t-1},\allowbreak a_{j_t}\sqrt{\EestObs_{t-1}+1}\}\), so the scale
rule sets \(\beta_t=s\). The scale is nondecreasing, so
\(\beta_r\le\beta_{t-1}<s\) for every \(r<t\). The identity
\(s^3=\kappa_{j_t}^{3/2}+2\kappa_{j_t}\twonorm{\widehat\Theta_{t-1}}\) gives
\(\twonorm{\widehat\Theta_{t-1}}=(s^3-\kappa_{j_t}^{3/2})/(2\kappa_{j_t})\).
If \(s^2\ge4\kappa_{j_t}\), then \(\kappa_{j_t}^{3/2}\le s^3/8\), so
\(\twonorm{\widehat\Theta_{t-1}}\ge7s^3/(16\kappa_{j_t})\ge7s/4>s\); since
\(\twonorm{\widehat\Theta_0}=0<s\), the set of low-norm indices is
nonempty, and Lemma~\ref{lem:growth-suffix} applies with \(k\ge2\) and
\(I\subseteq\{2,\ldots,T-1\}\).

\emph{The lagged square.}
If \(t=1\), the sum is empty. For \(t\ge2\), Lemma~\ref{lem:lagged-clock}
at prefix \(t-1\), strict floor activity \eqref{eq:support-binding}, and
\(j_{t-1}\le j_t\) (so \(a_{j_{t-1}}\ge a_{j_t}\)) give
\[
 \sum_{r=1}^{t-1}\frac{p_r\normt{G_r}{\widehat\theta_r}^2}{\beta_r}
 \le\frac{3}{a_{j_{t-1}}}\sqrt{\EestObs_{t-1}+1}
 <\frac{3}{a_{j_{t-1}}}\,\frac{s}{a_{j_t}}
 \le\frac{3s}{a_{j_t}^2},
\]
proving \eqref{eq:growth-lagged-at-binding}.
\end{proof}

\subsection{Interval drift and active scale updates}

Recall \(\mathsf M_T\) and \(\mathsf E_T\) from
\eqref{eq:intro-net-drawup}. By \eqref{eq:state-update} and
Lemma~\ref{lem:qix:bias},
\(\varpi_t\iprod{e_t}{\widehat\Theta_{t-1}-\E_t\widehat\Theta_t}
 =-\varpi_t\{\epsilon_t\twonorm{\widehat\Theta_{t-1}}+(1-\lambda_tp_t)\iprod{e_t}{\theta_t}\}\),
so \(\mathsf M_T\) is the largest weighted interval drift after crediting
inward increments and contractions (matching the suffix sum of
Lemma~\ref{lem:growth-suffix}; zero for \(T\le2\)). A loss with
\(\iprod{e_t}{\theta_t}<0\) contributes outward drift even when loss
directions never change.

By \eqref{eq:growth-Phi-signed-mean} and \eqref{eq:intro-net-drawup},
\begin{equation}
 0\le\mathsf M_T
 \le\sqrt{[T-2]_+\sum_{t=2}^{T-1}\normt{G_t}{\theta_t}^2},
 \qquad
 \mathsf M_T\le\sqrt{\kappa_0}\,(\Qcurv_T+1)+\mathsf E_T.
 \label{eq:growth-net-drawup-relaxation}
\end{equation}

\paragraph{Scale at update rounds.}

At an active update with tuning index \(j\ge0\) (so \(\kappa_j=q_j\kappa_0\)),
the fixed potential \(\Phi_{\kappa_0}\) scales as
\(s^2/(q_j^{2/3}\sqrt{\kappa_0})\). Collect the noise and rotation coefficients,
with \(\ell_\Phi\) from Appendix~\ref{sec:radial-event}, into
\[
 B_{\Phi}(j)
 =2\sqrt{\frac{6\ell_\Phi}{T}}
 +\frac{2\ell_\Phi}{3\ell_0}
 +\frac{12}{a_j^2}.
\]
Let \(U_{\Phi,T}(j)\) be the largest of the following four thresholds,
noting that \(\sqrt{\kappa_0}\,q_j^{1/2}=\sqrt{\kappa_j}\) and
\(\kappa_0^{2/7}q_j^{2/7}=\kappa_j^{2/7}\):
\begin{lemenum}
\item \emph{Confidence term:}
 \(\sqrt{\kappa_0}\,q_j^{1/3}\sqrt{8\ell_\Phi/(3\sqrt{\kappa_0})}\).
\item \emph{Entrance and rotation costs:}
 \(\sqrt{\kappa_j}\,\{8(2+B_\Phi(j))\}^{3/4}\).
\item \emph{Net interval drift:}
 \(\min\bigl\{q_j^{1/3}\kappa_0^{1/4}\sqrt{8\mathsf M_T},\;
  \kappa_j^{2/7}(8T)^{3/7}\bigr\}\).
\item \emph{Radial variance:}
 \(\kappa_j^{2/7}
 \bigl(16\sqrt{3\ell_\Phi\sum_{t=2}^{T-1}\normt{G_t}{\theta_t}^2}\bigr)^{3/7}\).
\end{lemenum}
Since \(q_j\ge1\), the second entry is at least
\(\sqrt{\kappa_j}\,16^{3/4}=8\sqrt{\kappa_j}>2\sqrt{\kappa_j}\), and
\(U_{\Phi,T}(j)\) is nondecreasing in \(j\).

\begin{proposition}[Scale bound at cubic-floor updates]
\label{thm:growth-Phi-cap}
On \(\mathcal E_{\mathrm{rad}}\) (Lemma~\ref{lem:growth-martingale}),
simultaneously for every cubic-floor-active round \(t\le T\),
\begin{equation}
 \rho_{\kappa_{j_t}}(\twonorm{\widehat\Theta_{t-1}})\le U_{\Phi,T}(j_t),
 \qquad\text{hence}\qquad
 \rho_{\mathrm{bind},T}\le U_{\Phi,T}(j(\beta_T)).
 \label{eq:growth-Phi-cap}
\end{equation}
\end{proposition}

\begin{proof}
Work pathwise on \(\mathcal E_{\mathrm{rad}}\).

\emph{Step 1: separate the low-norm case.}
Fix an active round \(t\), put \(j=j_t\), \(q=q_j\),
\(s=\rho_{\kappa_j}(\twonorm{\widehat\Theta_{t-1}})\), and \(x=s/\sqrt{\kappa_0}\).
By Lemma~\ref{lem:growth-binding-scaffold}, either
\(s<2\sqrt{\kappa_j}\le U_{\Phi,T}(j)\), or \(s^2\ge4\kappa_j\), in which case
\(\twonorm{\widehat\Theta_{t-1}}\ge7s^3/(16\kappa_j)>s\) and
Lemma~\ref{lem:growth-suffix} applies with suffix
\(I=\{k,\ldots,t-1\}\subseteq\{2,\ldots,T-1\}\). Assume the latter.

\emph{Step 2: apply the potential recursion.}
Apply Lemma~\ref{lem:growth-suffix} with the fixed potential
\(\Phi=\Phi_{\kappa_0}\), which is concave, increasing, and differentiable by
Lemma~\ref{lem:growth-Phi-geometry}.
Set \(\omega=\Phi_{\kappa_0}'(s)=\sqrt{\kappa_0}/\rho_{\kappa_0}(s)\). By the state recursion, the mean term in
\eqref{eq:growth-potential-recursion} is
\(\varpi_r\iprod{e_r}{\widehat\Theta_{r-1}-\E_r\widehat\Theta_r}\). The entrance bound
\(\twonorm{\widehat\Theta_{k-1}}\le2s\) and monotonicity of \(\Phi_{\kappa_0}\) give
\begin{equation}
 \Phi_{\kappa_0}(\twonorm{\widehat\Theta_{t-1}})
 \le\Phi_{\kappa_0}(2s)
 +\sum_{r\in I}\varpi_r\iprod{e_r}{\widehat\Theta_{r-1}-\E_r\widehat\Theta_r}
 -\sum_{r\in I}\varpi_r\iprod{e_r}{\xi_r}
 +4\omega\sum_{r\in I}\frac{p_r\normt{G_r}{\widehat\theta_r}^2}{\beta_r}.
 \label{eq:growth-Phi-recursion}
\end{equation}
If \(I=\varnothing\), all suffix sums vanish. Otherwise
\(I\subseteq\{2,\ldots,T-1\}\). By definition of \(\mathsf M_T\), the
drift sum is at most \(\mathsf M_T\). Moreover, on \(I\),
\(\twonorm{\widehat\Theta_{r-1}}\ge s\) implies \(\varpi_r\le\omega\) and
Lemma~\ref{lem:growth-Phi-geometry} gives
\(\iprod{e_r}{\widehat\Theta_{r-1}-\E_r\widehat\Theta_r}\le(1-\lambda_rp_r)[-\iprod{e_r}{\theta_r}]_+\le1\),
so the same drift sum is also at most \(\omega|I|\le\omega T\). Thus
\[
 \sum_{r\in I}\varpi_r\iprod{e_r}{\widehat\Theta_{r-1}-\E_r\widehat\Theta_r}
 \le\min\{\mathsf M_T,\;\omega T\}.
\]
Also, \eqref{eq:growth-lagged-at-binding} gives
\(\sum_{r\in I}p_r\normt{G_r}{\widehat\theta_r}^2/\beta_r\le3s/a_j^2\).

\emph{Step 3: bound the noise on the selected suffix.}
Let \(I\ne\varnothing\), so \(T\ge3\). For \(r\in I\), \(\twonorm{\widehat\Theta_{r-1}}\ge s\) and
\(\beta_r<s\). Since \(\Phi_{\kappa_0}'\) is nonincreasing,
\(\varpi_r\le\omega\), and therefore
\(\sum_{r\in I}\varpi_r^2\normt{G_r}{\theta_r}^2\le\omega^2\sum_{t=2}^{T-1}\normt{G_t}{\theta_t}^2\) and
\(\max_{r\in I}\varpi_r\beta_r\le\omega s\). The weighted event
controls this interval although its endpoints were selected from the
path, so \eqref{eq:growth-interval-martingale} and \(\max\{a,b\}\le a+b\) give
\[
 -\sum_{r\in I}\varpi_r\iprod{e_r}{\xi_r}
 \le\frac{\ell_\Phi}{3}
 +\omega\left\{
 2\sqrt{3\sum_{t=2}^{T-1}\normt{G_t}{\theta_t}^2\ell_\Phi}
 +2s\sqrt{\frac{6\ell_\Phi}{T}}
 +\frac{2s\ell_\Phi}{3\ell_0}\right\}.
\]

\emph{Step 4: bound the potential.}
Since \(\twonorm{\widehat\Theta_{t-1}}\ge7s^3/(16\kappa_j)=7s^3/(16q\kappa_0)\)
and \(x=s/\sqrt{\kappa_0}\ge2\sqrt q\ge2\),
\[
 \frac{\rho_{\kappa_0}(\twonorm{\widehat\Theta_{t-1}})^3}{\kappa_0^{3/2}}
 =1+\frac{2\twonorm{\widehat\Theta_{t-1}}}{\sqrt{\kappa_0}}
 \ge1+\frac{7x^3}{8q}
 \ge8.
\]
Thus \(\rho_{\kappa_0}(\twonorm{\widehat\Theta_{t-1}})^2\ge4\kappa_0\),
so \(\rho_{\kappa_0}(\twonorm{\widehat\Theta_{t-1}})^2-\kappa_0\ge\frac34\rho_{\kappa_0}(\twonorm{\widehat\Theta_{t-1}})^2\),
and \eqref{eq:overview-potential} gives (using \(\frac9{16}(7/8)^{2/3}>1/2\))
\[
 \Phi_{\kappa_0}(\twonorm{\widehat\Theta_{t-1}})
 \ge\frac{9}{16\sqrt{\kappa_0}}\rho_{\kappa_0}(\twonorm{\widehat\Theta_{t-1}})^2
 \ge\frac{9}{16}\Bigl(\frac78\Bigr)^{2/3}\frac{\sqrt{\kappa_0}\,x^2}{q^{2/3}}
 \ge\frac{\sqrt{\kappa_0}\,x^2}{2q^{2/3}}.
\]
For the entrance, \eqref{eq:overview-potential},
\((1+4x)^{2/3}\le1+(4x)^{2/3}\), and \(3\cdot4^{2/3}/4<2\) give
\[
 \Phi_{\kappa_0}(2s)
 =\frac{3\sqrt{\kappa_0}}4\bigl\{(1+4x)^{2/3}-1\bigr\}
 \le\frac{3\cdot4^{2/3}}4\sqrt{\kappa_0}\,x^{2/3}
 \le2\sqrt{\kappa_0}\,x^{2/3}.
\]
Finally, \(\rho_{\kappa_0}(s)=(\kappa_0^{3/2}+2\kappa_0s)^{1/3}\ge(\kappa_0s)^{1/3}=\sqrt{\kappa_0}x^{1/3}\)
gives \(\omega\le x^{-1/3}\).

\emph{Step 5: solve the scalar inequality.}
Insert Steps~2--4 in \eqref{eq:growth-Phi-recursion}, divide by
\(\sqrt{\kappa_0}\), and multiply by \(q^{2/3}\). The terms proportional
to \(s=\sqrt{\kappa_0}x\) inside the braces of Step~3, together with the rotation
term, contribute at most \(\omega B_\Phi(j)\sqrt{\kappa_0}x\le B_\Phi(j)\sqrt{\kappa_0}x^{2/3}\).
Hence
\begin{align*}
 \frac{x^2}{2}\le q^{2/3}\Biggl\{
 &\frac{\ell_\Phi}{3\sqrt{\kappa_0}}
 +(2+B_\Phi(j))x^{2/3}\\
 &+\min\Bigl\{\frac{\mathsf M_T}{\sqrt{\kappa_0}},\;\frac{T}{\sqrt{\kappa_0}}x^{-1/3}\Bigr\}
 +2\sqrt{\frac{3\sum_{t=2}^{T-1}\normt{G_t}{\theta_t}^2\ell_\Phi}{\kappa_0}}\,
   x^{-1/3}\Biggr\};
\end{align*}
for \(I=\varnothing\) the same inequality holds, since the terms dropped
there are nonnegative.

At least one of the four summands on the right is at least \(x^2/8\).
Solving for \(x\) in each case, taking either
\(\mathsf M_T/\sqrt{\kappa_0}\) or \((T/\sqrt{\kappa_0})x^{-1/3}\) as the third summand,
and multiplying by \(\sqrt{\kappa_0}\) gives
\(s\le U_{\Phi,T}(j)\), since
\(\sqrt{\kappa_0}(q^{2/3})^{3/4}=\sqrt{\kappa_j}\) and
\(\sqrt{\kappa_0}(q^{2/3}/\sqrt{\kappa_0})^{3/7}=\kappa_j^{2/7}\).
The bound holds simultaneously over all active rounds \(t\le T\) (including
\(T\le2\), where \([T-2]_+=\mathsf M_T=0\) and every active suffix is empty),
and \(j_t\le j(\beta_T)\) with monotonicity of \(U_{\Phi,T}(\cdot)\) yields
\(\rho_{\mathrm{bind},T}\le U_{\Phi,T}(j(\beta_T))\).
\end{proof}

\subsection{Final regularisation scale}

\begin{lemma}[Final regularisation scale]
\label{lem:final-scale}
Fix \(T\ge1\) and \(\bar\delta\in(0,1/8]\), and set
\(A=\log(T/\bar\delta)\) and
\(U_j:=\{1+A/\ell_0+\sqrt{A/T}+(j+1)/\ell_j\}^{3/4}\). On
\(\mathcal E_{\mathrm{up}}\cap\mathcal E_{\mathrm{rad}}\), with
\(j=j(\beta_T)\),
\[
 \begin{aligned}
 \beta_T\le C_0\max\Bigl\{
 &\sqrt{\kappa_j}\,U_j,\;
 a_j\sqrt{d(1+\Qcurv_T)},\\
 &\min\bigl\{q_j^{1/3}\kappa_0^{1/4}\sqrt{\mathsf M_T},\;
 \kappa_j^{2/7}T^{3/7}\bigr\},\;
 \kappa_j^{2/7}\bigl\{A(1+\Qcurv_T)\bigr\}^{3/14}\Bigr\},
 \end{aligned}
\]
and \(j+1\le C_0\min\{T,\log(eTA)\}\).
\end{lemma}

\begin{proof}
Write \(j=j(\beta_T)\) and \(q=q_j\).
Combine Lemma~\ref{lem:scale-envelope} with Proposition~\ref{thm:growth-Phi-cap}:
\[
 \beta_T\le\max\Bigl\{U_{\Phi,T}(j),\;
 \sqrt6\,a_j\sqrt{\tfrac32d\Qcurv_{T-1}+1}\Bigr\}.
\]
By \eqref{eq:ell-phi-bound} and \(a_j^{-2}=48(j+1)/\ell_j\), we have
\(B_\Phi(j)\le C_0(\sqrt{A/T}+A/\ell_0+(j+1)/\ell_j)\).
Also \(\kappa_0\ge2\ell_0^2\) gives \(\ell_\Phi/\sqrt{\kappa_0}\le C_0A/\ell_0\), and
\(q^{1/3}\le q^{1/2}\), so the first two entries of \(U_{\Phi,T}(j)\)
are at most
\(C_0\sqrt{\kappa_j}(1+A/\ell_0+\sqrt{A/T}+(j+1)/\ell_j)^{3/4}=C_0\sqrt{\kappa_j}\,U_j\).
The third and fourth entries of \(U_{\Phi,T}(j)\) and the second-order
entry \(a_j\sqrt{9d\Qcurv_T+6}\le3a_j\sqrt{d(1+\Qcurv_T)}\)
match the remaining entries, proving the first bound.

For the bound on \(j+1\), put \(x=\beta_T/\beta_0\ge1\), so
\(j+1\le1+\log_2x\). Since \(\ell_j=\ell_0+2\log(j+1)\le\ell_0(j+1)\)
(as \(\ell_0>2\)), we have \(q\le(j+1)^3\), \(a_j\le a_0<\beta_0\), and
\(B_\Phi(j)\le C_0(A+j+1)\); dividing the first bound by
\(\beta_0=\sqrt{\kappa_0}\) and using \(\Qcurv_T\le T\), \(\kappa_0\ge1\),
and \(A\ge1\) yields
\(x\le C_0(j+1)^3(A+j+1)(T+1)^{1/2}\le C_0(1+\log_2x)^4A\sqrt{T+1}\).
Since \((1+\log_2x)^4\le C_0x^{1/2}\) for all \(x\ge1\), this gives
\(x\le C_0A^2(T+1)\), so \(j+1\le1+\log_2x\le C_0\log(eTA)\).
Together with \(j(\beta_T)\le\floor{(T-1)\log_23}\le2(T-1)\) from
Lemma~\ref{lem:rate-jumps}, this gives
\(j+1\le C_0\min\{T,\log(eTA)\}\).
\end{proof}

\section{Proof of the main regret bound}
\label{sec:assembly}

We now bound the sampling noise, intersect the five concentration events
established in Appendices~\ref{sec:bandit-error}--\ref{sec:assembly}, and
combine Appendices~\ref{sec:local}, \ref{sec:bandit-error},
and~\ref{sec:radial-growth} to prove Proposition~\ref{prop:main-reduction},
Theorem~\ref{thm:headline}, and Theorem~\ref{thm:worst-case}.

\subsection{Sampling noise and common confidence event}
\label{sec:analysis-events}
\label{sec:sampling-error}

Fix the \(\delta\)-tuned run of Algorithm~\ref{alg:radial-qix}
(Appendix~\ref{sec:public-tuning}), an integer horizon \(T\ge1\), and an
analysis level \(0<\bar\delta\le\delta/8\); the learner keeps
\(\delta_0=\delta/8\), while \(\bar\delta\) enters only
\(\mathcal E_{\mathrm{play}}\) and \(\mathcal E_{\mathrm{rad}}\).
Put \(A:=\log(T/\bar\delta)\ge\log8>1\) and
\begin{equation}
 \Lambda:=\log\frac{1+\ceil{\log_2T}}{\bar\delta}\ge\log8>1.
 \label{eq:exact-logarithms}
\end{equation}
The conditional sampling variance on a round is
\begin{equation}
 \operatorname{Var}_t\!\left(\iprod{a_t-z_t}{\theta_t}\right)
 =p_t(1-p_t)\iprod{e_t}{\theta_t}^2+\frac{p_t}{d}\normt{G_t}{\theta_t}^2\le \normt{G_t}{\theta_t}^2.
 \label{eq:exact-sampling-clocks}
\end{equation}

\begin{lemma}[Sampling event \(\mathcal E_{\mathrm{play}}\)]
\label{lem:played}
With probability at least \(1-\bar\delta\), simultaneously for every
\(1\le\tau\le T\),
\begin{equation}
 \sum_{t=1}^\tau\iprod{a_t-z_t}{\theta_t}
 \le2\sqrt{(1+\Qcurv_\tau)\Lambda}
 +\frac{\sqrt2}{3}\Lambda.
 \label{eq:played-bound}
\end{equation}
\end{lemma}

\begin{proof}
The increments are centred and bounded in absolute value by \(\sqrt2\): on
a boundary draw \(\twonorm{a_t-z_t}=p_t\le1\), and on an exploratory draw,
with \(x=\iprod{U_t}{e_t}\),
\[
 \twonorm{G_t^{1/2}U_t-(1-p_t)e_t}^2
 =1+(1-p_t)^2-(1-p_t)x^2-2(1-p_t)\sqrt{p_t}\,x
 \le2-p_t\le2
\]
(maximised at \(x=-\sqrt{p_t}\)). By \eqref{eq:exact-sampling-clocks}, the
conditional variances satisfy \(\sigma_t^2\le\normt{G_t}{\theta_t}^2\).
Apply Lemma~\ref{lem:masked-freedman} with \(R=\sqrt2\),
\(\sigma_t^2\), \(k_t\equiv1\), and \(\ell=\Lambda\) on the dyadic grid
\(\{2^k:0\le k\le\ceil{\log_2T}\}\); the union over \(1+\ceil{\log_2T}\)
levels costs \(\bar\delta\). At any \(\tau\le T\), the least grid level
\(v_\tau\) covering \(\sum_{t=1}^\tau\sigma_t^2\le\Qcurv_\tau\) satisfies
\(v_\tau\le2(1+\Qcurv_\tau)\) and leaves every round unmasked through
\(\tau\), giving \eqref{eq:played-bound}.
\end{proof}

\begin{definition}[Common confidence event]
\label{def:common-event}
The \emph{common confidence event} is
\(\mathcal E:=\mathcal E_{\mathrm{up}}\cap\mathcal E_{\mathrm{rev}}
  \cap\mathcal E_{\mathrm{vec}}\cap\mathcal E_{\mathrm{play}}
  \cap\mathcal E_{\mathrm{rad}}\), intersecting the five events proved in
Lemmas~\ref{lem:shell-reverse-clock}, \ref{lem:lifted-uniform},
\ref{lem:upper-clock}, \ref{lem:growth-martingale}, and~\ref{lem:played}.
By a union bound, \(\bbP(\mathcal E^c)\le3\delta/8+5\bar\delta\).
\end{definition}

\subsection{Fixed-horizon regret bound}

\begin{proposition}[Fixed-horizon regret bound]
\label{prop:fixed-horizon}
On \(\mathcal E\), with \(j=j(\beta_T)\) and \(U_j\) as in
Lemma~\ref{lem:final-scale},
\begin{equation}
 \begin{split}
 \Reg_T\le C_0\Bigl\{
 &(j+1)\Bigl(\sqrt{\kappa_j}\,U_j
  +\min\bigl\{q_j^{1/3}\kappa_0^{1/4}\sqrt{\mathsf M_T},\;\kappa_j^{2/7}T^{3/7}\bigr\}
  +\kappa_j^{2/7}\bigl\{A(1+\Qcurv_T)\bigr\}^{3/14}\Bigr)\\
 &+\sqrt{(j+1)\ell_jd(1+\Qcurv_T)}
  +\sqrt{\Lambda(1+\Qcurv_T)}
  +\Lambda\Bigr\}.
 \end{split}
 \label{eq:growth-readable-finite}
\end{equation}
The same bound holds on \(\mathcal E\) at every prefix \(\tau\le T\),
replacing \((T,\Reg_T,\Qcurv_T,\mathsf M_T,j(\beta_T))\) by
\((\tau,\Reg_\tau,\Qcurv_\tau,\mathsf M_\tau,j(\beta_\tau))\) while
keeping \(A,\Lambda\) (and \(T\) inside \(U_j\)) fixed.
\end{proposition}

\begin{proof}
Work on \(\mathcal E\), fix any unit comparator \(v\), and set
\(j=j(\beta_T)\).

\emph{Step~1: FTRL decomposition.}
Combining \eqref{eq:regret-decomposition}, the pathwise FTRL bound
\eqref{eq:ftrl-pathwise}, and Lemma~\ref{lem:terminal-cost} gives
\[
 \begin{split}
 \sum_{t=1}^T\iprod{a_t-v}{\theta_t}
 \le{}&\beta_T\log(\beta_T^2/\kappa_0+1/\ell_0)
 +\sum_{t=1}^T\frac{p_t\normt{G_t}{\widehat\theta_t}^2}{\beta_t}
 +\sum_{t=1}^T\iprod{a_t-z_t}{\theta_t}\\
 &+\sum_{t=1}^T\left\{\iprod{z_t-v}{\theta_t-\widehat\theta_t}
                    -\epsilon_t\Gamma_{t-1}(v)\right\}
 +\sum_{t=1}^T\II{\epsilon_t>0}\beta_t(1-p_t).
 \end{split}
\]

\emph{Step~2: estimation error and sampling noise.}
Lemma~\ref{lem:bandit-correction} bounds the last two sums above uniformly in
\(v\) by \(C_0(j+1)\beta_T\), and \(\mathcal E_{\mathrm{play}}\) bounds the
sampling sum by \(C_0\{\sqrt{\Lambda(1+\Qcurv_T)}+\Lambda\}\).

\emph{Step~3: stability and terminal cost.}
Since \(\beta_T<2B_j\) and \(1/\ell_0<1\), we have
\(\beta_T\log(\beta_T^2/\kappa_0+1/\ell_0)\le\beta_T\log(4^{j+1}+1)\le3(j+1)\beta_T\).
Applying Lemma~\ref{lem:summed-stability} and combining Steps~1--3 gives,
uniformly for all unit \(v\) (and hence for all \(v\in\ball\) by affinity,
including \(v=u^\star\)),
\begin{equation}
 \sum_{t=1}^T\iprod{a_t-v}{\theta_t}\le C_0\bigl\{
 (j+1)\beta_T
 +a_j^{-1}\sqrt{d(1+\Qcurv_T)}
 +\sqrt{\Lambda(1+\Qcurv_T)}
 +\Lambda\bigr\}.
 \label{eq:regret-at-scale}
\end{equation}

\emph{Step~4: final scale.}
Substituting Lemma~\ref{lem:final-scale} for \((j+1)\beta_T\) in
\eqref{eq:regret-at-scale}, using \((j+1)a_j=\sqrt{(j+1)\ell_j/48}\) and
\(a_j^{-1}=\sqrt{48(j+1)/\ell_j}\le\sqrt{48(j+1)\ell_j}\) (as \(\ell_j>1\)),
yields \eqref{eq:growth-readable-finite}.

\emph{Prefixes.}
Fix \(\tau\le T\). All five events in \(\mathcal E\) hold at every prefix
of the horizon-\(T\) run, and Lemmas~\ref{lem:trigger}--\ref{lem:trigger-credit},
\ref{lem:contracted-ftrl}, \ref{lem:terminal-cost},
\ref{lem:bandit-correction}, and~\ref{lem:summed-stability} apply identically
at horizon \(\tau\) (with terminal covers \(H_{j,\tau}\)). In
Proposition~\ref{thm:growth-Phi-cap}, the suffix of any active round
\(t\le\tau\) lies in \(\{2,\ldots,\tau-1\}\) (so \(|I|\le\tau\)), yielding
Lemma~\ref{lem:final-scale} with \((\tau,\Qcurv_\tau,\mathsf M_\tau,j(\beta_\tau))\)
while keeping \(A,\Lambda\) (and \(T\) inside \(U_j\)) fixed. Steps~1--4
therefore apply verbatim at \(\tau\).
\end{proof}

\subsection{Anytime extension}
\label{sec:ball-headline-anytime}
\label{sec:headline-details}

To make the bound simultaneous over all horizons on the same \(\delta\)-tuned
run, apply Proposition~\ref{prop:fixed-horizon} on dyadic horizons \(T_k=2^k\)
(\(k\ge1\)) and cover every \(T\ge1\) as a prefix of \(T_{k_T}\), where
\(k_T=1\vee\ceil{\log_2T}\) satisfies \(T\le2^{k_T}\le2T\).

\begin{proofof}{Proposition~\ref{prop:main-reduction} and Theorem~\ref{thm:headline}.}
For each \(k\ge1\), set \(\bar\delta_k:=\delta/(20k^2)\) and let
\(\mathcal E_k\) intersect \(\mathcal E_{\mathrm{play}}\) and
\(\mathcal E_{\mathrm{rad}}\) at horizon \(T_k\) and level
\(\bar\delta_k\), so \(\bbP(\mathcal E_k^c)\le5\bar\delta_k=\delta/(4k^2)\).
Define
\(\mathcal E_{\mathrm{base}}
 :=\mathcal E_{\mathrm{up}}\cap\mathcal E_{\mathrm{rev}}
   \cap\mathcal E_{\mathrm{vec}}\cap\bigcap_{k\ge1}\mathcal E_k\).
Since the first three events are horizon-free at level \(\delta_0=\delta/8\)
each, a union bound gives
\(\bbP(\mathcal E_{\mathrm{base}}^c)\le\tfrac38\delta+\sum_{k\ge1}\delta/(4k^2)
 =\delta(\tfrac38+\tfrac{\pi^2}{24})<\delta\).

Fix \(T\ge1\), put \(k=k_T\) and \(j=j(\beta_T)\), and set
\[
 A_k:=\log\frac{20T_kk^2}{\delta},
 \qquad
 \Lambda_k:=\log\frac{20k^2(1+k)}{\delta},
 \qquad
 U_{k,j}:=\Bigl(1+\frac{A_k}{\ell_0}+\sqrt{\frac{A_k}{T_k}}+\frac{j+1}{\ell_j}\Bigr)^{3/4}.
\]
Because \(\bar\delta_k\le\delta/20<\delta/8\), the prefix clause of
Proposition~\ref{prop:fixed-horizon} at horizon \(T_k\) and prefix \(\tau=T\)
applies on \(\mathcal E_{\mathrm{base}}\). Its intermediate bound
\eqref{eq:regret-at-scale} (valid for all \(v\in\ball\)) proves
Proposition~\ref{prop:main-reduction}, since \(j+1,\Lambda_k=\widetilde O(1)\).
In \eqref{eq:growth-readable-finite}, applying
\eqref{eq:growth-net-drawup-relaxation} (with
\(q_j^{1/3}\sqrt{\kappa_0}\le\sqrt{\kappa_j}\)) and weighted AM--GM
(\(\kappa_j,A_k\ge1\)),
\begin{align*}
 q_j^{1/3}\kappa_0^{1/4}\sqrt{\mathsf M_T}
 &\le\sqrt{\kappa_j(1+\Qcurv_T)}+q_j^{1/3}\kappa_0^{1/4}\sqrt{\mathsf E_T},\\
 \kappa_j^{2/7}\bigl\{A_k(1+\Qcurv_T)\bigr\}^{3/14}
 &\le\sqrt{\kappa_j}\,A_k^{3/4}+\sqrt{\kappa_j(1+\Qcurv_T)},
\end{align*}
and absorbing \(\sqrt{(j+1)\ell_jd}\le(j+1)\sqrt{\kappa_j}\)
(by \eqref{eq:kappa-bounds}) gives, simultaneously for all
\(T\ge1\) on \(\mathcal E_{\mathrm{base}}\),
\begin{equation}
 \begin{split}
 \Reg_T\le C_0\Bigl\{
 &(j+1)\Bigl(\sqrt{\kappa_j}\,(U_{k,j}+A_k^{3/4})
  +\sqrt{\kappa_j(1+\Qcurv_T)}
  +q_j^{1/3}\kappa_0^{1/4}\sqrt{\mathsf E_T}\Bigr)\\
 &+\sqrt{\Lambda_k(1+\Qcurv_T)}+\Lambda_k\Bigr\}.
 \end{split}
 \label{eq:headline-explicit}
\end{equation}
Since \(j+1\le C_0\min\{T,\log(eT_kA_k)\}\), \(\ell_j\), \(q_j\), \(A_k\),
and \(\Lambda_k\) are polylogarithmic in \(T\) and \(1/\delta\) and
\(\kappa_j\le C_0(j+1)\ell_j^2d\), \eqref{eq:intro-readable-rate} follows.
\end{proofof}

\begin{proofof}{Theorem~\ref{thm:worst-case}.}
Work on \(\mathcal E_{\mathrm{base}}\), and write \(G=\log(eT)\ge1\) and
\(L=\ell_0+\log(e+G)\). Since
\(\ell_0=\log\{4\pi^2/(3\delta)\}\in[\log(e/\delta),3\log(e/\delta)]\), we
have \(L\le3\{\log(e/\delta)+\log\log(e^2T)\}\), so it suffices to show
\(\Reg_T\le C_0\sqrt{dTGL}\).

If \(T\le16dGL\), then trivially \(\Reg_T\le2T\le8\sqrt{dTGL}\). Assume
henceforth that \(T>16dGL\), which implies \(T>16\ell_0\) and \(L<T\).
Set \(k=k_T\) and \(j=j(\beta_T)\). From \(k\le2G\le eT\) and
\(T\le T_k\le2T\), we have \(A_k\le6G+\ell_0\le7GL<T\), so
\(A_k/T_k\le1\), \(A_k/\ell_0\le6G/\ell_0+1\), and \(\Lambda_k\le C_0L\).
Lemma~\ref{lem:final-scale} gives \(j+1\le C_0\log(eT_kA_k)\le C_0G\),
whence \(\ell_j\le C_0L\) and \(\kappa_j\le C_0(j+1)dL^2\).

Apply \eqref{eq:growth-readable-finite} at horizon \(T_k\) and prefix
\(T\le T_k\) (using \(\Qcurv_T\le T\)). Its second line is at most
\(C_0(\sqrt{dTGL}+\sqrt{TL}+L)\le C_0\sqrt{dTGL}\). On its first line,
\(A_k<T\) gives \(A_k(1+T)\le2T^2\), so
\(\{A_k(1+T)\}^{3/14}\le2T^{3/7}\), leaving only the two terms
\((j+1)\sqrt{\kappa_j}\,U_{k,j}\) and \((j+1)\kappa_j^{2/7}T^{3/7}\).

\emph{Case~1: \(G\le\ell_0\).}
Then \(L\le2\ell_0\), \(A_k\le7\ell_0\), and
\((j+1)/\ell_j\le C_0G/\ell_0\le C_0\), so \(U_{k,j}\le C_0\),
\(\ell_j\le C_0\ell_0\), and \(\kappa_j\le C_0(j+1)d\ell_0^2\). Put
\(r=T/\ell_0>16\). Dividing Lemma~\ref{lem:final-scale} by
\(\beta_0=\sqrt{\kappa_0}\asymp\sqrt d\,\ell_0\) (using \(\Qcurv_T\le T\)
and \(A_k\le7\ell_0\)) yields \(\beta_T/\beta_0\le C_0(j+1)^{1/2}r^{1/2}\),
and hence \(j+1\le1+\log_2(\beta_T/\beta_0)\le C_0\log(er)\) (absorbing
\(\tfrac12\log_2(j+1)\) into \((j+1)/2\)). Since \(T=r\ell_0\) and
\(\log(er)\le G\),
\begin{align*}
 (j+1)\sqrt{\kappa_j}\,U_{k,j}
 &\le C_0\log^{3/2}(er)\sqrt d\,\ell_0
  =C_0\frac{\log(er)}{\sqrt r}\,\sqrt{dT\log(er)\,\ell_0}
  \le C_0\sqrt{dTGL},\\
 (j+1)\kappa_j^{2/7}T^{3/7}
 &\le C_0\log^{9/7}(er)\,d^{2/7}\ell_0\,r^{3/7}
  =C_0\frac{\log^{11/14}(er)}{r^{1/14}d^{3/14}}\,\sqrt{dT\log(er)\,\ell_0}
  \le C_0\sqrt{dTGL},
\end{align*}
because \(\log(er)/\sqrt r\) and \(\log^{11/14}(er)/r^{1/14}\) are bounded
on \([1,\infty)\).

\emph{Case~2: \(G>\ell_0\).}
Then \(L\le2G\), \(A_k\le7G\), \(U_{k,j}\le C_0G^{3/4}\), and
\(\kappa_j\le C_0dG^3\), so
\((j+1)\sqrt{\kappa_j}\,U_{k,j}\le C_0G^{13/4}\sqrt d\) and
\((j+1)\kappa_j^{2/7}T^{3/7}\le C_0G^{13/7}d^{2/7}T^{3/7}\).
Since \(G=\log(eT)\), both \(G^{13/4}/T^{1/2}\) and \(G^{13/7}/T^{1/14}\)
are bounded for all \(T\ge1\), giving at most
\(C_0\sqrt{dT}\le C_0\sqrt{dTGL}\).
\end{proofof}

\section{Data-dependent consequences and separations}
\label{sec:consequences}

\subsection{Relaxations and simultaneous anytime events}
\label{sec:ball-headline-relaxations}

We relax the excess radial drift in Theorem~\ref{thm:headline} and bound it by
the number of rounds below the quadratic-form threshold.

Let
\begin{equation}
 T_{\mathrm{low}}=\bigl|\{2\le t\le T-1:\ \normt{G_t}{\theta_t}^2<\kappa_0^{-1}\}\bigr|
 \label{eq:low-energy-count}
\end{equation}
count the rounds whose loss quadratic form is below the
threshold \(\kappa_0^{-1}\).

\Needspace{10\baselineskip}
\begin{corollary}[Relaxations of the excess radial drift]
\label{cor:headline-relaxations}
Pathwise:
\begin{lemenum}
\item \(\mathsf E_T\le\sqrt{[T-2]_+\Qcurv_{T-1}}\).
\item \(\mathsf E_T\le T_{\mathrm{low}}/(4\sqrt{\kappa_0})\) and
 \(\sqrt d\,\mathsf E_T\le T_{\mathrm{low}}/8\); hence, on the event of
 Theorem~\ref{thm:headline},
 \begin{equation}
  \Reg_T=\widetilde O\bigl(\sqrt{d\Qcurv_T+T_{\mathrm{low}}}+\sqrt d\bigr).
  \label{eq:low-energy-relaxation}
 \end{equation}
\item \(\mathsf E_T\le\bigl[T/(4\sqrt{\kappa_0})
 -\sqrt{\kappa_0}\sum_{t=1}^T\twonorm{P_{\perp,t}\theta_t}^2\bigr]_+\);
 in particular, \(\mathsf E_T=0\) whenever
 \(\sum_{t=1}^T\twonorm{P_{\perp,t}\theta_t}^2\ge T/(4\kappa_0)\).
\end{lemenum}
\end{corollary}

\begin{proof}
\emph{Drift relaxation.}
By definition and \eqref{eq:growth-net-drawup-relaxation},
\[
 0\le\mathsf E_T\le\mathsf M_T
 \le\sqrt{[T-2]_+\sum_{t=2}^{T-1}\normt{G_t}{\theta_t}^2}
 \le\sqrt{[T-2]_+\Qcurv_{T-1}}.
\]

\emph{Counting low-quadratic-form rounds.}
For the count bounds leading to \eqref{eq:low-energy-relaxation}, if \(\mathsf M_T=0\)
there is nothing to prove; otherwise let \(I^\star\) attain the inner
maximum in \eqref{eq:intro-net-drawup}.  Since
\(\Qcurv_T\ge\sum_{t\in I^\star}\normt{G_t}{\theta_t}^2\) and
\(\varpi_t\iprod{e_t}{\widehat\Theta_{t-1}-\E_t\widehat\Theta_t}\le \normt{G_t}{\theta_t}\) by
\eqref{eq:growth-Phi-depth},
\[
 \mathsf E_T
 \le\Bigl[\sum_{t\in I^\star}\bigl(\varpi_t\iprod{e_t}{\widehat\Theta_{t-1}-\E_t\widehat\Theta_t}-\sqrt{\kappa_0}\normt{G_t}{\theta_t}^2\bigr)\Bigr]_+
 \le\sum_{t\in I^\star}\bigl[\normt{G_t}{\theta_t}-\sqrt{\kappa_0}\normt{G_t}{\theta_t}^2\bigr]_+.
\]
The summand is positive only when \(\normt{G_t}{\theta_t}<\kappa_0^{-1/2}\), and the
concave map \(x\mapsto x-\sqrt{\kappa_0}x^2\) is maximised at
\(x=1/(2\sqrt{\kappa_0})\) with value \(1/(4\sqrt{\kappa_0})\); this
proves the first bound.  The second
follows from \(d\ell_0^2\le\kappa_0\) (by \eqref{eq:kappa-bounds}) and
\(\ell_0>2\), which give
\(\sqrt{d/\kappa_0}\le1/\ell_0<1/2\).  Substituting the second
bound in \eqref{eq:intro-readable-rate} gives the third.

\emph{Tangential condition.}
For part~(c), we may assume \(\mathsf M_T>0\) and
use the maximising interval \(I^\star\) above. Put \(y_t=\sqrt{p_t}|\iprod{e_t}{\theta_t}|\).
Lemma~\ref{lem:growth-Phi-geometry} gives \(\varpi_t\le\sqrt{p_t}\), so
\eqref{eq:intro-net-drawup}, \(\epsilon_t\twonorm{\widehat\Theta_{t-1}}\ge0\), and
\(0\le\lambda_tp_t\le1\) imply
\(\varpi_t\iprod{e_t}{\widehat\Theta_{t-1}-\E_t\widehat\Theta_t}\le\varpi_t|\iprod{e_t}{\theta_t}|\le y_t\).
Using \(\normt{G_t}{\theta_t}^2=\twonorm{P_{\perp,t}\theta_t}^2+y_t^2\) and completing the square,
\begin{align*}
 \varpi_t\iprod{e_t}{\widehat\Theta_{t-1}-\E_t\widehat\Theta_t}-\sqrt{\kappa_0}\normt{G_t}{\theta_t}^2
 &\le y_t-\sqrt{\kappa_0}y_t^2-\sqrt{\kappa_0}\twonorm{P_{\perp,t}\theta_t}^2\\
 &=\frac1{4\sqrt{\kappa_0}}
   -\sqrt{\kappa_0}\Bigl(y_t-\frac1{2\sqrt{\kappa_0}}\Bigr)^2
   -\sqrt{\kappa_0}\twonorm{P_{\perp,t}\theta_t}^2\\
 &\le\frac1{4\sqrt{\kappa_0}}-\sqrt{\kappa_0}\twonorm{P_{\perp,t}\theta_t}^2.
\end{align*}
Apply this bound on \(I^\star\) and keep the rounds outside \(I^\star\):
\begin{align*}
 \mathsf M_T-\sqrt{\kappa_0}(1+\Qcurv_T)
 &\le\sum_{t\in I^\star}\bigl(\varpi_t\iprod{e_t}{\widehat\Theta_{t-1}-\E_t\widehat\Theta_t}-\sqrt{\kappa_0}\normt{G_t}{\theta_t}^2\bigr)
 -\sqrt{\kappa_0}\sum_{t\notin I^\star}\normt{G_t}{\theta_t}^2\\
 &\le|I^\star|/(4\sqrt{\kappa_0})-\sqrt{\kappa_0}\sum_{t=1}^T\twonorm{P_{\perp,t}\theta_t}^2,
\end{align*}
using \(\normt{G_t}{\theta_t}^2\ge\twonorm{P_{\perp,t}\theta_t}^2\) off \(I^\star\).
Since \(|I^\star|\le T\), taking positive parts proves part~(c).
\end{proof}

\paragraph{Simultaneous anytime events.}
\phantomsection
\label{sec:anytime-consequences}

Retain \(T_k=2^k\),
\(k_T=1\vee\ceil{\log_2T}\) (so \(T\le T_{k_T}\le2T\)),
\(\bar\delta_k=\delta/(20k^2)\), and the base event
\(\mathcal E_{\mathrm{base}}=\mathcal E_{\mathrm{up}}\cap\mathcal E_{\mathrm{rev}}\cap\mathcal E_{\mathrm{vec}}\cap\bigcap_{k\ge1}\mathcal E_k\)
of Appendix~\ref{sec:ball-headline-anytime}, which fails with probability
at most \(\delta(3/8+\pi^2/24)\). On \(\mathcal E_{\mathrm{base}}\),
both \eqref{eq:growth-readable-finite} and \eqref{eq:headline-explicit}
hold simultaneously for all \(T\ge1\) with \(k=k_T\), \(j=j(\beta_T)\),
and \(A_k,\Lambda_k,U_{k,j}\) as in
Appendix~\ref{sec:ball-headline-anytime}. Note that
\(\Lambda_k=\log\{20k^2(1+k)/\delta\}\ge\log40>3.68\).
We extend \(\mathcal E_{\mathrm{base}}\) in two ways, both using the
original \(\delta\)-tuned run and total failure probability at most
\(\delta(3/8+5\pi^2/96)<\delta\):
\begin{itemize}[leftmargin=*,itemsep=1pt,topsep=2pt]
\item \emph{Two-sided sampling extension \(\mathcal E_{\mathrm{canc}}\):}
for each \(k\ge1\), intersect \(\mathcal E_{\mathrm{base}}\) with
Lemma~\ref{lem:played} for \(\pm\iprod{a_t-z_t}{\theta_t}\) on prefixes
\(\tau\le T_k\), each side at level \(\delta/(32k^2)\) (extra failure
probability \(\sum_{k\ge1}2\delta/(32k^2)=\delta\pi^2/96\)). Since
\(\log\{32k^2(1+k)/\delta\}=\Lambda_k+\log(8/5)<\tfrac76\Lambda_k\), this
yields
\(\bigl|\sum_{t=1}^T\iprod{a_t-z_t}{\theta_t}\bigr|\le\tfrac52\sqrt{(\Qcurv_T+1)\Lambda_k}+\Lambda_k\)
for all \(T\le T_k\).
\item \emph{Random-loss extension \(\mathcal E_{\mathrm{rand}}\):}
for each \(k\ge1\), intersect \(\mathcal E_{\mathrm{base}}\) with the five
martingale events of Step~2 in Appendix~\ref{sec:random-losses} on
prefixes \(\tau\le T_k\), each at level \(\delta/(160k^2)\) (extra failure
probability \(\sum_{k\ge1}5\delta/(160k^2)=\delta\pi^2/192\)). Their
peeled logarithm satisfies
\(\log(160k^2/\delta)+\log\{\pi^2(1+k)^2/3\}<3\Lambda_k\).
\end{itemize}

\begin{lemma}[Absorbing square and fourth roots]
\label{lem:root-absorption}
For \(x,a,b,c\ge0\):
\begin{lemenum}
\item if \(x\le a+b\sqrt x\), then \(x\le2a+b^2\);
\item if \(x\le a+b\sqrt x+cx^{1/4}\), then
 \(x\le2a+2b^2+\tfrac32c^{4/3}\).
\end{lemenum}
\end{lemma}

\begin{proof}
Part~(a) follows from \(b\sqrt x\le x/2+b^2/2\). For~(b), Young's
inequality gives \(b\sqrt x\le x/4+b^2\) and
\(cx^{1/4}\le x/4+\tfrac34c^{4/3}\); subtracting \(x/2\) and doubling
yields the claim.
\end{proof}

\subsection{Cancellation}
\label{sec:cancellation}

Recall the cancellation slack
\(\Delta_T=\sum_{t=1}^T\twonorm{\theta_t}-\twonorm{S_T}\ge0\); it measures
the gain from choosing a separate best action on every round and, when
\(S_T\ne0\), the weighted angular scatter of the losses about the
cumulative loss direction. Changing magnitudes along one ray gives zero
slack; when losses sum to zero, the slack is their total norm.
Set \(\ell_t^+(u)=\iprod u{\theta_t}+\twonorm{\theta_t}\in[0,2]\).
Since \(\twonorm{u^\star}=1\) and \(\iprod{u^\star}{S_T}=-\twonorm{S_T}\),
\[
 \Delta_T
 =\sum_t\ell_t^+(u^\star)=\sum_t\twonorm{\theta_t}-\twonorm{S_T}
 =\frac12\sum_{t:\,\theta_t\ne0}\twonorm{\theta_t}
 \twonorm*{\frac{\theta_t}{\twonorm{\theta_t}}+u^\star}^2\ge0.
\]

\begin{lemma}[Quadratic form via regret and cancellation]
\label{lem:clock-regret-cancellation}
Pathwise,
\begin{equation}
 \sum_{t=1}^T\bigl(\twonorm{\theta_t}^2-\twonorm{z_t}\iprod{e_t}{\theta_t}^2\bigr)
 \le2\Reg_T+2\Delta_T-2\sum_{t=1}^T\iprod{a_t-z_t}{\theta_t}.
 \label{eq:clock-regret-cancellation}
\end{equation}
\end{lemma}

\begin{proof}
The one-round identity
\(2\twonorm{\theta_t}\ell_t^+(z_t)-\normt{G_t}{\theta_t}^2
 =p_t\twonorm{\theta_t}^2+(1-p_t)\ell_t^+(e_t)^2\ge0\)
gives \(\normt{G_t}{\theta_t}^2\le2\twonorm{\theta_t}\ell_t^+(z_t)\le2\ell_t^+(z_t)\)
(since \(\twonorm{\theta_t}\le1\) and \(\ell_t^+(z_t)\ge0\)).
Summing and inserting the definitions of regret and cancellation gives
\begin{align*}
 \Qcurv_T
 &\le2\sum_t\ell_t^+(z_t)
  =2\sum_t\{\iprod{a_t}{\theta_t}+\twonorm{\theta_t}\}
        -2\sum_t\iprod{a_t-z_t}{\theta_t}\\
 &=2\Reg_T+2\Delta_T-2\sum_t\iprod{a_t-z_t}{\theta_t}. \qedhere
\end{align*}
\end{proof}

Substituting Lemma~\ref{lem:clock-regret-cancellation} into
Theorem~\ref{thm:headline} and solving for the regret via a two-sided
sampling-noise bound gives a bound in the cancellation slack. With full
information, regret \(O(1+\sqrt{\Delta_T})\) is minimax on the Euclidean
ball and attainable without knowing \(T\) or \(\Delta_T\)
\citep{mayo2026smallloss}, so regret on aligned losses is constant.

\begin{corollary}[Cancellation bound]
\label{cor:cancellation}
With probability at least \(1-\delta\), simultaneously for all \(T\ge1\),
\begin{equation}
 \Reg_T=\widetilde O\!\left(
 d+\sqrt{d\,\Delta_T}+d^{1/4}\sqrt{\mathsf E_T}\right)
 =\widetilde O\!\left(d+\sqrt{d\,\Delta_T}+\sqrt{T_{\mathrm{low}}}\right).
 \label{eq:cancellation-form}
\end{equation}
\end{corollary}

\begin{proof}
Work on the simultaneous event \(\mathcal E_{\mathrm{canc}}\) of
Appendix~\ref{sec:anytime-consequences}, which has probability at least
\(1-\delta\). Fix any \(T\ge1\) with \(\Reg_T>0\) (the bound is immediate
otherwise) and put \(k=k_T\), so \(T\le T_k\le2T\).

\emph{Step 1: self-bound the loss quadratic form.}
Inserting the two-sided sampling bound on \(\mathcal E_{\mathrm{canc}}\)
into \eqref{eq:clock-regret-cancellation} gives
\(\Qcurv_T+1\le2\Reg_T+2\Delta_T+2\Lambda_k+1+5\sqrt{(\Qcurv_T+1)\Lambda_k}\),
and Lemma~\ref{lem:root-absorption}(a) yields
\begin{equation}
 \Qcurv_T+1\le4\Reg_T+4\Delta_T+29\Lambda_k+2.
 \label{eq:clock-self-bound}
\end{equation}

\emph{Step 2: absorb the regret term.}
Write \(j=j(\beta_T)\),
\(\alpha_k=(j+1)\sqrt{\kappa_j}+\sqrt{\Lambda_k}\), and
\(a=4\Delta_T+29\Lambda_k+2\). By \eqref{eq:clock-self-bound},
\(\sqrt{\Qcurv_T+1}\le2\sqrt{\Reg_T}+\sqrt a\). Substituting into
\eqref{eq:headline-explicit} gives \(\Reg_T\le B+2C_0\alpha_k\sqrt{\Reg_T}\), where
\[
 B=C_0\Bigl\{
 (j+1)\sqrt{\kappa_j}\,(U_{k,j}+A_k^{3/4})+\Lambda_k
 +\alpha_k\sqrt a+(j+1)q_j^{1/3}\kappa_0^{1/4}\sqrt{\mathsf E_T}\Bigr\},
\]
so Lemma~\ref{lem:root-absorption}(a) yields \(\Reg_T\le2B+4C_0^2\alpha_k^2\).
Since \(\kappa_j\le C_0(j+1)\ell_j^2d\),
\(j+1\le C_0\log(eT_kA_k)\), and
\(A_k,\Lambda_k,\ell_j,q_j,U_{k,j}\)
are polylogarithmic in \(T\) and \(1/\delta\), this gives the first form
in \eqref{eq:cancellation-form}; the second follows from
\(\kappa_0^{1/4}\sqrt{\mathsf E_T}\le\frac12\sqrt{T_{\mathrm{low}}}\)
by Corollary~\ref{cor:headline-relaxations}.
\end{proof}

\paragraph{Removing the excess radial drift.}
Using \(\mathsf M_T\le\sqrt{T\Qcurv_T}\) in the same self-bound and
absorbing the fourth root of regret via Young's inequality gives a
learner-independent bound.

\begin{corollary}[Learner-independent cancellation bound]
\label{cor:first-order}
With probability at least \(1-\delta\), simultaneously for all \mbox{\(T\ge1\)},
\begin{equation}
 \Reg_T=\widetilde O\!\left(\sqrt{d\,\Delta_T}+d^{1/4}(T\Delta_T)^{1/4}
 +(dT)^{1/3}+d\right).
 \label{eq:first-order-form}
\end{equation}
\end{corollary}

\begin{proofof}{Corollary~\ref{cor:first-order}.}
Work on \(\mathcal E_{\mathrm{canc}}\) and fix \(T\ge1\) with
\(\Reg_T>0\); put \(k=k_T\), \(j=j(\beta_T)\), and
\(a=4\Delta_T+29\Lambda_k+2\). Note that
\((j+1)\kappa_j^{2/7}\{A_k(1+\Qcurv_T)\}^{3/14}\) has exponent \(3/14<1/4\);
using \((1+\Qcurv_T)^{3/14}\le1+(1+\Qcurv_T)^{1/4}\) and
\(\mathsf M_T\le\sqrt{T\Qcurv_T}\) from
\eqref{eq:growth-net-drawup-relaxation}, Proposition~\ref{prop:fixed-horizon}
at horizon \(T_k\) and prefix \(T\) gives
\[
 \Reg_T\le C_0\Bigl\{
 (j+1)\sqrt{\kappa_j}\,U_{k,j}+(j+1)\kappa_j^{2/7}A_k^{3/14}+\Lambda_k
 +\alpha_1\sqrt{\Qcurv_T+1}
 +\alpha_2(\Qcurv_T+1)^{1/4}\Bigr\},
\]
where
\(\alpha_1=\sqrt{(j+1)\ell_jd}+\sqrt{\Lambda_k}\)
and
\(\alpha_2=(j+1)q_j^{1/3}(\kappa_0T)^{1/4}+(j+1)\kappa_j^{2/7}A_k^{3/14}\).
By \eqref{eq:clock-self-bound},
\(\sqrt{\Qcurv_T+1}\le2\sqrt{\Reg_T}+\sqrt a\) and
\((\Qcurv_T+1)^{1/4}\le\sqrt2\,\Reg_T^{1/4}+a^{1/4}\). Applying
Lemma~\ref{lem:root-absorption}(b) yields
\begin{equation}
 \begin{split}
 \Reg_T\le{}&2C_0\bigl\{
 (j+1)\sqrt{\kappa_j}\,U_{k,j}+(j+1)\kappa_j^{2/7}A_k^{3/14}+\Lambda_k
 +\alpha_1\sqrt a+\alpha_2a^{1/4}\bigr\}\\
 &+8C_0^2\alpha_1^2
 +\tfrac32\,2^{2/3}C_0^{4/3}\alpha_2^{4/3}.
 \end{split}
 \label{eq:first-order-explicit}
\end{equation}
Since \(\kappa_j\le C_0(j+1)\ell_j^2d\),
\(j+1\le C_0\log(eT_kA_k)\), all logarithmic factors are polylogarithmic
in \(T\) and \(1/\delta\), \(\sqrt d\le d\), and
\((dT)^{1/4}\le(dT)^{1/3}\), this proves \eqref{eq:first-order-form}.
\end{proofof}

\subsection{Limitations on aligned losses}
\label{sec:fixed-ray}

Throughout this subsection, \(\mathcal L_T\) is as in
Section~\ref{sec:aligned-losses}.

\begin{corollary}[Second-order and count bounds on aligned losses]
\label{cor:fixed-ray-count}
On the simultaneous event of Proposition~\ref{prop:fixed-ray-separation},
for every \(T\ge1\),
\begin{equation}
 \sqrt{d(1+\Qcurv_T)+\sqrt d\,\mathsf E_T}
 \le C_0\bigl\{\mathcal L_T^{3/2}d+\mathcal L_T(dT)^{1/3}\bigr\},
 \label{eq:ray-radial-bound}
\end{equation}
and
\begin{equation}
 T_{\mathrm{low}}
 \ge[T-2]_+-C_0d\bigl\{\mathcal L_T^5d+\mathcal L_T^{9/2}(dT)^{1/3}\bigr\}.
 \label{eq:ray-count-bound}
\end{equation}
\end{corollary}

\Needspace{4\baselineskip}
\begin{proofof}{Proposition~\ref{prop:fixed-ray-separation} and Corollary~\ref{cor:fixed-ray-count}.}
Note that \(2<\ell_0<\mathcal L_T=\log(40Tk_T^2/\delta)\) for all \(T\ge1\).

\emph{Step 1: upper-bound event and regret identity.}
Work on \(\mathcal E_{\mathrm{canc}}\)
(Appendix~\ref{sec:anytime-consequences}, failure probability at most
\(\delta(3/8+5\pi^2/96)\)). Since \(S_T=-(\sum_t\gamma_t)u\), the
hindsight comparator is \(u\), so
\(\Reg_T=\sum_{t=1}^T\gamma_t\{1-\iprod{a_t}{u}\}\ge0\) and \(\Delta_T=0\).

\emph{Step 2: explicit upper bounds on regret and \(\Qcurv_T\).}
Fix \(T\ge1\) and set \(k=k_T\), \(j=j(\beta_T)\). On prefix
\(T\le T_k\le2T\), we have \(A_k\le\mathcal L_T\) and
\(\Lambda_k,j+1,\ell_j\le C_0\mathcal L_T\), whence
\(q_j\le(j+1)(\ell_j/\ell_0)^2\), \(\kappa_j\le C_0(j+1)\ell_j^2d\),
and \(U_{k,j}\le C_0(\sqrt{\mathcal L_T}+\mathcal L_T/\ell_0)^{3/4}\).
In \eqref{eq:first-order-explicit} with \(\Delta_T=0\)
(\(a=29\Lambda_k+2\le C_0\mathcal L_T\)):
\(\alpha_1^2\le C_0\mathcal L_T^2d\);
\((j+1)\sqrt{\kappa_j}\,U_{k,j}\le C_0\mathcal L_T^3d\)
(since \(\ell_j/\ell_0^{3/4}\le C_0\mathcal L_T^{1/4}\));
\(\{(j+1)\kappa_j^{2/7}A_k^{3/14}\}^{4/3}\le C_0\mathcal L_T^{58/21}d\le C_0\mathcal L_T^3d\);
and, using subadditivity on \(\ell_j=\ell_0+2\log(j+1)\) to get
\(\ell_0^{-2/9}\ell_j^{8/9}\le\ell_0^{2/3}+\{2\log(j+1)\}^{8/9}\le C_0\mathcal L_T^{2/3}\),
\[
 \bigl\{(j+1)q_j^{1/3}(\kappa_0T)^{1/4}\bigr\}^{4/3}
 \le C_0(j+1)^{16/9}\ell_j^{8/9}\ell_0^{-2/9}(dT)^{1/3}
 \le C_0\mathcal L_T^{22/9}(dT)^{1/3}
 \le C_0\mathcal L_T^{5/2}(dT)^{1/3}.
\]
Substituting into \eqref{eq:first-order-explicit} proves the upper bound
\(\Reg_T\le C_0\{\mathcal L_T^3d+\mathcal L_T^{5/2}(dT)^{1/3}\}\), and
\eqref{eq:clock-self-bound} then gives
\begin{equation}
 \Qcurv_T+1\le4\Reg_T+29\Lambda_k+2
 \le C_0\bigl\{\mathcal L_T^3d+\mathcal L_T^{5/2}(dT)^{1/3}\bigr\}.
 \label{eq:ray-quadratic-bound}
\end{equation}

\emph{Step 3: bound the second-order expression and the selected scale.}
Write \(x_1=\mathcal L_T^{3/2}d\) and \(x_2=\mathcal L_T(dT)^{1/3}\). The pathwise
interval bound \(\mathsf E_T\le\mathsf M_T\le\sqrt{T\Qcurv_T}\) and
\eqref{eq:ray-quadratic-bound} give
\begin{align*}
 \sqrt{d(1+\Qcurv_T)+\sqrt d\,\mathsf E_T}
 &\le\sqrt{d(1+\Qcurv_T)}+(dT\Qcurv_T)^{1/4}\\
 &\le C_0\bigl\{\mathcal L_T^{3/2}d+\mathcal L_T^{5/4}d^{2/3}T^{1/6}
                  +\mathcal L_T^{3/4}d^{1/2}T^{1/4}+\mathcal L_T^{5/8}(dT)^{1/3}\bigr\}\\
 &\le C_0\bigl\{x_1+x_1^{1/2}x_2^{1/2}+x_1^{1/4}x_2^{3/4}+x_2\bigr\}
  \le C_0(x_1+x_2),
\end{align*}
where the last step uses weighted AM--GM, proving \eqref{eq:ray-radial-bound}.

To bound \(\beta_T/\sqrt{\kappa_0}\), put
\(x'_1=\mathcal L_T^{3/2}\sqrt d\) and \(x'_2=\mathcal L_T^{4/3}(dT)^{1/3}/\sqrt d\).
Dividing Lemma~\ref{lem:final-scale} at the same prefix by
\(\sqrt{\kappa_0}\asymp\ell_0\sqrt d\), using
\(\sqrt{q_j}\,U_{k,j}\le C_0\mathcal L_T^{3/2}\),
\(a_j/\ell_0<1\) (by \eqref{eq:base-scales}),
\(q_j^{1/3}\kappa_0^{-1/4}\le C_0(j+1)^{1/3}\ell_j^{2/3}\ell_0^{-7/6}d^{-1/4}\le C_0\mathcal L_T^{1/2}d^{-1/4}\),
and \(q_j^{2/7}(A_k/\kappa_0)^{3/14}\le C_0\mathcal L_T^{3/4}\),
together with \eqref{eq:ray-quadratic-bound} and weighted AM--GM, gives
\[
 \frac{\beta_T}{\sqrt{\kappa_0}}
 \le C_0(x'_1+x'_2)
 =C_0\Bigl\{\mathcal L_T^{3/2}\sqrt d+\mathcal L_T^{4/3}\frac{(dT)^{1/3}}{\sqrt d}\Bigr\}.
\]
Since \(\beta_t\) is nondecreasing and
\(\beta_t^2p_t\ge\kappa_{j_t}\ge\kappa_0\), we obtain pathwise on this event
\begin{equation}
 \sum_{t=1}^Tp_t\ge\frac{\kappa_0T}{\beta_T^2}
 \ge\frac{cdT}{\{\mathcal L_T^{3/2}d+\mathcal L_T^{4/3}(dT)^{1/3}\}^2}.
 \label{eq:ray-exploration-lower}
\end{equation}

\emph{Step 4: convert forced exploration to realised regret.}
Put \(X_t=\gamma_t\{1-\iprod{a_t}{u}\}/2\in[0,1]\); the mean-action
identity gives
\(\mu_t:=\E_tX_t=\frac{\gamma_t}{2}\{1-(1-p_t)\iprod{e_t}{u}\}\ge\gamma_0p_t/2\).
By convexity on \([0,1]\), \(e^{-x}\le1-\alpha x\) with \(\alpha=1-e^{-1}\),
so \(\E_t e^{-X_t}\le1-\alpha\mu_t\le e^{-\alpha\mu_t}\).
Consequently, \(Z_T=\exp\{\alpha\sum_{t\le T}\mu_t-\sum_{t\le T}X_t\}\)
is a nonnegative supermartingale with \(Z_0=1\). Ville's inequality gives,
outside an event of probability \(\delta/32\), simultaneously for every \(T\),
\[
 \Reg_T
 =2\sum_{t=1}^TX_t
 \ge2\alpha\sum_{t=1}^T\mu_t-2\log(32/\delta)
 \ge\alpha\gamma_0\sum_{t=1}^Tp_t-2\log(32/\delta).
\]
Combining with \eqref{eq:ray-exploration-lower}, and absorbing \(\alpha\)
into \(c\), proves the lower bound in \eqref{eq:ray-regret-bounds}, with
total failure probability at most
\(\delta(3/8+5\pi^2/96+1/32)<\delta\).

\emph{Step 5: count the low-quadratic-form rounds.}
Each eligible round not counted by \(T_{\mathrm{low}}\) contributes
at least \(\kappa_0^{-1}\) to \(\Qcurv_T\). Since
\(\kappa_0\le C_0d\ell_0^2\le C_0d\mathcal L_T^2\), \eqref{eq:ray-quadratic-bound} gives
\[
 [T-2]_+-T_{\mathrm{low}}\le\kappa_0\Qcurv_T
 \le C_0d\bigl\{\mathcal L_T^5d+\mathcal L_T^{9/2}(dT)^{1/3}\bigr\},
\]
which proves \eqref{eq:ray-count-bound}.
For fixed \(d,\delta,\gamma_0\), \(\mathcal L_T=O(\log(eT))\). The regret lower
bound then grows as \(T^{1/3}\) up to logarithms, whereas
\(d\{\mathcal L_T^5d+\mathcal L_T^{9/2}(dT)^{1/3}\}=o(T)\). Thus \(T_{\mathrm{low}}=T-o(T)\), and
\eqref{eq:ray-quadratic-bound} shows
\[
 \sqrt{d(1+\Qcurv_T)+T_{\mathrm{low}}}\asymp\sqrt T.
\]
The logarithmic prefactor and baseline in
Theorem~\ref{thm:headline} preserve the stated polynomial separation.
\end{proofof}

\subsection{Random loss vectors}
\label{sec:random-losses}

Let \(\theta_t=\bar\theta+\eta_t\in\ball\) with \(\bar\theta\) fixed,
generated by a nonanticipating adversary. Write \(\mathcal G_{t-1}\) for
the history containing the learner's state before the current noise and
fresh draw. Given this history, assume that \(\eta_t\) has mean zero and
covariance \(\Sigma_t\), and is independent of the learner's draw. Put
\(\mathcal T_T=\sum_{t=1}^T\tr\Sigma_t\).

\begin{corollary}[Random loss vectors]
\label{cor:random-losses}
Under the preceding conditional mean and covariance assumptions, with
probability at least \(1-\delta\), simultaneously for all \(T\ge1\),
\begin{equation}
 \Reg_T=\widetilde O\!\left(d+\sqrt{d\mathcal T_T}+\sqrt T\right).
 \label{eq:random-loss-bounds}
\end{equation}
\end{corollary}

The last term of \eqref{eq:random-loss-bounds} is absorbed when
\(\mathcal T_T\ge T/(4\kappa_0)\). The bound depends on the mean only
through feasibility: \(\tr\Sigma_t\le1-\twonorm{\bar\theta}^2\). The
horizon-tuned i.i.d.\ bound of \citet[Theorem~3.9]{ezer2026parameter} also
needs no trace information, because it estimates the noise; our bound uses
anytime tuning, allows adaptive adversaries, and has an additive
\(\sqrt T\) term (Appendix~\ref{sec:extended-consequences}).

\begin{proof}
Work on the probability space of Appendix~\ref{sec:notation}, where the
nonanticipating adversary determines \(\theta_t=\bar\theta+\eta_t\) from
\(\omega\) and past learner draws, so \(\mathcal E_{\mathrm{base}}\) holds
with respect to \(\mathcal F_t\). Let \(\mathcal G_{t-1}\subseteq\mathcal F_{t-1}\)
be the smaller history before both \(\eta_t\) and the learner's fresh draw
(revealing \(\eta_t\) only at round \(t\)), and write
\(\E_t^{\mathcal G}[\cdot]=\E[\cdot\mid\mathcal G_{t-1}]\).
Conditional Jensen gives \(r:=\twonorm{\bar\theta}\le1\) and
\(\twonorm{\eta_t}\le2\).

\emph{Step 1: separate mean loss and noise, and relate to regret.}
Write \(\bar\chi_t=\iprod{e_t}{\bar\theta}\). Then
\(\normt{G_t}{\theta_t}^2=x_t+\zeta_t+w_t\), where
\(x_t=\normt{G_t}{\bar\theta}^2=r^2-(1-p_t)\bar\chi_t^2\),
\(\zeta_t=2\iprod{G_t\bar\theta}{\eta_t}\), and
\(w_t=\normt{G_t}{\eta_t}^2\in[0,4]\), with
\(\E_t^{\mathcal G}w_t=\tr(G_t\Sigma_t)\le\tr\Sigma_t\).
Let \(g_t=(1-p_t)\bar\chi_t+r\) and \(\bar R_T=\sum_{t=1}^Tg_t\). Since
\(\bar\chi_t\ge-r\),
\(2rg_t-x_t=p_tr^2+(1-p_t)(r+\bar\chi_t)^2\ge0\) and
\(g_t-p_tr=(1-p_t)(r+\bar\chi_t)\ge0\), so
\(\sum_{t=1}^Tx_t\le2r\bar R_T\) and \(r\sum_{t=1}^Tp_t\le\bar R_T\).
With \(M^1_T=\sum_{t=1}^T\iprod{a_t-z_t}{\bar\theta}\) and
\(M^2_T=\sum_{t=1}^T\iprod{a_t}{\eta_t}\), the reverse triangle
inequality gives
\(\Reg_T=\bar R_T-Tr+M^1_T+M^2_T+\twonorm{T\bar\theta+\sum_{t=1}^T\eta_t}
 \ge\bar R_T+M^1_T+M^2_T-\twonorm{\sum_{t=1}^T\eta_t}\), hence
\begin{equation}
 \bar R_T
 \le\Reg_T+\twonorm*{\sum_{t=1}^T\eta_t}+|M^1_T|+|M^2_T|.
 \label{eq:mean-regret-bound}
\end{equation}

\emph{Step 2: verify the five concentration inputs.}
Given \(\mathcal G_{t-1}\), the following five increments are centred
(with (b) using conditional independence of \(\eta_t\) and \(a_t\)):
\begin{enumerate}[label=(\alph*),leftmargin=*,itemsep=1pt,topsep=2pt]
\item \(\iprod{a_t-z_t}{\bar\theta}\), bounded by \(\sqrt2\,r\)
(as in Lemma~\ref{lem:played}), with conditional variance
\(p_t(1-p_t)\bar\chi_t^2+p_t\normt{G_t}{\bar\theta}^2/d\le2p_tr^2\);
\item \(\iprod{a_t}{\eta_t}\), bounded by \(2\), with conditional second
moment \(\E_t^{\mathcal G}[a_t\transpose\Sigma_ta_t]\le\tr\Sigma_t\);
\item \(\zeta_t\), bounded by \(4r\), with conditional variance
\(4\bar\theta\transpose G_t\Sigma_tG_t\bar\theta\le4r^2\tr\Sigma_t\);
\item \(w_t-\E_t^{\mathcal G}w_t\), bounded by \(4\), with conditional
variance at most \(\E_t^{\mathcal G}w_t^2\le4\tr\Sigma_t\);
\item \(\eta_t\), with norm at most \(2\) and conditional second moment
\(\tr\Sigma_t\).
\end{enumerate}
For each \(k\ge1\), apply Proposition~\ref{prop:contractive-classical:peeled}
with \(K_t=I\) and \(\ell=\log(160k^2/\delta)\) to each sequence on
prefixes \(T\le T_k\) (omitting (a) and (c) when \(r=0\)); this defines
the event \(\mathcal E_{\mathrm{rand}}\) of
Appendix~\ref{sec:anytime-consequences}, of probability at least
\(1-\delta\).

\emph{Step 3: absorb mean regret and finish.}
Work on \(\mathcal E_{\mathrm{rand}}\), fix any \(T\ge1\) with
\(\Reg_T>0\), and put \(k=k_T\) and \(j=j(\beta_T)\).
By \eqref{eq:headline-explicit} and
\(\kappa_0^{1/4}\sqrt{\mathsf E_T}\le\tfrac12\sqrt T\)
(Corollary~\ref{cor:headline-relaxations}), writing
\(\alpha_k:=(j+1)\sqrt{\kappa_j}+\sqrt{\Lambda_k}\),
\begin{equation}
 \Reg_T\le C_0\Bigl\{
 (j+1)\sqrt{\kappa_j}\,(U_{k,j}+A_k^{3/4})+\Lambda_k
 +\alpha_k\sqrt{1+\Qcurv_T}+\tfrac12(j+1)q_j^{1/3}\sqrt T\Bigr\}.
 \label{eq:random-finite-form}
\end{equation}
Since the peeled logarithm of \(\mathcal E_{\mathrm{rand}}\) is at most
\(3\Lambda_k\), for range \(R\) and cumulative variance \(V\le T_kR^2\),
Proposition~\ref{prop:contractive-classical:peeled} bounds each sequence
by \(c\sqrt{V\Lambda_k}+cR\Lambda_k\) for an absolute constant \(c\). With
\(r^2\sum_{t=1}^Tp_t\le r\bar R_T\), on \(\mathcal E_{\mathrm{rand}}\),
\[
 |M^1_T|\le c\sqrt{r\bar R_T\Lambda_k}+cr\Lambda_k,
 \qquad
 \Bigl|\sum_{t=1}^T\zeta_t\Bigr|\le cr\sqrt{\mathcal T_T\Lambda_k}+cr\Lambda_k,
\]
and \(|M^2_T|\), \(\twonorm{\sum_{t=1}^T\eta_t}\), and
\(\bigl|\sum_{t=1}^T(w_t-\E_t^{\mathcal G}w_t)\bigr|\) are each at most
\(c\sqrt{\mathcal T_T\Lambda_k}+c\Lambda_k\).
By \eqref{eq:mean-regret-bound}, \(r\le1\), and
Lemma~\ref{lem:root-absorption}(a),
\(\bar R_T\le2\Reg_T+4c\sqrt{\mathcal T_T\Lambda_k}+(6c+c^2)\Lambda_k\).
Summing \(\normt{G_t}{\theta_t}^2=x_t+\zeta_t+w_t\), substituting the
bound on \(\bar R_T\), and applying AM--GM to \(\sqrt{\mathcal T_T\Lambda_k}\) yields
\[
 \Qcurv_T
 \le2r\bar R_T+\mathcal T_T+2c\sqrt{\mathcal T_T\Lambda_k}+2c\Lambda_k
 \le4r\Reg_T+2\mathcal T_T+C_0\Lambda_k.
\]
Substituting into \eqref{eq:random-finite-form}, splitting the square
root, and applying Lemma~\ref{lem:root-absorption}(a) with coefficient
\(2C_0\alpha_k\sqrt r\) gives
\[
 \Reg_T\le C_0\bigl\{(j+1)\sqrt{\kappa_j}\,(U_{k,j}+A_k^{3/4})+\Lambda_k
 +\alpha_k\sqrt{\mathcal T_T+\Lambda_k+1}
 +\alpha_k^2r+(j+1)q_j^{1/3}\sqrt T\bigr\}.
\]
Since \(\kappa_j\le C_0(j+1)\ell_j^2d\), \(r\le1\), and all
logarithmic factors are polylogarithmic in \(T\) and \(1/\delta\), this
proves \eqref{eq:random-loss-bounds}. When
\(\mathcal T_T\ge T/(4\kappa_0)\),
\(\sqrt T\le2\sqrt{\kappa_0\mathcal T_T}=\widetilde O(\sqrt{d\mathcal T_T})\).
\end{proof}

\paragraph{The isotropic example.}
Let \(d\ge4\), \(\theta_t=\bar\theta+d^{-1/2}W_t\), and
\(\twonorm{\bar\theta}=1/2\), with independent uniform unit vectors
\(W_t\), independent of the learner. Then
\(\theta_t\in\ball\), \(\Sigma_t=d^{-2}I_d\), and \(\mathcal T_T=T/d\),
so Corollary~\ref{cor:random-losses} gives
\(\Reg_T=\widetilde O(d+\sqrt T)\). This improves on the worst-case rate
once \(T\gg d\), and by the full factor \(\sqrt d\) once \(T\ge d^2\).

\section{Proof of the second-order lower bound}
\label{sec:lower-bound-proof}

We prove Theorem~\ref{thm:lower-bound} in a slightly more general form.
Fix a density \(\nu\) supported on \([1/4,1/2]\) with
\(\sqrt\nu\in H^1(\R)\) and a number \(\gamma\in(0,1/8]\); the proof
gives \eqref{eq:lower-conclusion} with a constant
\(c=c_{\nu,\gamma}>0\) depending only on \((\nu,\gamma)\), using
\eqref{eq:lower-assumption} at any \(\delta_{\mathrm{lb}}\in(0,1/(16T)]\).
An explicit choice such as \(\gamma=1/8\) and
\(\nu(r)=8\sin^2(4\pi(r-1/4))\) on \([1/4,1/2]\) yields the absolute
constant of Theorem~\ref{thm:lower-bound}.

Fix orthonormal vectors \(e,f\in\R^d\) (\(d\ge2\)), put
\(J_\nu=\int_{\R}|(\sqrt\nu)'(r)|^2\,dr\) and
\(C_{\nu,\gamma}=4\vee6J_\nu\gamma^2\), and write \(\bbP_{\mathrm{alg}}\)
for probability over the learner's randomisation on a fixed loss sequence.
We use the unnormalised Hellinger convention
\(H^2(P,Q)=\int(\sqrt{dP}-\sqrt{dQ})^2\), so \(H^2\le2\) and
\(\operatorname{TV}(P,Q)\le H(P,Q)\) \citep[Section~14.3]{lattimore2020bandit}.
Draw \(R_1,\ldots,R_T\stackrel{\mathrm{i.i.d.}}{\sim}\nu\) independently of
the learner. For an interval \(I\subseteq\{1,\ldots,T\}\) and sign
\(v\in\{-1,1\}\), we compare the null losses \(\theta_t^0=-R_te\) with the
perturbed losses \(\theta_t^{I,v}=-R_te+\gamma vf\II{t\in I}\).

\emph{Transcript and joint laws.}
Because the learner observes scalar feedback but not the radii, we measure
information in the transcript after marginalising out \(R_{1:T}\), while
retaining \(R_{1:T}\) in the joint law to evaluate regret.
With \(\Xi_t\) the learner's private state on round \(t\), collect
\(U_t=(\Xi_t,e_t,p_t,a_t)\) and \(Y_t=\iprod{a_t}{\theta_t}\) into the
augmented transcript \(H_T=(U_1,Y_1,\ldots,U_T,Y_T)\), regarding algorithmic
objects as measurable kernels. Let \(\bbP_0,\bbP_{I,v}\) denote the joint
laws of \((R_{1:T},H_T)\) and \(P_0,P_{I,v}\) the marginal laws of \(H_T\);
write \(\E_0\) for expectation under \(\bbP_0\) (equivalently \(P_0\) on
transcript-measurable variables).

\subsection{Information in the marginal feedback laws}

The learner's departure from the null sequence's best action \(e\)
bounds statistical distinguishability. The one-round bound concerns
feedback alone; the sequential bound concerns the augmented transcript
and allows the learner's actions to adapt.

\begin{lemma}[Marginalised Hellinger comparison]
\label{lem:lower-radial-nuisance}
Let \(R\sim\nu\), let
\(a=\alpha e+\zeta f+a_\perp\in\ball\), where
\(a_\perp\perp\operatorname{span}\{e,f\}\), and fix
\(v\in\{-1,1\}\). For
\(Y_0=-\alpha R\) and \(Y_v=-\alpha R+\gamma v\zeta\),
\begin{equation}
 H^2(\mathcal L(Y_v),\mathcal L(Y_0))
 \le C_{\nu,\gamma}(1-\alpha).
 \label{eq:lower-one-step-hellinger}
\end{equation}
For the full marginal transcript, every \(\eta>0\) satisfies, with
\(\mu:=C_{\nu,\gamma}\E_0\sum_{t\in I}(1-\iprod{a_t}{e})\),
\begin{equation}
 \operatorname{TV}(P_{I,v},P_0)
 \le\frac{\mu}{\eta}+\sqrt{2(\mu+\eta)}.
 \label{eq:lower-hellinger}
\end{equation}
\end{lemma}

\begin{proof}
\emph{Step 1: compare the one-round feedback laws.}
If \(\alpha\le1/2\), then
\[
 H^2(\mathcal L(Y_v),\mathcal L(Y_0))\le2\le4(1-\alpha)
 \le C_{\nu,\gamma}(1-\alpha),
\]
since \(H^2=2-2\int\sqrt{dP\,dQ}\le2\).

For \(\alpha>1/2\), hence \(\alpha>0\), put
\(g=\sqrt\nu\), \(b=\gamma v\zeta\), \(h=b/\alpha\).
The Lebesgue densities of \(Y_0\) and \(Y_v\) are
\(p_0(y)=\alpha^{-1}\nu(-y/\alpha)\) and
\(p_v(y)=\alpha^{-1}\nu((b-y)/\alpha)\).
Substituting \(r=-y/\alpha\) and \(|dy|=\alpha\,dr\) gives
\begin{equation}
 H^2(\mathcal L(Y_v),\mathcal L(Y_0))
 =\int_{\R}\{g(r+h)-g(r)\}^2\,dr.
 \label{eq:lower-scaled-translation}
\end{equation}
For the locally absolutely continuous representative of \(g\in H^1(\R)\),
\(g(r+h)-g(r)=h\int_0^1g'(r+sh)\,ds\); Cauchy--Schwarz on \([0,1]\),
Tonelli, and translation invariance bound \eqref{eq:lower-scaled-translation}
by \(h^2J_\nu\). With \(v^2=1\), this gives
\(H^2(\mathcal L(Y_v),\mathcal L(Y_0))\le\gamma^2\zeta^2J_\nu/\alpha^2\).
Orthogonality and \(a\in\ball\) give
\(\alpha^2+\zeta^2+\twonorm{a_\perp}^2\le1\); since
\(\alpha\mapsto(1+\alpha)/\alpha^2\) decreases on \((0,\infty)\) and equals
\(6\) at \(\alpha=1/2\),
\[
 \frac{\zeta^2}{\alpha^2}
 \le\frac{1-\alpha^2}{\alpha^2}
 =\frac{1+\alpha}{\alpha^2}(1-\alpha)
 \le6(1-\alpha).
\]
Therefore
\(H^2(\mathcal L(Y_v),\mathcal L(Y_0))\le6J_\nu\gamma^2(1-\alpha)\le C_{\nu,\gamma}(1-\alpha)\),
proving \eqref{eq:lower-one-step-hellinger}.

\emph{Step 2: factor the conditional interaction laws.}
Work with the transcript laws after marginalising the radii (retaining
\(R_t\) can make \((R_t,Y_t)\) mutually singular whenever
\(\iprod{a_t}{f}\ne0\)). Write \(H_{t-1}=(U_1,Y_1,\ldots,U_{t-1},Y_{t-1})\)
and let \(K_t(du\mid h_{t-1})\) be the learner's conditional kernel for
\(U_t\) (shared by both experiments). For \(u\), put
\(\alpha(u)=\iprod{a(u)}{e}\), \(\zeta(u)=\iprod{a(u)}{f}\), and let
\(F_t^0(\cdot\mid u)=\mathcal L\{-\alpha(u)R_t\}\) and
\(F_t^{I,v}(\cdot\mid u)=\mathcal L\{-\alpha(u)R_t+\gamma v\zeta(u)\II{t\in I}\}\)
be the conditional feedback laws (independent of the past because \(R_t\)
is fresh). The conditional laws of \((U_t,Y_t)\) factor as
\(M_t^j(du,dy\mid h_{t-1})=K_t(du\mid h_{t-1})F_t^j(dy\mid u)\) for
\(j\in\{0,(I,v)\}\). Integrating over \(K_t(du\mid h_{t-1})\) via Tonelli
(with dominating kernel \(F_t^0+F_t^{I,v}\)) and applying
\eqref{eq:lower-one-step-hellinger} gives
\begin{align}
 d_t(h_{t-1})
 &:=H^2\bigl(M_t^0(\cdot\mid h_{t-1}),M_t^{I,v}(\cdot\mid h_{t-1})\bigr)
  =\int K_t(du\mid h_{t-1})H^2\bigl(F_t^0(\cdot\mid u),F_t^{I,v}(\cdot\mid u)\bigr)
  \nonumber\\
 &\le C_{\nu,\gamma}\II{t\in I}\E_0[1-\iprod{a_t}{e}\mid H_{t-1}=h_{t-1}].
 \label{eq:lower-common-kernel-hellinger}
\end{align}

\emph{Step 3: compose the one-round comparisons.}
Put \(S_t(h_{t-1})=\sum_{s=1}^td_s(h_{s-1})\) (\(S_0=0\)); each \(S_t\)
is \(\sigma(H_{t-1})\)-measurable and \(S_1\le\cdots\le S_T\). By Step~2
and the tower property,
\(\mu_0:=\E_0[S_T(H_{T-1})]\le C_{\nu,\gamma}\E_0\sum_{t\in I}(1-\iprod{a_t}{e})=\mu\).
Fix \(\eta>0\) and define an auxiliary transcript law \(\widetilde P_{I,v}\)
by setting \(\widetilde M_t^{I,v}=M_t^{I,v}\) when \(S_t(h_{t-1})\le\eta\)
and \(\widetilde M_t^{I,v}=M_t^0\) otherwise. Its one-step squared
Hellinger distance from \(M_t^0\) is
\(\widetilde d_t=\II{S_t\le\eta}d_t\); since \((S_t)\) is nondecreasing,
\(\sum_{t=1}^T\widetilde d_t\le\eta\) pathwise.

Write \(\rho_t=1-\tfrac12H^2(P_{0,1:t},\widetilde P_{I,v,1:t})\)
(\(\rho_0=1\)). Conditioning on \(H_{t-1}\) with dominating measure
\(\lambda_{t-1}=P_{0,1:t-1}+\widetilde P_{I,v,1:t-1}\) (densities
\(p_{t-1},\tilde q_{t-1}\)) gives
\(\rho_t=\rho_{t-1}-\tfrac12\int\sqrt{p_{t-1}\tilde q_{t-1}}\,\widetilde d_t\,d\lambda_{t-1}\).
Telescoping over \(t=1,\ldots,T\) and using
\(\sqrt{p_{t-1}\tilde q_{t-1}}\le\tfrac12(p_{t-1}+\tilde q_{t-1})\) yields
\[
 H^2(P_0,\widetilde P_{I,v})
 \le\frac12\E_0\Bigl[\sum_{t=1}^T\widetilde d_t\Bigr]
   +\frac12\E_{\widetilde P_{I,v}}\Bigl[\sum_{t=1}^T\widetilde d_t\Bigr]
 \le\frac{\mu_0+\eta}{2}\le\frac{\mu+\eta}{2},
\]
so \(\operatorname{TV}(P_0,\widetilde P_{I,v})\le\sqrt{(\mu+\eta)/2}\).
On \(G=\{S_T(H_{T-1})\le\eta\}\), \(P_{I,v}\) and \(\widetilde P_{I,v}\)
agree on every measurable subset, so
\(\operatorname{TV}(P_{I,v},\widetilde P_{I,v})\le\widetilde P_{I,v}(G^c)
 \le P_0(G^c)+\operatorname{TV}(P_0,\widetilde P_{I,v})\).
By Markov's inequality, \(P_0(G^c)\le\mu_0/\eta\le\mu/\eta\), and the
triangle inequality gives
\[
 \operatorname{TV}(P_{I,v},P_0)
 \le\operatorname{TV}(P_{I,v},\widetilde P_{I,v})+\operatorname{TV}(\widetilde P_{I,v},P_0)
 \le\frac{\mu}{\eta}+\sqrt{2(\mu+\eta)}. \qedhere
\]
\end{proof}

\subsection{From a low-information interval to the trade-off}

\begin{proofof}{Theorem~\ref{thm:lower-bound}.}
The zero-loss sequence in \eqref{eq:lower-assumption} forces \(\mathsf F\ge0\).
Put \(Z=\mathsf F+\mathsf B^2+1\ge2\) and
\(c_0=\min\{2^{-6},(2^{26}C_{\nu,\gamma})^{-1}\}\). Since \(R_{1:T}\) is
independent of the learner and \(\mathsf F,\mathsf B\) are deterministic,
averaging \eqref{eq:lower-assumption} over \(R_{1:T}\sim\nu^{\otimes T}\)
gives \(\bbP(\mathcal E_{\mathrm{lb}})\ge1-\delta_{\mathrm{lb}}\) for the
regret event \(\mathcal E_{\mathrm{lb}}\) under both \(\bbP_0\) and
\(\bbP_{I,v}\).

\emph{Step 1: control deviation under the null experiment.}
Under \(\theta_t^0=-R_te\), put \(c_t=\iprod{e_t}{e}\),
\(m_t=\iprod{x_t}{e}=(1-p_t)c_t\), \(\alpha_t=\iprod{a_t}{e}\),
\(D_T^x=\sum_{t=1}^T(1-m_t)\), and \(\bar X_T=\sum_{t=1}^TR_t(1-m_t)\).
The realised null regret is \(X_T=\sum_{t=1}^TR_t(1-\alpha_t)\in[0,T]\),
and independence of \(R_t\) from the round-\(t\) draw gives
\(z:=\E_0X_T=\E_0\bar X_T\). Since \(2(1-m_t)-\{1-(1-p_t)c_t^2\}=p_t+(1-p_t)(1-c_t)^2\ge0\) and
\(R_t\le1/2\),
\[
 \normt{G_t}{\theta_t^0}^2
 =R_t^2\{1-(1-p_t)c_t^2\}
 \le2R_t^2(1-m_t)\le R_t(1-m_t),
\]
so \(\Qcurv_T^0=\sum_{t=1}^T\normt{G_t}{\theta_t^0}^2\le\bar X_T\) pathwise.
By \eqref{eq:lower-assumption}, \(X_T\le T\), and Jensen's inequality,
\[
 z=\E_0[X_T\mathds 1_{\mathcal E_{\mathrm{lb}}}]+\E_0[X_T\mathds 1_{\mathcal E_{\mathrm{lb}}^c}]
  \le\mathsf F+\mathsf B\E_0\sqrt{\Qcurv_T^0}+T\delta_{\mathrm{lb}}
  \le\mathsf F+\mathsf B\sqrt z+\frac1{16}.
\]
Lemma~\ref{lem:root-absorption}(a) gives
\(z\le2\mathsf F+\mathsf B^2+1/8\le2Z\); since \(\E_0\alpha_t=\E_0m_t\)
and \(R_t\ge1/4\),
\(\E_0D_T^x=\E_0\sum_{t=1}^T(1-\alpha_t)\le4z\le8Z\).

\emph{Step 2: select a low-information interval.}
If \(Z\ge c_0T/2\), then \(Z\ge(c_0/2)T^{1/3}\mathsf B^{-2/3}\) directly.
Otherwise set \(k=\floor{c_0T/Z}\); since \(c_0T/Z\in(2,T/128]\),
\begin{equation}
 \frac{c_0T}{2Z}\le k\le\frac{c_0T}{Z},
 \qquad 1\le k\le T/8.
 \label{eq:lower-k-bounds}
\end{equation}
Partition \(\{\floor{T/2}+1,\ldots,T\}\) into
\(N=\floor{(T-\floor{T/2})/k}\ge T/(2k)-1\ge T/(4k)\) disjoint intervals
of length \(k\). Their expected deviations sum to at most
\(\E_0D_T^x\le8Z\), so some deterministic interval \(I\) satisfies
\begin{equation}
 \E_0\sum_{t\in I}(1-\alpha_t)\le\frac{32kZ}{T}\le32c_0.
 \label{eq:lower-low-info-interval}
\end{equation}

\emph{Step 3: turn an incorrect sign into regret.}
By \eqref{eq:lower-low-info-interval},
\(\mu:=C_{\nu,\gamma}\E_0\sum_{t\in I}(1-\alpha_t)\le32C_{\nu,\gamma}c_0\le2^{-21}\).
Lemma~\ref{lem:lower-radial-nuisance} with \(\eta=2^{-15}\) and the
triangle inequality yield
\begin{equation}
 \operatorname{TV}(P_{I,v},P_0)\le\frac{2^{-21}}{2^{-15}}+\sqrt{2(2^{-21}+2^{-15})}<\frac1{32},
 \qquad
 \operatorname{TV}(P_{I,+},P_{I,-})\le\frac1{16}.
 \label{eq:lower-tv}
\end{equation}
Put \(A=\sum_{t=1}^TR_t\le T/2\), \(L=(A^2+\gamma^2k^2)^{1/2}\), and
\(Z_I=\sum_{t\in I}\iprod{a_t}{f}\). Since
\(\twonorm{-R_te+\gamma vf}^2\le1/4+1/64<1\), the perturbed losses are
feasible with cumulative vector \(-Ae+\gamma vkf\) and optimal loss
\(-L\). Because \(L+A\le2T\),
\[
 \Reg_T^{I,v}=\sum_{t=1}^TR_t(1-\iprod{a_t}{e})+\gamma vZ_I+(L-A)
 \ge\gamma vZ_I+D,
 \qquad
 D:=\frac{\gamma^2k^2}{2T}\le L-A.
\]
Whenever \(\Reg_T^{I,v}<D/2\), we have \(\gamma vZ_I<-D/2<0\), so the
transcript-measurable test \(\widehat v=-\operatorname{sgn}(Z_I)\)
(\(\operatorname{sgn}(0)=1\)) recovers \(v\). Setting
\(E_+=\{\widehat v=+1\}\), \eqref{eq:lower-tv} gives
\(P_{I,+}(E_+^c)+P_{I,-}(E_+)\ge1-\operatorname{TV}(P_{I,+},P_{I,-})\ge15/16\),
so some sign \(v\in\{-1,1\}\) satisfies
\begin{equation}
 \bbP_{I,v}\{\Reg_T^{I,v}\ge D/2\}\ge P_{I,v}\{\widehat v\ne v\}\ge\frac{15}{32}.
 \label{eq:lower-testing-event}
\end{equation}

\emph{Step 4: control the quadratic form on the testing event.}
Since \(D_T^x\) is transcript-measurable, Markov's inequality and
\eqref{eq:lower-tv} give
\(P_{I,v}\{D_T^x>64Z\}\le P_0\{D_T^x>64Z\}+\operatorname{TV}(P_{I,v},P_0)\le1/8+1/32\),
hence
\begin{equation}
 \bbP_{I,v}\{D_T^x\le64Z\}\ge\frac{27}{32}.
 \label{eq:lower-small-mass-event}
\end{equation}
For every round, \(0\preceq G_t\preceq I\) and
\(e\transpose G_te=1-(1-p_t)c_t^2\le2(1-m_t)\), so
\(\normt{G_t}{\theta_t^{I,v}}^2\le2R_t^2e\transpose G_te+2\gamma^2\II{t\in I}
 \le(1-m_t)+2\gamma^2\II{t\in I}\) (using \(R_t\le1/2\)). Summing yields
\(\Qcurv_T^{I,v}\le64Z+2\gamma^2k\) on \eqref{eq:lower-small-mass-event}.

\emph{Step 5: solve the scalar trade-off.}
Intersecting \(\mathcal E_{\mathrm{lb}}\) with \eqref{eq:lower-testing-event}
and \eqref{eq:lower-small-mass-event} leaves probability at least
\(15/32+27/32-1-\delta_{\mathrm{lb}}\ge1/4>0\). On this nonempty
intersection, deterministic bounds force
\[
 \frac{\gamma^2k^2}{4T}=\frac D2
 \le\mathsf F+\mathsf B\sqrt{64Z+2\gamma^2k}
 \le9Z+\sqrt2\,\gamma\mathsf B\sqrt k
\]
(using \(\mathsf F\le Z\) and \(\mathsf B\le\sqrt Z\)). Substituting the
bounds on \(k\) from \eqref{eq:lower-k-bounds} yields
\begin{equation}
 \frac{\gamma^2c_0^2T}{16Z^2}
 \le9Z+\sqrt{2c_0}\,\gamma\mathsf B\sqrt{\frac{T}{Z}}.
 \label{eq:lower-two-cases}
\end{equation}
If \(9Z\) dominates the right side, then \(Z^3\ge\gamma^2c_0^2T/288\) and
\(Z\ge c_1T^{1/3}\ge c_1T^{1/3}\mathsf B^{-2/3}\) with
\(c_1=(\gamma^2c_0^2/288)^{1/3}\). Otherwise
\(Z^{3/2}\ge\gamma c_0^{3/2}\sqrt T/(32\sqrt2\,\mathsf B)\), giving
\(Z\ge c_2T^{1/3}\mathsf B^{-2/3}\) with
\(c_2=(\gamma c_0^{3/2}/(32\sqrt2))^{2/3}\). Combined with Step~2, the
first bound holds with \(c_{\nu,\gamma}=\min\{c_0/2,c_1,c_2\}>0\).

\emph{Step 6: verify prefix drift and eliminate the coefficient.}
For every realisation of the radii and either sign,
\(\twonorm{\sum_{s\le t}\theta_s^{I,v}}^2
 =(\sum_{s\le t}R_s)^2+\gamma^2|I\cap\{1,\ldots,t\}|^2\ge t^2/16\)
(and similarly on the null sequence). Finally, \(Z\ge\mathsf B^2\) gives
\(\mathsf B^{-2/3}\ge Z^{-1/3}\), so \eqref{eq:lower-conclusion} yields
\(Z\ge c_{\nu,\gamma}T^{1/3}Z^{-1/3}\) and hence
\(Z\ge c_{\nu,\gamma}^{3/4}T^{1/4}\).
\end{proofof}

\Needspace{8\baselineskip}
\section{Comparisons with earlier work}
\label{sec:extended-consequences}

\subsection{Regret rates and computational costs}
The tunings cited in Table~\ref{tab:closest-adversarial} all use the horizon
\(T\).

\emph{Oblivious and fixed-comparator rates.}
\citet[Theorem~6]{bubeck2012minimax} use
\(\eta=\sqrt{\log T/(2dT)}\) and \(\gamma=T^{-1/2}\) (with \(\eta d\le1/2\));
their radial gradient and coordinate sampler give \(O(d)\) ball updates.
They bound pseudo-regret, which equals expected hindsight regret for
deterministic losses.
\citet[Section~5.2]{abernethy2009beating} give
\(\widetilde O(d\sqrt T)\) with high probability for a fixed comparator
and \(\widetilde O(d^{3/2}\sqrt T)\) uniformly after a comparator-net
union. Their ball algorithm requires, per round, the radial minimiser of
the regularised objective and one element of an orthonormal basis of
\(e_t^\perp\); a Householder reflection supplies it in \(O(d)\) operations
(our count).

\emph{\citet[Theorems~4--5]{zimmert2022bias}.}
For general convex sets, they obtain adaptive high-probability
\(\widetilde O(d^2\sqrt T)\) via a polynomial-time entropic-barrier
method (assuming linear optimisation access) and \(\widetilde O(d\sqrt T)\)
via a lifted moment-body log-determinant method without an efficient
implementation supplied.
Their Remark~7 notes that the entropic-barrier analysis uses only
self-concordance and a sampler whose information matrix is proportional
to the inverse Hessian. Its comparator union bound over
a \(1/\sqrt T\)-cover, with \(\log|C|\le d\log(6\sqrt T)\), enters the
confidence term \(\iota=O(d\log T+\log(1/\delta))\), which sets the
bias scale \(\gamma\propto\sqrt\iota\); the positive bias cost is of order
\(\sqrt\iota\,\nu\sqrt{dT\log T}\) (their Appendix~F, Steps~3--4). Hence
\(\nu=O(1)\) on the ball still gives \(\widetilde O(d\sqrt T)\) (our
reading), while \(\nu=d\) for the entropic barrier gives \(d^2\sqrt T\).

\emph{\citet[Theorem~3.1]{lee2020bias}.}
For a \(\nu\)-self-concordant barrier (with \(\nu=O(1)\) on the ball),
they use barrier optimisation and matrix operations as in SCRiBLe and
obtain adaptive high-probability regret
\(\widetilde O(d^2\nu\sqrt{dT}+d^3\nu)\).
Specifically, for a fixed comparator their deviation bound is of order
\(d(\sqrt\nu+\rho)\sqrt{T\log(C/\delta)}\), where \(\rho\) is the
comparator's largest local norm and \(C\) is polylogarithmic (their
Lemma~B.10). Cancelling it with the negative terms of the learning-rate
increases requires \(\eta\) of order
\(1/(d^2\sqrt{T\log(C/\delta)})\) up to logarithms (their
Lemma~B.15 and proof of Theorem~3.1): one factor \(d\) comes from the
deviation and the other from the increase factor
\(\kappa=\exp\{1/(100d\log(\nu T))\}\), which keeps the learning rate
within a factor \(5\) over at most \(100d\log_2(\nu T)\) increases (their
Lemma~B.12). Against an adaptive adversary, a union bound over
\(e^{O(d\log T)}\) points replaces \(\log(C/\delta)\) by \(O(d\log T)\);
the regularisation cost \(\nu/\eta\) then gives \(\widetilde O(d^2\nu\sqrt{dT})\),
and \(\widetilde O(d^{5/2}\sqrt T)\) on the ball (our reading).

\paragraph{Stochastic parameter noise.}
For i.i.d.\ losses with covariance \(\Sigma\),
\citet[Theorems~3.7 and~3.9]{ezer2026parameter} give horizon-tuned, high-probability
pseudo-regret \(\widetilde O(d+\sqrt{dT\tr\Sigma})\) with known \(\tr\Sigma\) and
\(\widetilde O(dT^{1/3}+\sqrt{dT\tr\Sigma})\) with unknown
\(\tr\Sigma\), without minimum noise.
Our anytime bound (Corollary~\ref{cor:random-losses}) holds against adaptive
adversaries and adds \(\sqrt T\) to the known-trace rate, in place of their
\(dT^{1/3}\); the \(\sqrt T\) is absorbed for \(\tr\Sigma\ge1/(4\kappa_0)\), even
with singular covariance.

Their bounds hold at the same order for realised hindsight regret.
Theorems~3.7 and~3.9 bound the pseudo-regret
\(R_T^{\mathrm{pseudo}}=\sum_t\iprod{a_t}{\bar\theta}
+T\twonorm{\bar\theta}\) of the realised actions with probability at
least \(1-2\delta'\) on \(\ell_p\) balls, \(p\in(1,2]\); at \(p=2\) their
variance statistic is \(\tr\Sigma\). Write \(\theta_t=\bar\theta+\eta_t\),
where \(\eta_t\) is independent of the history and of the learner's draw
\(a_t\). The triangle inequality gives
\[
 \Reg_T
 =R_T^{\mathrm{pseudo}}+\sum_t\iprod{a_t}{\eta_t}
 +\twonorm*{T\bar\theta+\sum_t\eta_t}-T\twonorm{\bar\theta}
 \le R_T^{\mathrm{pseudo}}
 +\sum_t\iprod{a_t}{\eta_t}
 +\twonorm*{\sum_t\eta_t}.
\]
Both sums are martingales, for the filtration that reveals \(a_t\)
before \(\eta_t\), with increments of norm at most \(2\) and conditional
second moments at most \(\tr\Sigma\).
Proposition~\ref{prop:contractive-classical:fixed} with \(K_t=I\) bounds each
by \(\widetilde O(\sqrt{T\tr\Sigma}+1)\), which both rates absorb.

\subsection{Relation to earlier bias arguments}
\label{sec:relation}
Paying for estimation error with negative regret earned on trigger rounds
follows \citet{lee2020bias}, who raise the learning rate when their matrix
test fails, adding a negative term of the order of the comparator's local
norm (their Lemma~B.13) that dominates the deviation in every direction
(their Lemma~B.15). \citet[Section~2.1]{zimmert2022bias}, citing
\citet{foster2020adapting}, observe that such increases act as linear
biases along the regulariser's gradient and enter FTRL linearly (their
Theorem~1). Our test adapts \citeauthor{lee2020bias}; each contraction
adds the bias \(\epsilon_t\widehat\Theta_{t-1}=-\epsilon_t\beta_t\nabla F(z_t)\),
and both works lift actions by a constant coordinate as we lift
comparators in Step~4. On the ball,
\citet[Section~5.2]{abernethy2009beating} add a roundwise linear bias
bounding the comparator's variance, of the form of our gap
(Lemma~\ref{lem:directional-geometry}).

We differ in three ways driven by ball geometry. First, contraction credit
is an exact identity linear in the gap (Step~2), whereas
\citeauthor{lee2020bias} (Lemma~B.13) and
\citet[Appendix~F, Step~4]{zimmert2022bias} bound their negative terms
through local norms and self-concordance; similarly,
Lemma~\ref{lem:contracted-ftrl} replaces the infinite boundary barrier
value by \(\Gamma_T(v)\). Second,
\citet[Section~3.1]{zimmert2022bias} and
\citet[Section~5.2]{abernethy2009beating} bias every round and
\citeauthor{lee2020bias} trigger \(\widetilde O(d)\) times (Lemma~B.12),
whereas our block structure limits contractions to logarithmically many per
epoch, independently of \(d\) (Lemma~\ref{lem:trigger}). Third, all three
works take a union bound over an \(e^{O(d\log T)}\)-net
(\citealp[Theorem~3.1]{lee2020bias};
\citealp[Appendix~F and Theorem~5]{zimmert2022bias};
\citealp[Section~5.2]{abernethy2009beating}). For unit comparators, the
gap is a quadratic form in the lifted comparator, so accumulated credit is
a squared cover norm; one lifted vector-martingale bound then controls all
comparators (Step~4), removing a factor \(\sqrt d\).

\end{document}